\documentclass{article}

\usepackage[eandd, final]{neurips_2026}
\usepackage[utf8]{inputenc} % allow utf-8 input
\usepackage[T1]{fontenc}    % use 8-bit T1 fonts
\usepackage{hyperref}       % hyperlinks
\usepackage{url}            % simple URL typesetting
\usepackage{booktabs}       % professional-quality tables
\usepackage{amsfonts}       % blackboard math symbols
\usepackage{nicefrac}       % compact symbols for 1/2, etc.
\usepackage{microtype}      % microtypography
\usepackage{graphicx}
\usepackage{subcaption}
\usepackage{xspace}
\usepackage{multirow}
\usepackage{array}

\usepackage{pifont}
\usepackage{threeparttable}

\usepackage{fvextra}
\usepackage{caption} % Required for \captionof
\usepackage{tikz}
\usepackage{amsmath}

\usepackage[table,dvipsnames]{xcolor}
\usepackage{booktabs}
\usepackage{graphicx}
\usepackage{longtable}
\usepackage[ruled,vlined]{algorithm2e}
\usepackage{subcaption}

\usepackage[most]{tcolorbox}
\usepackage{tabularx}
\definecolor{HeatBlue}{rgb}{0.3,0.5,0.7}

\newcounter{prompt}

\usepackage{rotating}
\usepackage{tabularx}
\usepackage{array}
\usepackage{makecell}

\usepackage{adjustbox}

\usepackage{graphicx}
\usepackage{xcolor}
\usepackage{tikz}
\usepackage{array}
\usepackage{booktabs}
\usepackage{adjustbox}
\usepackage{booktabs}
\usepackage{pdflscape}

\newenvironment{PromptBlock}[1]{%
  \refstepcounter{prompt}%
  \begin{center}
  \textbf{Prompt~\theprompt: #1}
  \end{center}
}{}

\DefineVerbatimEnvironment{PromptText}{Verbatim}{
    frame=single,
    framerule=0.5pt,
    framesep=10pt,
    fontsize=\small,
    breaklines=true,    % Wraps long lines
    breakanywhere=true, % Forces wrapping of long URLs
    fontfamily=courier
}

\definecolor{colCAUS}{HTML}{D62728} % Red
\definecolor{colDETL}{HTML}{1F77B4} % Blue
\definecolor{colEVNT}{HTML}{2CA02C} % Green
\definecolor{colPROS}{HTML}{9467BD} % Purple
\definecolor{colSOC}{HTML}{8C564B}  % Brown
\definecolor{colSPAT}{HTML}{FF7F0E} % Orange
\definecolor{colTEMP}{HTML}{E377C2} % Pink

\newcommand{\stackedbar}[7]{%
  \begin{tikzpicture}[x=5cm, y=0.25cm, baseline=0ex]
    \fill[colCAUS] (0,0) rectangle (#1,1);
    \fill[colDETL] (#1,0) rectangle (#1+#2,1);
    \fill[colEVNT] (#1+#2,0) rectangle (#1+#2+#3,1);
    \fill[colPROS] (#1+#2+#3,0) rectangle (#1+#2+#3+#4,1);
    \fill[colSOC]  (#1+#2+#3+#4,0) rectangle (#1+#2+#3+#4+#5,1);
    \fill[colSPAT] (#1+#2+#3+#4+#5,0) rectangle (#1+#2+#3+#4+#5+#6,1);
    \fill[colTEMP] (#1+#2+#3+#4+#5+#6,0) rectangle (#1+#2+#3+#4+#5+#6+#7,1);
  \end{tikzpicture}%
}

\newcommand{\legendsquare}[1]{\tikz[baseline=0.4ex]{\fill[#1] (0,0) rectangle (0.25,0.25);}}

\newcommand{\egostream}{\textsc{EgoStream}\xspace}

\newcommand{\cmark}{\textcolor{green!60!black}{\ding{51}}}
\newcommand{\xmark}{\textcolor{red!75!black}{\ding{55}}}
\newcommand{\pmark}{\textcolor{orange!90!black}{$\triangle$}}

\newcolumntype{L}[1]{>{\raggedright\arraybackslash}p{#1}}
\newcolumntype{C}{>{\centering\arraybackslash}c}

\newcommand{\badge}[2]{%
  \begingroup
  \setlength{\fboxsep}{1.4pt}%
  \colorbox{#1!12}{\textcolor{#1!65!black}{\scriptsize\textsf{\textbf{#2}}}}%
  \endgroup
}

\newcommand{\bSink}{\badge{green}{SINK}}
\newcommand{\bHat}{\badge{pink}{HAT}}

\newcommand{\bFIFO}{\badge{gray}{FIFO}}
\newcommand{\bMerge}{\badge{teal}{MERGE}}
\newcommand{\bPrune}{\badge{red}{PRUNE}}

\newcommand{\bIntraPA}{\badge{violet}{INTRA-PA}}
\newcommand{\bInterPA}{\badge{purple}{INTER-PA}}

\newcommand{\bIntraCos}{\badge{orange}{INTRA-COS}}
\newcommand{\bInterCos}{\badge{brown}{INTER-COS}}

\newcommand{\bIntraCS}{\badge{magenta}{INTRA-CS}}
\newcommand{\bIntraVN}{\badge{olive}{INTRA-VN}}

\newcommand{\bOff}{\badge{NavyBlue}{OFFLOAD}}
\newcommand{\bQOff}{\badge{black}{Q-OFF}}

\newcommand{\bAdaHead}{\badge{cyan}{ADA-HEAD}}
\newif\ifshowrev \showrevtrue
\DeclareRobustCommand{\rev}[1]{\ifshowrev\textcolor{red}{#1}\else#1\fi}

\title{\egostream: A Diagnostic Benchmark for \\Streaming Episodic Memory in Egocentric Vision}
\author{%
  Rosario Forte \quad Giuseppe Lando \quad Antonino Furnari\\
  Department of Mathematics and Computer Science\\
  University of Catania\\
  \texttt{rosario.forte@phd.unict.it}\\
  \texttt{giuseppe.lando@studium.unict.it}\\
  \texttt{antonino.furnari@unict.it}
}

\newcommand{\regimetimeline}[7]{%
  \begin{tikzpicture}[baseline=-0.5ex, scale=0.48]
    \draw[gray!30, thick] (0,0) -- (6,0);
    \foreach \i/\act/\txt/\scl/\wgt in {
      0/#1/I/1/\bfseries,
      1/#2/S/1/\bfseries,
      2/#3/SM/0.75/\mdseries,
      3/#4/M/1/\bfseries,
      4/#5/ML/0.75/\mdseries,
      5/#6/L/1/\bfseries,
      6/#7/U/1/\bfseries%
    } {
      \ifnum\act=1
        \fill[teal!80] (\i,0) circle (0.35);
        \node[white, font=\sffamily\tiny\wgt, scale=\scl] at (\i,0) {\txt};
      \else
        \fill[gray!20] (\i,0) circle (0.35);
        \node[gray!50, font=\sffamily\tiny\wgt, scale=\scl] at (\i,0) {\txt};
      \fi
    }
  \end{tikzpicture}%
}

\begin{document}

\maketitle

\begin{abstract}
  Continuous episodic memory is a core capability for autonomous agents operating in dynamic, real-world environments, yet current streaming video benchmarks provide limited tools for diagnosing what models remember and for how long. We introduce \egostream, a diagnostic benchmark for streaming episodic memory evaluation in egocentric vision. \egostream organizes 2,250 curated questions along seven cognitive dimensions: detail, spatial, temporal, event, social, causal, and prospective memory. We introduce the Answer Validity Window (AVW), which specifies the temporal span an answer remains valid as the observed scene evolves. This allows us to expand the questions into 8,528 recall-conditioned evaluations, enabling controlled testing from instant to ultra-long-term recall while separating genuine model forgetting from natural world-state changes. We rigorously establish baseline performance through a unified streaming MLLM framework that compares several state-of-the-art memory-management mechanisms, covering sliding windows, attention sinks, KV-cache pruning, merging, adaptive head budgets, and offloading. 
  Experiments on a unified Qwen3-VL backbone, with consistent trends on Molmo2-8B and InternVL3.5-8B, reveal that comparable aggregate accuracies mask starkly different memory profiles. For instance, token pruning preserves fine-grained details and temporal structure significantly better than token merging, adaptive head budgets yield the highest aggregate semantic accuracy, while quantized offloading rescues ultra-long-term recall. Ultimately, all in-cache memory policies operate well below real-time (>1s per frame), and top performing in-cache methods ceil at about 45\% accuracy under a 9,804-token budget, exposing critical gaps in current architectures. \egostream provides the diagnostic testbed needed to close these gaps.
  \noindent\textbf{Project website:} \url{https://coai-iplab.github.io/EgoStream}
  %Experiments reveal that similar aggregate accuracies often mask different memory profiles; for instance, token pruning preserves fine-grained episodic details and temporal structure significantly better than token merging, while different distinct state-of-the-art memory management mechanisms only exhibit small performance differences when compared within our unified framework. \egostream provides a diagnostic testbed to build reliable, long-horizon embodied memory systems.
  %This innovation enables controlled evaluation from instant to ultra-long-term recall while separating genuine forgetting from natural state changes. The benchmark contains 2,250 curated questions expanded into 8,528 recall-conditioned evaluations under a strict streaming protocol. We further propose a unified streaming MLLM framework covering sliding windows, attention sinks, KV-cache pruning, merging, and offloading, enabling controlled comparison of memory-management mechanisms. Experiments reveal that aggregate accuracy masks substantial trade-offs among recall categories and regimes, demonstrating the validity of \egostream as a diagnostic testbed for building reliable embodied memory systems.
\end{abstract}

\section{Introduction}

To operate successfully in the physical world, an autonomous agent must store and recall episodic events across both short and long temporal horizons~\cite{tulving1972episodic,tulving2002episodic,clayton2003elements}. For visual embodied AI systems, streaming egocentric video offers a natural testbed for studying continuous memory, as it captures first-person experience as it unfolds over time~\cite{grauman2022ego4d,baermann2022keys}, unlike detached exocentric media such as movies or surveillance footage. This makes the egocentric domain ideal for real-world memory tasks, including wearable assistants and embodied agents that must provide context-aware support~\cite{grauman2022ego4d,plizzari2024outlook,yang2025egolife,EgoSchema,perrett2025hdepic}. In this setting, egocentric assistants must answer questions about past events (e.g., whether the user locked the door or where an object was placed),
%. Unlike offline video understanding systems, they cannot revisit the full video at query time. Instead, they must 
processing observations incrementally and maintaining a compact memory under bounded resources. This makes memory in egocentric vision particularly challenging: relevant evidence may appear long before it is queried, requiring the model to decide what to retain during streaming perception.

Despite recent progress in egocentric Video Question Answering (VideoQA), existing offline benchmarks still provide only a limited view of episodic memory~\cite{grauman2022ego4d,EgoSchema,di2024groundvqa,chen2025multihop,plizzari2025egotempo,perrett2025hdepic}.
While they have substantially advanced long-form understanding and reasoning,
%Most egocentric VideoQA benchmarks are designed for offline evaluation, where the entire video is available at query time~\cite{grauman2022ego4d,EgoSchema,di2024groundvqa,chen2025multihop,plizzari2025egotempo,perrett2025hdepic}.
%They have substantially advanced long-form video understanding, temporal grounding, and multi-step reasoning, but 
they do not isolate memory as a measurable variable. The temporal distance between the supporting evidence and the query is rarely controlled, and evaluation does not distinguish \textit{what} information must be remembered, \textit{how long} it must be retained, or \textit{when} a previously correct answer becomes invalid in a changing environment. 
Recent online and streaming benchmarks move closer to this setting~\cite{lin2024streamingbench,niu2025ovobench,xiong2025streamchat,wang2025omnimmi}, but 
%by posing questions during continuous video streams or multi-round interactions~\cite{lin2024streamingbench,niu2025ovobench,xiong2025streamchat,wang2025omnimmi}. 
%However, they 
primarily measure online  responsiveness or aggregate accuracy rather than providing a diagnostic account of memory quality.
%, causal interaction, or aggregate VideoQA accuracy, rather than providing a diagnostic account of memory quality. 
Consequently, models with very different memory profiles may obtain similar scores. One may preserve coarse event gist while forgetting object locations, whereas another may retain short-term spatial details but fail at long-horizon causal recall. 
This flattened assessment makes it difficult to determine which forms of episodic information are preserved by current streaming models and which degrade first.
This limitation has led to a proliferation of memory-management methods that address streaming constraints
%through techniques such as sliding window, attention sinks, KV-cache pruning, token compression, and retrieval-based memory
~\cite{xu2025streamingvlm,zhang2023ho,liu2023scissorhands,li2024snapkv,bolya2023token,yang2025streammem,kim2025infinipotv}. 
Evaluating these methods solely through aggregate performance masks the qualitative trade-offs induced by their architectural choices. 
%For instance, two models might achieve similar overall accuracy while exhibiting very different memory profiles: one may extend its retention window by discarding fine-grained visual details, while another may preserve high-fidelity spatial information but forget it over a shorter timeframe. 
We argue that streaming memory evaluation should measure not only whether a model answers correctly, but also what kind of episodic information is retained and for how long.

\begin{figure}
    \centering
    \includegraphics[width=\linewidth]{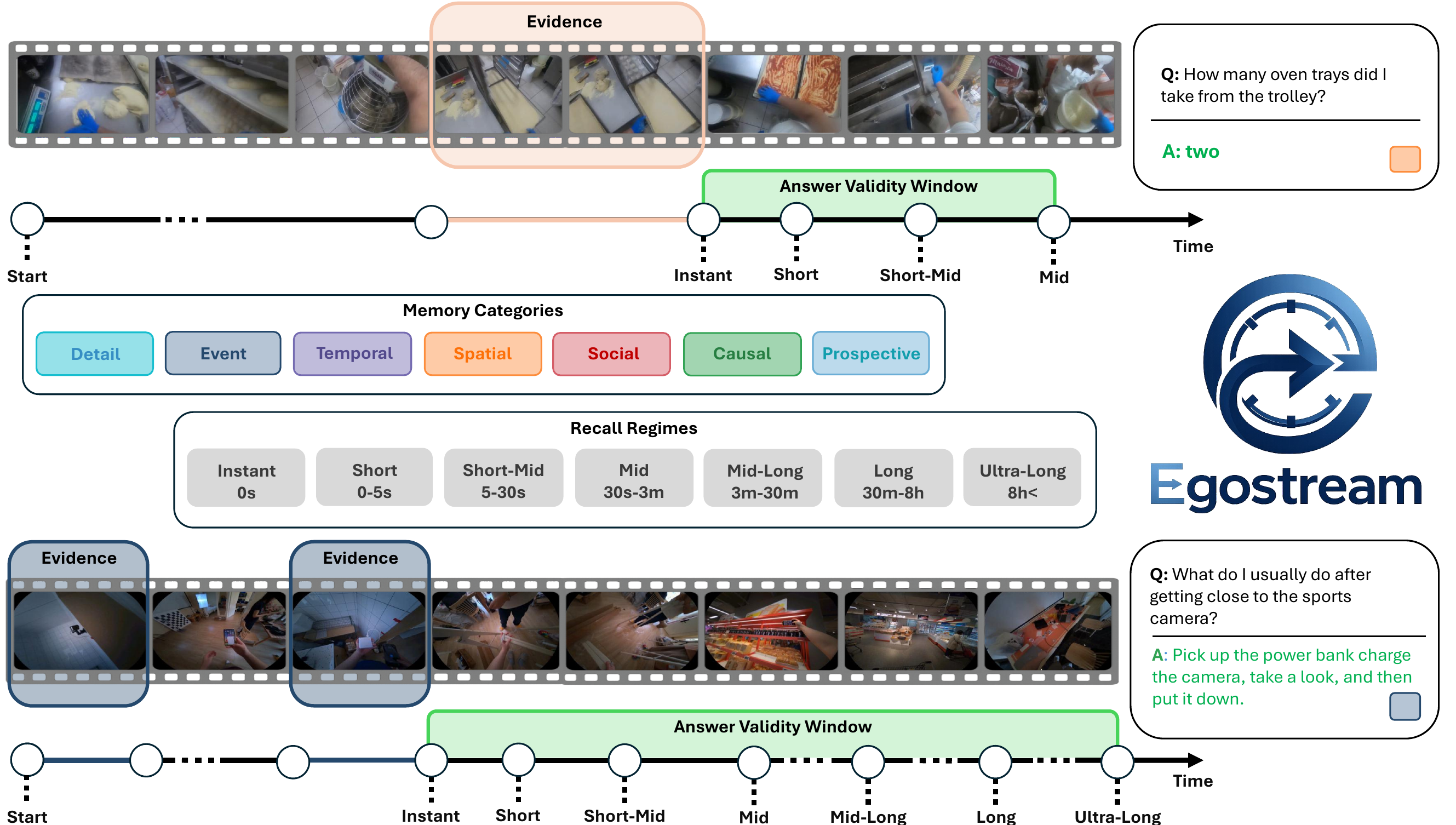}
    \hfill
    %\caption{Isn't it the best teaser image you have ever seen? I bet ya!}
    \caption{\textbf{Overview of the \egostream benchmark.} Each question is grounded in specific visual evidence, assigned a semantic category, and evaluated across multiple recall regimes. We introduce the Answer Validity Window as the temporal span over which an answer can be verified to remain factually correct, constraining recall evaluations to this window to ensure consistency.}
    \label{fig:teaser}
\end{figure}

To support progress in 
%the evaluation and design of 
visual episodic memory systems, we introduce \egostream(Figure~\ref{fig:teaser}), a new benchmark curated from existing egocentric VideoQA datasets~\cite{di2024groundvqa,chen2025multihop,plizzari2025egotempo,perrett2025hdepic,yang2025egolife}
%, where questions are systematically 
and reorganized through a cognitive- and application-oriented framework.
Drawing inspiration from cognitive accounts of episodic memory and related memory functions~\cite{tulving1972episodic,tulving2002episodic,burgess2002spatial,friedman1993memory,radvansky2014event,trabasso1985causal,mcdaniel2007prospective}, we curate queries to test seven 
%into standardized formats aimed at testing distinct 
dimensions of memory: \textit{detail}, \textit{spatial}, \textit{temporal}, \textit{event}, \textit{social}, \textit{causal}, and \textit{prospective} memory. 
To support a consistent evaluation of memory retention over time, we introduce the concept of \textit{Answer Validity Window} (AVW)---a temporal span indicating the timeframe during which a specific past observation remains factually true before being overturned by subsequent events, or before its validity can no longer be verified from the observed stream. 
Because continuous egocentric vision is dynamic, annotating AVWs
%While most VideoQA benchmarks implicitly assume that questions can be answered independently of when they are asked, continuous egocentric vision is dynamic, and subsequent observations frequently invalidate past facts. 
ensures that forgetting is not conflated with scene evolution. Each question is then evaluated at up to seven recall horizons: \textit{instant }, \textit{short-term },
\textit{short-mid-term }, \textit{mid-term }, \textit{mid-long-term }, \textit{long-term }, and \textit{ultra-long-term }.
This organization is also consistent with classic studies showing that memory should be evaluated across controlled retention intervals~\cite{ebbinghaus1885,murre2015replication,dudai2004memory,moscovitch2006consolidation}.

%To establish a clear picture of current progress, we evaluate existing state-of-the-art streaming methods on our benchmark~\cite{chen2024videollmonline,di2025rekv,ning2025livevlm,kim2025infinipotv,zhang2025flashvstream,xu2025streamingvlm,chen2025streamingtom,zhang2026hermes,liu2026thinkstream}. 
%However, evaluating SOTA architectures can confound memory mechanisms with differences in visual encoders, language backbones, and training data. 
To establish a picture of current progress we implement core memory-management strategies (e.g., sliding windows, attention sinks, token pruning, token merging)
%: sliding windows, attention sinks, token pruning, token merging, intra-/inter-frame reduction, and offloading 
within a unified MLLM architecture. 
This controlled setup reveals that similar aggregate accuracy can hide distinct memory profiles, with policies preserving different semantic information and degrading differently across recall horizons.
Our experiments reveal that established state-of-the-art memory management mechanisms exhibit only marginal performance differences when evaluated within our unified framework and that aggregate accuracies mask divergent memory profiles. For instance, token pruning preserves fine-grained details and temporal structure significantly better than token merging. This highlights the need to evaluate streaming memory by the specific episodic information preserved and its temporal accessibility, rather than just average accuracy.

Our contributions are threefold. 1) We introduce \egostream, a benchmark for streaming episodic memory in egocentric vision, built from five egocentric datasets and comprising 2,250 validated memory questions expanded into 8,528 recall-conditioned evaluations. 2) We organize episodic memory along seven semantic dimensions and introduce the Answer Validity Window (AVW) to measure temporal retention as the world evolves.
%: detail, spatial, temporal, event, social, causal, and prospective, introducing the Answer Validity Window(AVW) concept, which measures how long an answer remains valid as the world evolves. 
3) We provide a unified streaming MLLM framework for the controlled comparison of memory-management techniques.
%controlled comparison of memory-management techniques under a fixed backbone. 
\noindent\textbf{Annotations, code, and evaluation scripts are publicly available at:} \url{https://coai-iplab.github.io/EgoStream}.
\section{Related Work}
%Our work relates to previous efforts in the formulation of episodic memory in egocentric vision, the design of benchmarks for continuous video understanding, and the development of memory-constrained streaming architectures, which we discuss in the following.

\textbf{Episodic Memory in Egocentric Vision}
Episodic memory in egocentric vision initially focused on offline temporal localization~\cite{grauman2022ego4d} and grounded QA~\cite{baermann2022keys, di2024groundvqa, jiang2023single}. Recognizing the continuous nature of first-person experience, recent efforts have shifted toward online streaming. While some approaches track memory through object-centric representations~\cite{manigrasso2026esom, goletto2024amego, fan2025embodied}, VideoQA emerge as a suitable protocol for episodic retrieval, accelerated by MLLMs~\cite{di2024groundvqa}. This shift has driven diverse continuous memory strategies, including dynamic KV-cache retrieval~\cite{di2025rekv}, cache compression~\cite{kim2025infinipotv, chen2025streamingtom}, and hierarchical memory reuse~\cite{zhang2026hermes, zhang2025flashvstream, xu2025streamingvlm}.
However, these works treat streaming memory as a modeling challenge, evaluated implicitly through end-task VideoQA. Repeated stream queries directly probe whether an observed fact is retained, forgotten, or superseded. \egostream builds on this by making episodic recall the primary evaluation target, not merely a means for downstream VideoQA.

\begin{table*}[t]
\centering
\scriptsize
\caption{\egostream versus related egocentric and streaming VideoQA benchmarks. 
%\cmark indicates the presence of a property. 
%\xmark indicates its absence, and \pmark indicates a \emph{partial} match. 
\pmark\xspace denotes that a property is partially covered, but not in an explicit, controlled, and systematic form.}
\label{tab:benchmark_comparison}
%\resizebox{\textwidth}{!}{%
\begin{tabular}{lccccccc}
\toprule
\textbf{Benchmark} &
\shortstack{\textbf{Egocentric}} &
\shortstack{\textbf{Online /}\\\textbf{streaming}} &
\shortstack{\textbf{Evidence}\\\textbf{grounding}} &
\shortstack{\textbf{Memory}\\\textbf{taxonomy}} &
\shortstack{\textbf{Answer}\\\textbf{Validity Window}} &
\shortstack{\textbf{Multiple recall}\\\textbf{regimes per QA}} \\% &
%\shortstack{\textbf{Num.}\\\textbf{questions}} \\
\midrule
MovieChat-1K~\cite{Song2023MovieChatFD}        & \xmark & \xmark & \xmark & \pmark & \xmark & \xmark \\%& 14K \\
StreamingBench~\cite{lin2024streamingbench}    & \xmark & \cmark & \xmark & \xmark & \xmark & \xmark \\%& 4.5K \\
EgoSchema~\cite{EgoSchema}                     & \cmark & \xmark & \xmark & \xmark & \xmark & \xmark \\%& 5.03K \\
MM-Ego~\cite{ye2025mmego}         & \cmark & \xmark & \xmark & \pmark & \xmark & \xmark \\%& 7.03K \\
EgoTextVQA~\cite{Zhou_2025_CVPR}               & \cmark & \cmark & \xmark & \xmark & \xmark & \xmark \\%& 7.06K \\
EgoTempo~\cite{plizzari2025egotempo}           & \cmark & \xmark & \pmark & \pmark & \xmark & \xmark \\%& 0.5K \\
OVO-Bench~\cite{niu2025ovobench}               & \pmark & \cmark & \xmark & \pmark & \xmark & \xmark \\%& 2.81K \\
GroundVQA~\cite{di2024groundvqa}               & \cmark & \xmark & \cmark & \xmark & \xmark & \xmark \\%& 0.5K \\
Multi-Hop EgoQA~\cite{chen2025multihop}        & \cmark & \xmark & \cmark & \xmark & \xmark & \xmark \\%& 1.08K \\
HD-EPIC~\cite{perrett2025hdepic}               & \cmark & \xmark & \cmark & \xmark & \xmark & \xmark \\%& 26.65K \\
EgoLifeQA~\cite{yang2025egolife}               & \cmark & \xmark & \cmark & \pmark & \xmark & \xmark \\%& 0.5K \\
\midrule
\textbf{\egostream{} (ours)}                   & \textbf{\cmark} & \textbf{\cmark} & \textbf{\cmark} & \textbf{\cmark} & \textbf{\cmark} & \textbf{\cmark} \\%& \textbf{2.3K} \\
\bottomrule
\end{tabular}%
%}
\end{table*}

\textbf{VideoQA and Streaming Benchmarks.}
Existing benchmarks have significantly advanced long-form video comprehension~\cite{Song2023MovieChatFD}, with recent egocentric datasets specializing this setting for first-person reasoning. These works cover a broad spectrum of abilities, including long-form QA~\cite{EgoSchema,ye2025mmego,Zhou_2025_CVPR}, temporal grounding and multi-step reasoning~\cite{di2024groundvqa,chen2025multihop,plizzari2025egotempo}, and fine-grained daily-life assistance~\cite{perrett2025hdepic,yang2025egolife}. In parallel, streaming-oriented benchmarks like StreamingBench~\cite{lin2024streamingbench} and OVO-Bench~\cite{niu2025ovobench} evaluate online responsiveness by posing queries during continuous video streams rather than assuming full offline access.
Despite this progress, memory remains an implicit byproduct of broader video understanding or online interaction. As shown in Table~\ref{tab:benchmark_comparison}, prior benchmarks address aspects like egocentric perception, streaming evaluation, or temporal reasoning. However, none jointly provide an explicit memory taxonomy, Answer Validity Windows, and controlled recall regimes. \egostream addresses this gap by turning streaming episodic memory into a specific, diagnostic evaluation target.

\textbf{Streaming Memory Architectures in Video LLMs.}
To bound memory growth in continuous video streams, recent Video-LLMs employ mechanisms such as recency-based retention~\cite{xiao2024efficient}, KV-cache pruning~\cite{zhang2023ho,li2024snapkv,kim2025infinipotv,chen2025streamingtom}, token merging~\cite{bolya2023token,zhang2024cam,wang2025model}, and external retrieval~\cite{di2025rekv}. These methods reduce spatial and temporal redundancy~\cite{kim2025infinipotv,chen2025streamingtom,yang2025streammem}, often aided by quantization~\cite{chen2025streamingtom}.While they improve latency and efficiency, they are typically evaluated only via aggregate online accuracy, leaving unclear what each policy preserves and how abilities decay over time. \egostream addresses this by evaluating representative memory families under a unified backbone, enabling controlled analysis of what streaming models remember, for how long, and which dimensions fail first.

\section{The \egostream Benchmark}
\label{sec:benchmark}

%We build \egostream to serve as a controlled, diagnostic evaluation of streaming episodic memory. 
%In the following sections, we detail the conceptual framework and construction of the benchmark.

\subsection{Source Curation and Initial Set of Questions}
\label{sec:curation}
% GOAL: Establish the physical foundation of the dataset so the reviewer accepts the empirical data-driven claims in later sections.
% TO DO:
% 1. Source Datasets: Briefly introduce the 5 foundation datasets (Ego4D, EgoLife, EgoTempo, Multi-Hop EgoQA, HD-EPIC). Explain that combining them is essential for capturing a heterogeneous mix of memory demands.
% 2. Base Curation: Explain how annotations were adapted for streaming. 
%    - Rewriting non-recall queries.
%    - Generating multiple-choice distractors via Gemini 3.1 and filtering them.
% 3. Evidence Grounding: Explain how the exact supporting moments were identified and clipped to ensure the answer is observed only once.
% 4. The Baseline Result: Conclude by stating this process yields ~2,400 clean, human-verified, temporally grounded question-answer pairs. 
% (Note: Do NOT discuss the categories or the AVW yet. Save those for the next sections).
We curate question-answer sets from five egocentric VideoQA datasets: \textbf{Ego4D Episodic Memory VQA}~\cite{baermann2022keys}, \textbf{EgoLife}~\cite{yang2025egolife}, \textbf{EgoTempo}~\cite{plizzari2025egotempo}, \textbf{Multi-Hop EgoQA}~\cite{chen2025multihop}, and \textbf{HD-EPIC}~\cite{perrett2025hdepic}. These sources cover complementary aspects of egocentric perception, including episodic memory and temporal localization, long-context life assistance, fine-grained temporal reasoning, multi-hop grounded reasoning, and kitchen activity recognition. %Aggregating them captures diverse episodic-memory challenges while mitigating single-dataset biases.
To unify these annotations under a streaming evaluation protocol, we apply a standardized curation pipeline. First, we ensure that all queries are explicit retrospective memory probes. Questions from Ego4D~\cite{baermann2022keys}, EgoLife~\cite{yang2025egolife}, and EgoTempo~\cite{plizzari2025egotempo} are already suitable, while those from HD-EPIC~\cite{perrett2025hdepic} and Multi-Hop EgoQA~\cite{chen2025multihop} are rewritten using Gemini 3.1~\cite{gemini}, e.g., \textit{``What is the best description for why the person performed the action <tip bowl> in video 1?''} is rewritten into \textit{``What was the person trying to achieve by tipping the bowl?''}

To support a standardized $4$-way multiple-choice evaluation, we use Gemini 3.1~\cite{gemini} to generate three plausible distractors for open-ended datasets, namely EgoTempo~\cite{plizzari2025egotempo} and Multi-Hop EgoQA~\cite{chen2025multihop}. We then apply an LLM-based quality-control step to remove ambiguous, trivial, or overly similar negatives, while preserving the original multiple-choice options for the remaining datasets (see Appendix~\ref{app:negative_generation}).
Finally, each question is attached the temporal segment containing the visual evidence needed to infer the correct answer, which we refer to as the \textit{evidence moment} (see Appendix~\ref{app:evidence_moments}). These annotations are inherited from the source datasets.
%Finally, measuring memory retention requires identifying the temporal origin of each fact. For each curated question-answer pair, we inherit from the source datasets the temporal annotation of the video segment containing the visual evidence needed to infer the correct answer, which we refer to as the \textit{evidence moment} (see Appendix~\ref{app:evidence_moments}). 
%This curation process yields an initial set of $2,634$ question-answer pairs.
% Finally, all curated samples undergo a comprehensive human validation step. Expert annotators manually verify the accuracy of the question-answer pairs, the plausibility of the distractors, and the temporal exactness of the evidence moments. This quality assurance step filters out roughly 10\% of the initial pool, yielding a high-fidelity foundation of $2,400$ clean, human-verified, temporally grounded question--answer--video instances. This robust base collection serves as the prerequisite substrate for the diagnostic taxonomy and temporal annotations detailed in the following sections.
%
This yields $2,634$ QA sets, that undergo a comprehensive human validation step. Annotators manually verify the accuracy of the question-answer pairs and the plausibility of the distractors (See Appendix~\ref{app:human_validation}). This step filters out roughly 11\% of the initial pool, yielding $2,335$ human-verified, temporally grounded video QA instances.% See Appendix~\ref{app:egostream_benchmark_construction} for more details on the curation and validation process.

\subsection{The Cognitive Dimensions of Episodic Memory}
\label{sec:cognitive_dimensions}
% GOAL: Define the Semantic Axis (what the model remembers) and prove the dataset is semantically diverse.
% TO DO:
% 1. The Concept: Introduce the 7 cognitive dimensions grounded in the "W-question" framework (Detail, Spatial, Temporal, Event, Social, Causal, Prospective). Reference Table 2 for definitions and examples.
% 2. The HITL Annotation: Explain how these labels were assigned to the 2,400 base questions. 
%    - Mention the iterative cycle (LLM seed -> human correction -> in-context learning -> human verification).
%    - Emphasize that this was done using *text only* to keep the taxonomy independent of visual biases.
%    - Explain the difference between strict Primary labels (using the priority cascade) and inclusive Secondary tags.
% 3. Empirical Proof: Direct the reviewer to the figures to prove the benchmark is balanced.
%
% FIGURES TO REFERENCE HERE:
% - Table 2 (Taxonomy definitions and examples)
% - Figure 3 (Distribution of question categories across datasets)
% - Figure 8 (Co-occurrences between primary and secondary labels)

\begin{figure}[t]
\begin{minipage}[b]{0.48\linewidth}
    \centering
    \parbox[c][3cm][c]{\linewidth}{%
        \centering
        \includegraphics[width=\linewidth,height=5cm,keepaspectratio]{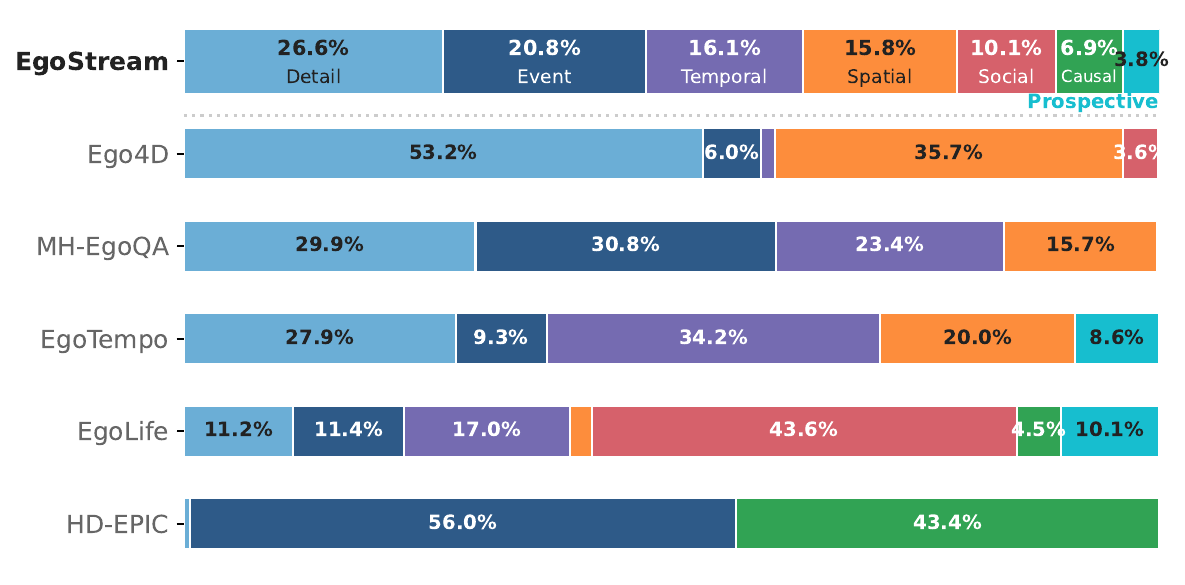}
    }
    \captionof{figure}{Distribution of question categories in \egostream and across subsets.}
    \label{fig:category_sunburst}
\end{minipage}
\hfill
\begin{minipage}[b]{0.48\linewidth}
    \centering
    \parbox[c][3cm][c]{\linewidth}{%
        \centering
        \includegraphics[width=\linewidth,height=5cm,keepaspectratio]{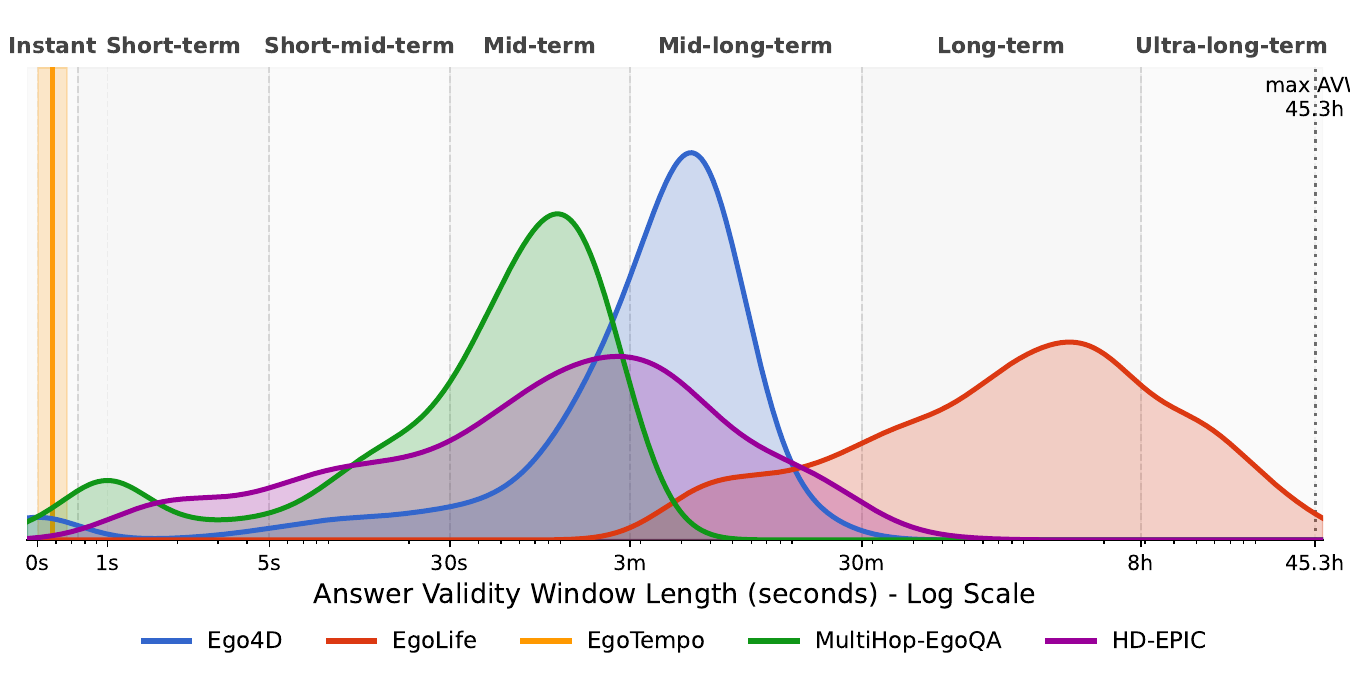}
    }
    \captionof{figure}{Recall regimes align with the empirical distribution of AVW lengths across datasets.}
    \label{fig:avw_taxonomy}
\end{minipage}
\end{figure}

\egostream isolates \textit{what} a model remembers by partitioning queries into seven dimensions grounded in foundational cognitive theory~\cite{tulving1972episodic, tulving2002episodic}. We structure this taxonomy around the natural ``W-question'' framework of human recall. Building upon the core episodic triad of \textit{what}, \textit{where}, and \textit{when}~\cite{clayton2003elements, burgess2002spatial, friedman1993memory}, we incorporate expanded dimensions critical for embodied agents: procedural actions (\textit{how})~\cite{radvansky2014event}, social context (\textit{who})~\cite{grunewald2026episodic, stewardson2023episodic}, causal reasoning (\textit{why})~\cite{trabasso1985causal}, and prospective intentions (\textit{what next})~\cite{mcdaniel2007prospective}. These categories are defined and summarized with examples in Table~\ref{tab:memory_taxonomy}. %, are defined below.
%
% \textbf{Detail Memory} (\textit{What}). Evaluates the retention of high-fidelity perceptual details, such as fine-grained object identities, counts, and states, rather than just coarse semantic concepts. 
%
% \textbf{Spatial Memory} (\textit{Where}). Probes the location, placement, or spatial relation of entities, testing whether an agent can anchor past observations to a persistent map of the physical environment.
%
% \textbf{Temporal Memory} (\textit{When}). Concerns the ordering of events and time-indexed states, testing whether a model can reconstruct the sequential timeline of an episode.
%
% \textbf{Event Memory} (\textit{How / Occurred}). Isolates the model's understanding of action boundaries, gist, and state changes, distinguishing the execution of an event from its mere temporal sequence.
%
% \textbf{Social Memory} (\textit{Who}). Tests the model's ability to recall the identities, co-presence, and social roles of people involved in past events, as observed from the egocentric point of view.
%
% \textbf{Causal Memory} (\textit{Why}). Goes beyond retrieving \textit{what} happened to probe whether the model can recover the underlying reasoning or physical causality behind a past event or decision.
%
% \textbf{Prospective Memory} (\textit{What next}). Tests the ability to encode a plan or intention and successfully retrieve it at the appropriate future moment to guide deferred behavior.
\begin{table}[t]
\centering
\caption{Taxonomy of episodic memory dimensions, cognitive probes, and example queries.}
\label{tab:memory_taxonomy}
\resizebox{\columnwidth}{!}{%
\begin{tabular}{@{}p{1.5cm}llp{6.5cm}@{}}
\toprule
\textbf{Category} & \textbf{Cognitive Probe} & \textbf{Focus} & \textbf{Example Query} \\ \midrule
\textbf{Detail} & \textit{What} & Object identity, attributes, counts & \textit{``What did I put in the microwave?''} \\
\textbf{Spatial} & \textit{Where} & Location, placement, spatial relations & \textit{``Where did I put the pair of scissors?''} \\
\textbf{Temporal} & \textit{When} & Order, sequence, time-indexed states & \textit{``Where was the knife before I picked it?''} \\
\textbf{Event} & \textit{How / Occurred} & Action gist, execution, state changes & \textit{``Did I close the circuit breaker box?''} \\
\textbf{Social} & \textit{Who} & People, interactions, co-presence & \textit{``Who was with me when I ate?''} \\
\textbf{Causal} & \textit{Why} & Reasons, explanations for past events & \textit{``Why didn't we buy the chestnut kernels?''} \\
\textbf{Prospective} & \textit{What next} & Future plans and intended actions & \textit{``What should I buy to decorate my room?''} \\ \bottomrule
\end{tabular}%
}
\end{table}
Because episodic queries frequently involve overlapping cognitive demands (for instance, \textit{``Who was with me when I operated the machine?''} contains both social and temporal cues) we assign each question one primary label and zero or more secondary tags, following the taxonomy in Table~\ref{tab:memory_taxonomy}. To efficiently scale this process, we employed an iterative Human-in-the-Loop (HITL) annotation pipeline (see Appendix~\ref{app:annotation_protocol}).
%Initial LLM predictions were manually corrected to form a high-quality seed set, which then guided subsequent LLM labeling via in-context learning. This cycle was repeated until every primary and secondary label was completely human-verified. Crucially, this semantic categorization was performed using only the text of the questions and answers, ensuring the taxonomy remains strictly independent of the visual evidence. 
After this process $85$ questions were discarded for not aligning to any of the category or being unclear, leaving $2,250$ valid questions (comprising 578 from Multi-Hop EgoQA~\cite{chen2025multihop}, 498 from Ego4D~\cite{baermann2022keys}, 433 from EgoLife~\cite{yang2025egolife}, 425 from EgoTempo~\cite{plizzari2025egotempo}, and 316 from HD-EPIC~\cite{perrett2025hdepic}). Figure~\ref{fig:category_sunburst} reports the distribution of primary categories and across data subsets. For reference, the text-only baseline obtains an average accuracy of 25.42\% demonstrating no strong text bias. See Appendix~\ref{app:text_baseline} for more details.
%Complete details of the HITL workflow, instructions given to annotators, and the exact prompt templates are provided in the Appendix~\ref{app:annotation_protocol}.

\subsection{The Answer Validity Window and Recall Regimes}
\label{sec:avw_and_regimes}

Time is the second fundamental dimension of episodic memory recall. To systematically evaluate this axis, we introduce the concept of \emph{Answer Validity Window} (AVW), i.e., the temporal interval during which a question’s original answer remains semantically valid after all its supporting visual evidence has been observed; this window is a conservative bound, since the answer may remain true beyond it but its validity can no longer be verified from the observed stream.
%This notion is related to prior attempts to characterize temporal difficulty, such as EgoSchema's temporal certificate sets~\cite{EgoSchema}, but it serves a different role. Rather than quantifying how much temporal context is needed to answer a question, AVW constrains whether that answer is still admissible as the world state evolves. 
Identifying this validity window is critical in continuous egocentric streams, where facts are inherently transient. An initially correct answer to a query such as ``Is the glass full?'' can rapidly become obsolete due to subsequent actions (e.g., drinking), object displacements (e.g., moving the glass), or broader scene evolutions.
%
%We systematically annotate the AVW for each curated question according to the specific characteristics of its source dataset. For \textbf{EgoTempo}, which explicitly targets immediate temporal reasoning, the AVW corresponds exactly to 0, corresponding to the question asked at the end of the input trimmed video. For \textbf{Ego4D-VQA} and \textbf{Multi-Hop EgoQA}, the AVW spans from the end of the ground-truth evidence moment to the end of the provided clip. For \textbf{EgoLife}, it extends from the end of the evidence moment up to the specific long-term query time originally provided by the dataset authors. For \textbf{HD-EPIC}, we prompt Gemini 3.1 to evaluate the provided narrations and identify all video chunks where the answer is observable. If an identified chunk overlaps significantly ($\text{IoU} > 0.5$) with the ground-truth moment $M$, we isolate the immediately preceding and succeeding valid chunks, $A$ and $B$. To prevent models from exploiting duplicate evidence, we trim the input video strictly from the end of $A$ to the beginning of $B$. Consequently, the AVW spans from the end of $M$ to the start of $B$, guaranteeing the visual evidence appears exactly once. A manual review confirmed the accuracy and consistency of these automated annotations(See Appendix~\ref{app:avw_recall_Regimes}).
%
We annotate the AVW using the temporal structure of each source dataset. For EgoTempo, the AVW is $0$, as questions are asked at the end of the trimmed input video. For Ego4D-VQA and Multi-Hop EgoQA, it spans from the end of the evidence moment to the end of the provided clip, while for EgoLife it extends to the long-term query time defined by the original dataset. For HD-EPIC, where validity intervals are not directly provided, we use Gemini 3.1~\cite{gemini} to identify narration chunks in which the answer is observable and derive the AVW by trimming duplicate evidence around the ground-truth moment. All annotations were manually reviewed for consistency. The review covered both the evidence moment and the expiration time for every question, and only 1.96\% of the windows required an adjustment (see Appendix~\ref{app:avw_audit}).
%Full dataset-specific procedures are reported in Appendix~\ref{app:avw_recall_Regimes}.

Associating an AVW to each question enables us to evaluate episodic memory at multiple recall regimes, reminiscent of classic memory testing paradigms~\cite{ebbinghaus1885, murre2015replication}. In practice, we sample multiple query times between the end of the relevant moment and the end of the validity interval, ensuring that forgetting is not conflated with answer invalidation caused by scene evolution.  Drawing on established memory consolidation timescales~\cite{dudai2004memory, moscovitch2006consolidation}, application-oriented considerations~\cite{plizzari2024outlook}, and empirical support in our data (see Figure~\ref{fig:avw_taxonomy}), we define seven recall regimes spanning a log-scaled progression. Table~\ref{tab:recall_regimes} reports the proposed recall regimes and terminology, with supporting references from psychology and the empirical distribution of memory categories within each temporal window.
\begin{table}[t]
\centering
\caption{Taxonomy of recall regimes aligned with memory consolidation timescales and distribution of memory categories that resolve within each temporal window.}
\label{tab:recall_regimes}
\resizebox{\columnwidth}{!}{%
\begin{tabular}{lllll} 
\toprule
\textbf{Recall Regime} & \textbf{Window} & \textbf{Memory Mechanism} & \textbf{Psychological Basis} & \textbf{AVW Lengths Query Distribution} \\
\midrule
Instant         & $0s$         & Perceptual buffer & Encoding/immediate retrieval~\cite{cowan2008differences} & \stackedbar{0.000}{0.200}{0.074}{0.059}{0.002}{0.143}{0.235} \\
Short-term      & $0s$--$5s$   & Working memory    & Active maintenance~\cite{baddeley2003working}            & \stackedbar{0.017}{0.032}{0.063}{0.000}{0.002}{0.017}{0.019} \\
Short-mid-term  & $5s$--$30s$  & Working memory    & Attention-dependent retention~\cite{baddeley2003working}& \stackedbar{0.057}{0.079}{0.097}{0.000}{0.003}{0.044}{0.036} \\
Mid-term        & $30s$--$3m$  & Transitional      & Early Synaptic Consolidation~\cite{dudai2004memory}      & \stackedbar{0.078}{0.273}{0.307}{0.000}{0.008}{0.176}{0.158} \\
Mid-long-term   & $3m$--$30m$  & Intermediate      & Synaptic consolidation~\cite{mcgaugh2000memory}          & \stackedbar{0.062}{0.317}{0.147}{0.022}{0.090}{0.182}{0.014} \\
Long-term       & $30m$--$8h$  & Long-term         & Late Synaptic Consolidation~\cite{dudai2004memory}       & \stackedbar{0.035}{0.052}{0.060}{0.044}{0.201}{0.006}{0.078} \\
Ultra-long-term & $>8h$        & Long-term         & Hippocampal-cortical transfer~\cite{moscovitch2006consolidation} & \stackedbar{0.003}{0.021}{0.013}{0.008}{0.065}{0.003}{0.049} \\
\bottomrule
\multicolumn{5}{c}{
    \vspace{0.5em} 
    \scriptsize 
    \legendsquare{colCAUS} Causal \quad
    \legendsquare{colDETL} Detail \quad
    \legendsquare{colEVNT} Event \quad
    \legendsquare{colPROS} Prospective \quad
    \legendsquare{colSOC} Social \quad
    \legendsquare{colSPAT} Spatial \quad
    \legendsquare{colTEMP} Temporal
} \\
\end{tabular}
}
\end{table}
%Each question in \egostream is evaluated at one or more recall times, constrained by its AVW. We first assign a natural query time anchored to the conventions of the source dataset: an instant query ($0$\,s delay) for \textbf{EgoTempo}, the physical end of the provided clip for \textbf{Ego4D} and \textbf{Multi-Hop EgoQA}, the end of our custom-trimmed clip for \textbf{HD-EPIC}, and the specifically annotated long-term query time for \textbf{EgoLife}. 
%
%To test whether models can comprehend and answer the queries independently of memory decay, all questions are explicitly evaluated in the \textit{instant} regime (i.e., queried immediately after the ground-truth evidence moment is observed). Beyond these anchored points, we associate each question with additional recall delays by sampling query times from a mixture of distributions. The parameters of these distributions are derived from our taxonomy of recall regimes and adjusted to fit the empirical distribution of the source data. This expansion maps each question to a realistic set of temporal delays that fall within its AVW, ultimately yielding the \textcolor{red}{ $8,528$} distinct question-recall evaluations that comprise the benchmark. See Appendix~\ref{app:recall_regimes} for more details.
%
Given the AVW annotations, we expand each curated question into one or more recall evaluations. Each question is first evaluated in the \textit{instant} regime, immediately after the evidence moment, to measure whether the model can answer before memory decay becomes a factor. We then sample additional query times within the AVW according to our recall-regime taxonomy, with distribution parameters adjusted to match the empirical temporal structure of the source data. This produces a realistic set of recall delays across short- and long-horizon settings, yielding $8,528$ question-recall evaluations. Further details are provided in Appendix~\ref{app:recall_regimes}.

\subsection{Evaluation Protocol and Metrics}
\label{sec:evaluation}

To establish \egostream as a standardized reference for future researchers, we explicitly define inference rules and scoring metrics, as well as provide scripts to standardize evaluation.

\textbf{Streaming Constraints.} The benchmark strictly enforces a streaming inference protocol. Models must process the video sequentially: a query issued at timestamp $t$ must be answered utilizing only the visual content observed up to $t$. Models are expected to maintain a compact internal representation of the past, avoiding to store raw video frames in an unbounded buffer to be reprocessed at query time. The visual processing required to form this memory must be purely query-agnostic, as the evaluation query is only revealed to the model at the exact moment of recall.

\textbf{Evaluation.} \egostream evaluates models by submitting the $2,250$ valid questions across their applicable recall regimes, resulting in $8,528$ individual inferences. Performance is assessed via standard accuracy (broken down across semantic categories and recall regimes) on the $4$-way multiple-choice setup, with a chance-level baseline corresponding to $0.25$. While individual methods are free to choose their specific system prompts, we utilize a unified prompt and answer-selection scheme for fair comparisons (see Appendix~\ref{app:model_inference}). Methods are evaluated under limited memory budgets for active memory, and reporting the average processing time per frame. Options are shuffled once with a fixed
seed, identically for all methods. See Appendix~\ref{app:mcq_bias} for more details.

% \textbf{Efficiency Metrics.} Beyond accuracy, practical streaming systems must be evaluated on their operational footprint. \textcolor{red}{We assess models under constrained memory budgets, reported uniformly in Gigabytes (GB) to ensure architecture-independent comparisons. We report }

% \textbf{Diagnostic Breakdown.} To enable granular diagnostic assessments, we break down accuracy across the semantic question categories and the temporal recall regimes. The overall benchmark performance is computed by averaging across these categories and regimes. This unweighted averaging prevents the evaluation from being dominated by short-term retrieval or dataset-specific biases, ensuring models are rewarded for balanced retention across the full spectrum of episodic memory demands.

% ---------------------------------------------------------
% SECTION 4: THE UNIFIED PIPELINE
% ---------------------------------------------------------
\section{A Unified Framework for Streaming Episodic Memory}
\label{sec:unified_framework}

Current evaluations of state-of-the-art models often confound memory mechanisms with orthogonal design choices, such as varying vision encoders or pre-training mixtures. To enable controlled diagnostic evaluations, we introduce a unified streaming Multimodal Large Language Model (MLLM) framework that isolates memory management as an independent variable. We instantiate it on a frozen Qwen3-VL (8B)~\cite{bai2025qwen3vltechnicalreport} backbone, which we use for our main experiments; since the framework is backbone-agnostic, swapping the vision encoder and decoder is enough to port it to other MLLMs, as we show in Section~\ref{sec:backbones} with InternVL~3.5 (8B)~\cite{wang2025internvl35} and Molmo~2 (8B)~\cite{molmo2}. In all cases, the framework adapts to streaming VideoQA by continuously appending newly encoded visual tokens to the KV cache. This enables incremental video processing without recomputing past contexts at every query. By keeping the encoder, projector, LLM decoder, and prompting scheme strictly fixed, we provide a testbed for benchmarking distinct memory techniques. Our unified pipeline allows us to replicate popular state-of-the-art streaming video understanding approaches~\cite{chen2025streamingtom,kim2025infinipotv,yang2025streammem,xu2025streamingvlm,di2025rekv} on a plug-and-play memory manager that operates directly in KV-cache.

\begin{figure}[t]
    \centering
     \includegraphics[width=\linewidth]{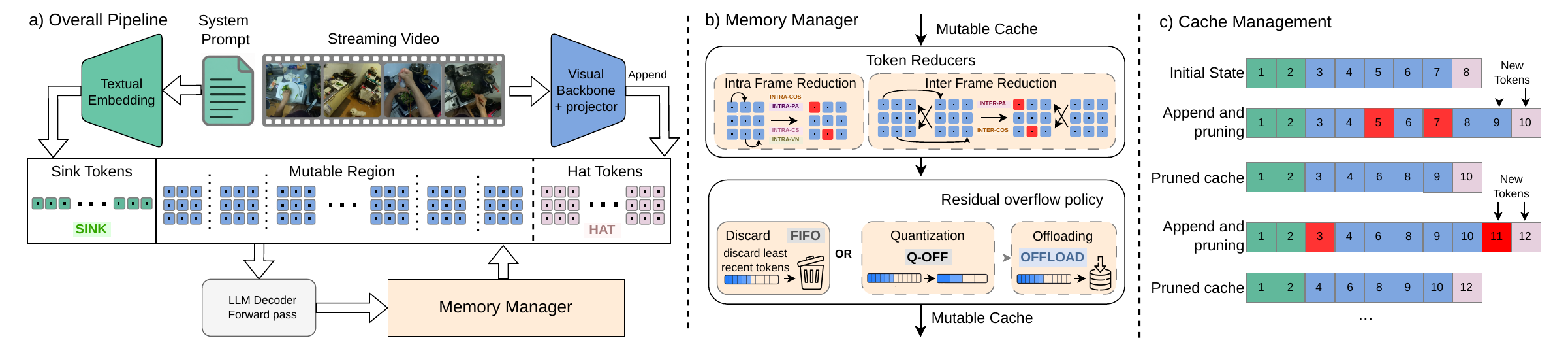}

    \caption{\textbf{Unified streaming framework.} The proposed framework processes frames sequentially while maintaining an incrementally updated KV cache. 
    (a) The cache is partitioned into three logical regions: protected \emph{sink tokens} (\bSink), corresponding to the system prompt; a \emph{mutable region}, containing historical visual tokens subject to memory management; and protected \emph{hat tokens} (\bHat), corresponding to the most recent visual observations. 
    The pipeline incrementally appends new visual tokens to the cache, passes the updated cache to the LLM decoder, and calls the memory manager whenever the mutable cache exceeds its budget. 
    (b) The memory manager operates only on the mutable cache. Optional reducer modules apply pruning or merging according to intra-frame, inter-frame, or global Ada Head token-selection criteria. If the mutable cache still exceeds the live budget, an overflow policy either discards excess tokens or externalizes them through offloading or quantized offloading. 
    (c) During the streaming inference, each new frame is encoded and appended to the cache. A fixed number of recent visual tokens enter the protected region, older tokens move into the mutable region, and the selected memory policy updates the mutable cache to satisfy the memory budget.}
    \label{fig:unified_pipeline}
\end{figure}

Figure~\ref{fig:unified_pipeline} illustrates our unified streaming memory framework, detailing the cache organization, memory management, and update dynamics. The core principle of our approach is to restrict memory operations—such as merging, pruning, and offloading—strictly to historical visual evidence. This ensures that both the system prompt and the most recent observations are always preserved. As the video stream progresses, the memory manager continuously enforces a strict budget on the active cache. At question time, the model answers using this curated cache, drawing upon the most relevant available evidence selected from live blocks or, when retrieval is enabled, offloaded memory.

\subsection{Memory Management Strategies}
\label{sec:memory_strategies}
% TO DO: Provide a brief survey of the techniques you are integrating. Use paragraph headings for readability.
% - Sliding Window / FIFO: Discuss naive context windows and their limitations for long-term recall.
% - KV-Cache Pruning/Eviction: Discuss token eviction strategies (e.g., H2O, StreamingLLM's attention sinks).
% - Token Merging/Compression: Discuss techniques that fuse redundant visual tokens.
% - Retrieval-Augmented / Bank-based: Discuss methods that externalize past KV states to CPU/disk and retrieve them dynamically.

% Our controlled setting evaluates a common family of bounded-memory mechanisms--- pruning, merging, retrieval, and offloading---under the same MLLM backbone, following prior KV-cache and streaming-video memory work~\cite{zhang2023ho,xiao2024efficient,li2024snapkv,kim2025infinipotv,chen2025streamingtom,xu2025streamingvlm,di2025rekv}. 
% %
% In the following, we discuss the main Memory Management Strategies evaluated within our unified framework.

In the following, we detail the main Memory Management Strategies evaluated within our framework, which draw upon established techniques from prior KV-cache and streaming-video memory research. See Appendix~\ref{app:implementation_details} for more details.

\textbf{Protected cache regions.}
Following attention-sink and streaming-inference methods, which show that structural tokens are critical for stable long-context decoding~\cite{xu2025streamingvlm, kim2025infinipotv, zhang2023ho}, we preserve the system prompt as fixed \emph{sink tokens} (\bSink). 
We also reserve a small protected region for the most recent visual tokens, denoted \emph{hat tokens} (\bHat), analogous to recent-token budgets in KV-cache reduction and streaming methods~\cite{zhang2023ho,xiao2024efficient,kim2025infinipotv}. 
This organization (Figure~\ref{fig:unified_pipeline}a) preserves instruction-following and grounding for recent content, restricting memory management to the mutable historical cache.

\textbf{Intra-frame reduction.}
Treating the mutable cache as an undifferentiated pool can discard temporal coverage. To prevent this, we evaluate frame-local reducers that compress redundant evidence within individual frames (Figure~\ref{fig:unified_pipeline}b). We apply pruning or merging based on four criteria: cosine redundancy (\bIntraCos) between adjacent key states; pseudo-attention salience (\bIntraPA) to discard low-attended tokens~\cite{chen2025streamingtom}; value-norm ranking (\bIntraVN) to prune low-norm states~\cite{kim2025infinipotv}; and sink-guided cross-attention (\bIntraCS) to remove tokens least attended by the system prompt~\cite{yang2025streammem}.

\textbf{Inter-frame reduction.}
Adjacent egocentric frames often share highly redundant content, enabling temporal compression. Following prior streaming-video methods, our memory manager compares neighboring frames in the mutable cache to prune redundant tokens when the budget is exceeded. We evaluate two cross-frame reducers: cosine redundancy (\bInterCos), which matches tokens using cosine similarity~\cite{chen2025streamingtom}, and pseudo-attention salience (\bInterPA), which uses cross-frame pseudo-attention to estimate token importance~\cite{yang2025streammem} (Figure~\ref{fig:unified_pipeline}b).

\textbf{Adaptive head budget.} We also evaluate the Ada-KV~\cite{feng2025adakv} (\bAdaHead) technique, which scores mutable tokens with the attention of recent visual queries and redistributes a fixed per-layer budget non-uniformly among KV heads within the mutable token budget (Appendix~\ref{app:adakv}).

\textbf{Residual overflow policy.}
When the reduced cache still exceeds its budget, we apply an overflow policy (Figure~\ref{fig:unified_pipeline}b): either \bFIFO, which discards the oldest tokens~\cite{xiao2024efficient}, or \emph{offloading} (\bOff), which preserves the oldest excess chunks on disk for retrieval at question time via cosine similarity~\cite{di2025rekv}. To minimize storage costs, we also evaluate an \texttt{int8} \emph{quantized offloading} variant (\bQOff)~\cite{chen2025streamingtom}.

% ---------------------------------------------------------
% SECTION 5: EXPERIMENTS AND RESULTS
% ---------------------------------------------------------
\section{Experiments and Results}
\label{sec:experiments}

We evaluate the capabilities of current streaming architectures on \textsc{EgoStream} via a controlled diagnostic evaluation using our unified framework, an analysis of semantic profiles and temporal decay, and a comparison of state-of-the-art models. Configuration and hardware detailed in Appendix~\ref{app:hardware_configuration}.

\begin{table}[t]
\centering
\caption{\textbf{Evaluation of memory-management strategies.} Accuracy is reported across semantic categories and temporal recall regimes, alongside macro-averages and per-frame processing time. Configurations are mapped to their closest equivalent in literature wrt isolated memory strategies.}
\label{tab:controlled_eval}
\ifcsname c@rowcount\endcsname\else\newcounter{rowcount}\fi
\setcounter{rowcount}{0}
\renewcommand{\arraystretch}{1.2}
\resizebox{\textwidth}{!}{%
\begin{tabular}{@{}r p{4cm} c c c c c c c c c c c c c c c c c c@{}}
\toprule
 &  &  & \multicolumn{8}{c}{\textbf{Semantic Categories}} & \multicolumn{8}{c}{\textbf{Temporal Recall Regimes}} & \multicolumn{1}{c}{\textbf{Proc.}} \\
\cmidrule(lr){4-11}
\cmidrule(lr){12-19}
%\cmidrule(l){20-20}
 & \textbf{Strategy} & \textbf{Configuration} & \textbf{Avg} & \textbf{SPAT} & \textbf{TEMP} & \textbf{DETL} & \textbf{EVNT} & \textbf{SOC} & \textbf{CAUS} & \textbf{PROS} & \textbf{Avg} & \textbf{Inst} & \textbf{Short} & \textbf{S-Mid} & \textbf{Mid} & \textbf{ML} & \textbf{Long} & \textbf{Ultra} & \textbf{(s)} \\
\midrule
% \multicolumn{20}{@{}l}{\textit{Text based inference}} \\

% \refstepcounter{rowcount}\arabic{rowcount}\label{row:qwen8b_text_only} & Qwen 8B text only & Text-only MCQA & \cellcolor[HTML]{FFF5F0}\textcolor[HTML]{000000}{25.42} & \cellcolor[HTML]{FFF5F0}\textcolor[HTML]{000000}{23.06} & \cellcolor[HTML]{FFF5F0}\textcolor[HTML]{000000}{24.06} & \cellcolor[HTML]{FFF5F0}\textcolor[HTML]{000000}{24.27} & \cellcolor[HTML]{FFF5F0}\textcolor[HTML]{000000}{27.71} & \cellcolor[HTML]{FFF5F0}\textcolor[HTML]{000000}{34.63} & \cellcolor[HTML]{FFF5F0}\textcolor[HTML]{000000}{23.27} & \cellcolor[HTML]{FFF5F0}\textcolor[HTML]{000000}{20.93} & -- & -- & -- & -- & -- & -- & -- & -- & \cellcolor[HTML]{FFF5F0}\textcolor[HTML]{000000}{0.06} \\

% \midrule
\multicolumn{20}{@{}l}{\textit{Sliding-window baselines}} \\

\refstepcounter{rowcount}\arabic{rowcount}\label{row:q3vl8b_sim_sw_pura} & \bFIFO & Sliding Window & \cellcolor[HTML]{F7FCF5}\textcolor[HTML]{000000}{29.53} & \cellcolor[HTML]{F7FCF5}\textcolor[HTML]{000000}{30.50} & \cellcolor[HTML]{F7FCF5}\textcolor[HTML]{000000}{33.53} & \cellcolor[HTML]{F7FCF5}\textcolor[HTML]{000000}{30.62} & \cellcolor[HTML]{F7FCF5}\textcolor[HTML]{000000}{30.02} & \cellcolor[HTML]{D5EFCF}\textcolor[HTML]{000000}{26.85} & \cellcolor[HTML]{F7FCF5}\textcolor[HTML]{000000}{25.94} & \cellcolor[HTML]{D3EECD}\textcolor[HTML]{000000}{29.28} & \cellcolor[HTML]{F7FCF5}\textcolor[HTML]{000000}{28.33} & \cellcolor[HTML]{F7FCF5}\textcolor[HTML]{000000}{31.83} & \cellcolor[HTML]{F7FCF5}\textcolor[HTML]{000000}{31.09} & \cellcolor[HTML]{F7FCF5}\textcolor[HTML]{000000}{28.71} & \cellcolor[HTML]{F7FCF5}\textcolor[HTML]{000000}{27.80} & \cellcolor[HTML]{CBEAC4}\textcolor[HTML]{000000}{28.42} & \cellcolor[HTML]{00441B}\textcolor[HTML]{FFFFFF}{28.12} & \cellcolor[HTML]{AEDEA7}\textcolor[HTML]{000000}{22.37} & \cellcolor[HTML]{FFF5F0}\textcolor[HTML]{000000}{1.05} \\

\refstepcounter{rowcount}\arabic{rowcount}\label{row:q3vl8b_sw_sink}\label{row:q3vl8b_sw_sink_aaaaa} & \bFIFO\ \bSink & StreamVLM~\cite{xu2025streamingvlm} & \cellcolor[HTML]{005924}\textcolor[HTML]{FFFFFF}{44.74} & \cellcolor[HTML]{005F26}\textcolor[HTML]{FFFFFF}{57.95} & \cellcolor[HTML]{248C46}\textcolor[HTML]{FFFFFF}{51.22} & \cellcolor[HTML]{005D25}\textcolor[HTML]{FFFFFF}{54.90} & \cellcolor[HTML]{004A1E}\textcolor[HTML]{FFFFFF}{46.36} & \cellcolor[HTML]{2C944C}\textcolor[HTML]{FFFFFF}{28.87} & \cellcolor[HTML]{006027}\textcolor[HTML]{FFFFFF}{38.68} & \cellcolor[HTML]{005E26}\textcolor[HTML]{FFFFFF}{35.20} & \cellcolor[HTML]{117B38}\textcolor[HTML]{FFFFFF}{39.05} & \cellcolor[HTML]{005120}\textcolor[HTML]{FFFFFF}{55.06} & \cellcolor[HTML]{00441B}\textcolor[HTML]{FFFFFF}{55.10} & \cellcolor[HTML]{006D2C}\textcolor[HTML]{FFFFFF}{49.18} & \cellcolor[HTML]{208843}\textcolor[HTML]{FFFFFF}{40.00} & \cellcolor[HTML]{7FC97F}\textcolor[HTML]{000000}{29.34} & \cellcolor[HTML]{6BC072}\textcolor[HTML]{000000}{24.93} & \cellcolor[HTML]{D3EECD}\textcolor[HTML]{000000}{19.74} & \cellcolor[HTML]{FFF4EF}\textcolor[HTML]{000000}{1.07} \\

\midrule
\multicolumn{20}{@{}l}{\textit{Similarity-based merge \bFIFO\ \bSink\ \bHat +}} \\

\refstepcounter{rowcount}\arabic{rowcount}\label{row:q3vl8b_sim_merge_intra_pseudo_att} & \bMerge\ \bIntraPA & CaM-like~\cite{zhang2024cam} & \cellcolor[HTML]{218943}\textcolor[HTML]{FFFFFF}{41.87} & \cellcolor[HTML]{17813D}\textcolor[HTML]{FFFFFF}{54.25} & \cellcolor[HTML]{43AC5E}\textcolor[HTML]{FFFFFF}{48.20} & \cellcolor[HTML]{17803C}\textcolor[HTML]{FFFFFF}{51.49} & \cellcolor[HTML]{248C46}\textcolor[HTML]{FFFFFF}{42.48} & \cellcolor[HTML]{5DB96B}\textcolor[HTML]{FFFFFF}{28.27} & \cellcolor[HTML]{0D7836}\textcolor[HTML]{FFFFFF}{37.48} & \cellcolor[HTML]{98D594}\textcolor[HTML]{000000}{30.92} & \cellcolor[HTML]{2C944C}\textcolor[HTML]{FFFFFF}{37.73} & \cellcolor[HTML]{067230}\textcolor[HTML]{FFFFFF}{52.47} & \cellcolor[HTML]{016E2D}\textcolor[HTML]{FFFFFF}{51.98} & \cellcolor[HTML]{3BA458}\textcolor[HTML]{FFFFFF}{44.00} & \cellcolor[HTML]{52B365}\textcolor[HTML]{FFFFFF}{37.20} & \cellcolor[HTML]{BAE3B3}\textcolor[HTML]{000000}{28.65} & \cellcolor[HTML]{B7E2B1}\textcolor[HTML]{000000}{23.48} & \cellcolor[HTML]{66BD6F}\textcolor[HTML]{000000}{26.32} & \cellcolor[HTML]{FC8B6B}\textcolor[HTML]{000000}{2.87} \\

\refstepcounter{rowcount}\arabic{rowcount}\label{row:q3vl8b_sim_merge_intra_cosine} & \bMerge\ \bIntraCos & CaM-like~\cite{zhang2024cam} & \cellcolor[HTML]{067230}\textcolor[HTML]{FFFFFF}{43.42} & \cellcolor[HTML]{006729}\textcolor[HTML]{FFFFFF}{57.30} & \cellcolor[HTML]{329B51}\textcolor[HTML]{FFFFFF}{49.83} & \cellcolor[HTML]{04702F}\textcolor[HTML]{FFFFFF}{53.28} & \cellcolor[HTML]{005522}\textcolor[HTML]{FFFFFF}{45.87} & \cellcolor[HTML]{A4DA9E}\textcolor[HTML]{000000}{27.52} & \cellcolor[HTML]{006328}\textcolor[HTML]{FFFFFF}{38.53} & \cellcolor[HTML]{7CC87C}\textcolor[HTML]{000000}{31.58} & \cellcolor[HTML]{258D47}\textcolor[HTML]{FFFFFF}{38.10} & \cellcolor[HTML]{005924}\textcolor[HTML]{FFFFFF}{54.44} & \cellcolor[HTML]{005E26}\textcolor[HTML]{FFFFFF}{53.20} & \cellcolor[HTML]{0C7735}\textcolor[HTML]{FFFFFF}{48.18} & \cellcolor[HTML]{2A924A}\textcolor[HTML]{FFFFFF}{39.40} & \cellcolor[HTML]{F7FCF5}\textcolor[HTML]{000000}{27.50} & \cellcolor[HTML]{CFECC9}\textcolor[HTML]{000000}{22.90} & \cellcolor[HTML]{C2E7BB}\textcolor[HTML]{000000}{21.05} & \cellcolor[HTML]{FCB296}\textcolor[HTML]{000000}{2.33} \\

\midrule
\multicolumn{20}{@{}l}{\textit{Similarity-based hybrid merge + prune \bFIFO\ \bSink\ \bHat +}} \\

\refstepcounter{rowcount}\arabic{rowcount}\label{row:q3vl8b_sim_merge_intra_pseudo_att_inter_pseudo_att} & 
\multicolumn{2}{l}{ \bMerge\ \bIntraPA\ \bPrune\ \bInterPA }& \cellcolor[HTML]{067230}\textcolor[HTML]{FFFFFF}{43.44} & \cellcolor[HTML]{006D2C}\textcolor[HTML]{FFFFFF}{56.66} & \cellcolor[HTML]{147E3A}\textcolor[HTML]{FFFFFF}{52.62} & \cellcolor[HTML]{006A2B}\textcolor[HTML]{FFFFFF}{53.89} & \cellcolor[HTML]{005F26}\textcolor[HTML]{FFFFFF}{45.33} & \cellcolor[HTML]{F7FCF5}\textcolor[HTML]{000000}{26.10} & \cellcolor[HTML]{006B2B}\textcolor[HTML]{FFFFFF}{38.23} & \cellcolor[HTML]{8ACE88}\textcolor[HTML]{000000}{31.25} & \cellcolor[HTML]{268E47}\textcolor[HTML]{FFFFFF}{38.01} & \cellcolor[HTML]{005723}\textcolor[HTML]{FFFFFF}{54.61} & \cellcolor[HTML]{006227}\textcolor[HTML]{FFFFFF}{52.87} & \cellcolor[HTML]{16803C}\textcolor[HTML]{FFFFFF}{47.35} & \cellcolor[HTML]{228A44}\textcolor[HTML]{FFFFFF}{39.87} & \cellcolor[HTML]{A8DCA2}\textcolor[HTML]{000000}{28.88} & \cellcolor[HTML]{F7FCF5}\textcolor[HTML]{000000}{21.45} & \cellcolor[HTML]{C2E7BB}\textcolor[HTML]{000000}{21.05} & \cellcolor[HTML]{8A0812}\textcolor[HTML]{FFFFFF}{5.32} \\

\midrule
\multicolumn{20}{@{}l}{\textit{Similarity-based prune \bFIFO\ \bSink\ \bHat +}} \\

\refstepcounter{rowcount}\arabic{rowcount}\label{row:q3vl8b_sim_prune_intra_cross_sink} & \bPrune\ \bIntraCS & StreamMem~\cite{yang2025streammem} & \cellcolor[HTML]{18813D}\textcolor[HTML]{FFFFFF}{42.39} & \cellcolor[HTML]{29914A}\textcolor[HTML]{FFFFFF}{52.25} & \cellcolor[HTML]{1D8640}\textcolor[HTML]{FFFFFF}{51.80} & \cellcolor[HTML]{107B37}\textcolor[HTML]{FFFFFF}{52.10} & \cellcolor[HTML]{0D7836}\textcolor[HTML]{FFFFFF}{43.85} & \cellcolor[HTML]{BCE4B5}\textcolor[HTML]{000000}{27.23} & \cellcolor[HTML]{03702E}\textcolor[HTML]{FFFFFF}{37.93} & \cellcolor[HTML]{7CC87C}\textcolor[HTML]{000000}{31.58} & \cellcolor[HTML]{157F3B}\textcolor[HTML]{FFFFFF}{38.88} & \cellcolor[HTML]{0A7633}\textcolor[HTML]{FFFFFF}{52.12} & \cellcolor[HTML]{117B38}\textcolor[HTML]{FFFFFF}{50.64} & \cellcolor[HTML]{289049}\textcolor[HTML]{FFFFFF}{45.82} & \cellcolor[HTML]{2F974E}\textcolor[HTML]{FFFFFF}{39.07} & \cellcolor[HTML]{CBEAC4}\textcolor[HTML]{000000}{28.42} & \cellcolor[HTML]{3BA458}\textcolor[HTML]{FFFFFF}{25.80} & \cellcolor[HTML]{278F48}\textcolor[HTML]{FFFFFF}{30.26} & \cellcolor[HTML]{B01217}\textcolor[HTML]{FFFFFF}{4.91} \\

\refstepcounter{rowcount}\arabic{rowcount}\label{row:q3vl8b_sim_prune_intra_vnorm} & \bPrune\ \bIntraVN & InfiniPot-V (VaN)~\cite{kim2025infinipotv} & \cellcolor[HTML]{006D2C}\textcolor[HTML]{FFFFFF}{43.79} & \cellcolor[HTML]{0B7734}\textcolor[HTML]{FFFFFF}{55.54} & \cellcolor[HTML]{077331}\textcolor[HTML]{FFFFFF}{53.67} & \cellcolor[HTML]{0D7835}\textcolor[HTML]{FFFFFF}{52.49} & \cellcolor[HTML]{005221}\textcolor[HTML]{FFFFFF}{45.98} & \cellcolor[HTML]{238B45}\textcolor[HTML]{FFFFFF}{29.02} & \cellcolor[HTML]{006B2B}\textcolor[HTML]{FFFFFF}{38.23} & \cellcolor[HTML]{7CC87C}\textcolor[HTML]{000000}{31.58} & \cellcolor[HTML]{0E7936}\textcolor[HTML]{FFFFFF}{39.17} & \cellcolor[HTML]{005B25}\textcolor[HTML]{FFFFFF}{54.30} & \cellcolor[HTML]{026F2E}\textcolor[HTML]{FFFFFF}{51.87} & \cellcolor[HTML]{1C8540}\textcolor[HTML]{FFFFFF}{46.88} & \cellcolor[HTML]{006B2B}\textcolor[HTML]{FFFFFF}{41.93} & \cellcolor[HTML]{9FD899}\textcolor[HTML]{000000}{29.00} & \cellcolor[HTML]{5AB769}\textcolor[HTML]{FFFFFF}{25.22} & \cellcolor[HTML]{80CA80}\textcolor[HTML]{000000}{25.00} & \cellcolor[HTML]{DC2924}\textcolor[HTML]{FFFFFF}{4.23} \\

\refstepcounter{rowcount}\arabic{rowcount}\label{row:q3vl8b_sim_prune_inter_cosine} & \bPrune\ \bInterCos & InfiniPot-V (TaR)~\cite{kim2025infinipotv} & \cellcolor[HTML]{005020}\textcolor[HTML]{FFFFFF}{45.23} & \cellcolor[HTML]{006027}\textcolor[HTML]{FFFFFF}{57.87} & \cellcolor[HTML]{067230}\textcolor[HTML]{FFFFFF}{53.78} & \cellcolor[HTML]{005421}\textcolor[HTML]{FFFFFF}{55.60} & \cellcolor[HTML]{005723}\textcolor[HTML]{FFFFFF}{45.76} & \cellcolor[HTML]{73C476}\textcolor[HTML]{000000}{28.05} & \cellcolor[HTML]{00471C}\textcolor[HTML]{FFFFFF}{39.73} & \cellcolor[HTML]{00441B}\textcolor[HTML]{FFFFFF}{35.86} & \cellcolor[HTML]{006529}\textcolor[HTML]{FFFFFF}{40.16} & \cellcolor[HTML]{005522}\textcolor[HTML]{FFFFFF}{54.75} & \cellcolor[HTML]{006328}\textcolor[HTML]{FFFFFF}{52.76} & \cellcolor[HTML]{05712F}\textcolor[HTML]{FFFFFF}{48.82} & \cellcolor[HTML]{00441B}\textcolor[HTML]{FFFFFF}{43.87} & \cellcolor[HTML]{067230}\textcolor[HTML]{FFFFFF}{30.84} & \cellcolor[HTML]{258D47}\textcolor[HTML]{FFFFFF}{26.38} & \cellcolor[HTML]{98D594}\textcolor[HTML]{000000}{23.68} & \cellcolor[HTML]{AB1016}\textcolor[HTML]{FFFFFF}{4.99} \\

\refstepcounter{rowcount}\arabic{rowcount}\label{row:q3vl8b_sim_prune_intra_vnorm_inter_cosine} & \bPrune\ \bIntraVN\ \bInterCos & InfiniPot-V~\cite{kim2025infinipotv} & \cellcolor[HTML]{005F26}\textcolor[HTML]{FFFFFF}{44.52} & \cellcolor[HTML]{005522}\textcolor[HTML]{FFFFFF}{58.91} & \cellcolor[HTML]{005221}\textcolor[HTML]{FFFFFF}{56.23} & \cellcolor[HTML]{005622}\textcolor[HTML]{FFFFFF}{55.51} & \cellcolor[HTML]{004A1E}\textcolor[HTML]{FFFFFF}{46.36} & \cellcolor[HTML]{F7FCF5}\textcolor[HTML]{000000}{26.10} & \cellcolor[HTML]{004A1E}\textcolor[HTML]{FFFFFF}{39.58} & \cellcolor[HTML]{DDF2D8}\textcolor[HTML]{000000}{28.95} & \cellcolor[HTML]{006328}\textcolor[HTML]{FFFFFF}{40.22} & \cellcolor[HTML]{005924}\textcolor[HTML]{FFFFFF}{54.44} & \cellcolor[HTML]{005321}\textcolor[HTML]{FFFFFF}{53.93} & \cellcolor[HTML]{006027}\textcolor[HTML]{FFFFFF}{50.12} & \cellcolor[HTML]{006428}\textcolor[HTML]{FFFFFF}{42.27} & \cellcolor[HTML]{268E47}\textcolor[HTML]{FFFFFF}{30.38} & \cellcolor[HTML]{F7FCF5}\textcolor[HTML]{000000}{21.45} & \cellcolor[HTML]{37A055}\textcolor[HTML]{FFFFFF}{28.95} & \cellcolor[HTML]{D11E1F}\textcolor[HTML]{FFFFFF}{4.42} \\

\refstepcounter{rowcount}\arabic{rowcount}\label{row:0_12_q3vl8b_sim_prune_intra_pseudo_att_inter_cosine_octer} & \bPrune\ \bIntraPA\ \bInterCos & StreamTOM~\cite{chen2025streamingtom}* & \cellcolor[HTML]{005020}\textcolor[HTML]{FFFFFF}{45.19} & \cellcolor[HTML]{006027}\textcolor[HTML]{FFFFFF}{57.87} & \cellcolor[HTML]{00441B}\textcolor[HTML]{FFFFFF}{57.28} & \cellcolor[HTML]{00471C}\textcolor[HTML]{FFFFFF}{56.68} & \cellcolor[HTML]{00471C}\textcolor[HTML]{FFFFFF}{46.58} & \cellcolor[HTML]{55B567}\textcolor[HTML]{FFFFFF}{28.35} & \cellcolor[HTML]{077331}\textcolor[HTML]{FFFFFF}{37.78} & \cellcolor[HTML]{72C375}\textcolor[HTML]{000000}{31.79} & \cellcolor[HTML]{005E26}\textcolor[HTML]{FFFFFF}{40.43} & \cellcolor[HTML]{004C1E}\textcolor[HTML]{FFFFFF}{55.42} & \cellcolor[HTML]{004D1F}\textcolor[HTML]{FFFFFF}{54.37} & \cellcolor[HTML]{005622}\textcolor[HTML]{FFFFFF}{50.88} & \cellcolor[HTML]{016E2D}\textcolor[HTML]{FFFFFF}{41.73} & \cellcolor[HTML]{006027}\textcolor[HTML]{FFFFFF}{31.07} & \cellcolor[HTML]{37A055}\textcolor[HTML]{FFFFFF}{25.88} & \cellcolor[HTML]{98D594}\textcolor[HTML]{000000}{23.68} & \cellcolor[HTML]{77040F}\textcolor[HTML]{FFFFFF}{5.51} \\

\refstepcounter{rowcount}\arabic{rowcount}\label{row:q3vl8b_sim_prune_intra_cosine} & \bPrune\ \bIntraCos & -- & \cellcolor[HTML]{00471C}\textcolor[HTML]{FFFFFF}{45.63} & \cellcolor[HTML]{00441B}\textcolor[HTML]{FFFFFF}{60.51} & \cellcolor[HTML]{16803C}\textcolor[HTML]{FFFFFF}{52.39} & \cellcolor[HTML]{00491D}\textcolor[HTML]{FFFFFF}{56.61} & \cellcolor[HTML]{00451C}\textcolor[HTML]{FFFFFF}{46.64} & \cellcolor[HTML]{45AD5F}\textcolor[HTML]{FFFFFF}{28.50} & \cellcolor[HTML]{00441B}\textcolor[HTML]{FFFFFF}{39.88} & \cellcolor[HTML]{006B2B}\textcolor[HTML]{FFFFFF}{34.87} & \cellcolor[HTML]{0B7734}\textcolor[HTML]{FFFFFF}{39.34} & \cellcolor[HTML]{00491D}\textcolor[HTML]{FFFFFF}{55.60} & \cellcolor[HTML]{00451C}\textcolor[HTML]{FFFFFF}{54.99} & \cellcolor[HTML]{00441B}\textcolor[HTML]{FFFFFF}{52.18} & \cellcolor[HTML]{0A7633}\textcolor[HTML]{FFFFFF}{41.27} & \cellcolor[HTML]{16803C}\textcolor[HTML]{FFFFFF}{30.61} & \cellcolor[HTML]{6BC072}\textcolor[HTML]{000000}{24.93} & \cellcolor[HTML]{F7FCF5}\textcolor[HTML]{000000}{15.79} & \cellcolor[HTML]{FCB69B}\textcolor[HTML]{000000}{2.28} \\

\refstepcounter{rowcount}\arabic{rowcount}\label{row:q3vl8b_sim_prune_intra_pseudo_att}\label{row:q3vl8b_sim_prune_intra_pseudo_att_aaaaa} & \bPrune\ \bIntraPA & -- & \cellcolor[HTML]{005220}\textcolor[HTML]{FFFFFF}{45.12} & \cellcolor[HTML]{004E1F}\textcolor[HTML]{FFFFFF}{59.47} & \cellcolor[HTML]{0E7936}\textcolor[HTML]{FFFFFF}{53.08} & \cellcolor[HTML]{004A1D}\textcolor[HTML]{FFFFFF}{56.52} & \cellcolor[HTML]{005522}\textcolor[HTML]{FFFFFF}{45.87} & \cellcolor[HTML]{359E53}\textcolor[HTML]{FFFFFF}{28.72} & \cellcolor[HTML]{005221}\textcolor[HTML]{FFFFFF}{39.28} & \cellcolor[HTML]{3EA75A}\textcolor[HTML]{FFFFFF}{32.89} & \cellcolor[HTML]{00441B}\textcolor[HTML]{FFFFFF}{41.50} & \cellcolor[HTML]{00441B}\textcolor[HTML]{FFFFFF}{56.04} & \cellcolor[HTML]{00491D}\textcolor[HTML]{FFFFFF}{54.71} & \cellcolor[HTML]{005F26}\textcolor[HTML]{FFFFFF}{50.18} & \cellcolor[HTML]{16803C}\textcolor[HTML]{FFFFFF}{40.60} & \cellcolor[HTML]{067230}\textcolor[HTML]{FFFFFF}{30.84} & \cellcolor[HTML]{DAF0D4}\textcolor[HTML]{000000}{22.61} & \cellcolor[HTML]{00441B}\textcolor[HTML]{FFFFFF}{35.53} & \cellcolor[HTML]{FC8F6F}\textcolor[HTML]{000000}{2.82} \\

\refstepcounter{rowcount}\arabic{rowcount}\label{row:q3vl8b_sim_prune_inter_pseudo_attn} & \bPrune\ \bInterPA & -- & \cellcolor[HTML]{004D1E}\textcolor[HTML]{FFFFFF}{45.42} & \cellcolor[HTML]{004A1E}\textcolor[HTML]{FFFFFF}{59.87} & \cellcolor[HTML]{005B25}\textcolor[HTML]{FFFFFF}{55.53} & \cellcolor[HTML]{004E1F}\textcolor[HTML]{FFFFFF}{56.17} & \cellcolor[HTML]{005924}\textcolor[HTML]{FFFFFF}{45.65} & \cellcolor[HTML]{B6E2AF}\textcolor[HTML]{000000}{27.30} & \cellcolor[HTML]{006328}\textcolor[HTML]{FFFFFF}{38.53} & \cellcolor[HTML]{006B2B}\textcolor[HTML]{FFFFFF}{34.87} & \cellcolor[HTML]{006227}\textcolor[HTML]{FFFFFF}{40.30} & \cellcolor[HTML]{004C1E}\textcolor[HTML]{FFFFFF}{55.42} & \cellcolor[HTML]{004E1F}\textcolor[HTML]{FFFFFF}{54.32} & \cellcolor[HTML]{006B2B}\textcolor[HTML]{FFFFFF}{49.35} & \cellcolor[HTML]{005924}\textcolor[HTML]{FFFFFF}{42.80} & \cellcolor[HTML]{067230}\textcolor[HTML]{FFFFFF}{30.84} & \cellcolor[HTML]{8BCF89}\textcolor[HTML]{000000}{24.35} & \cellcolor[HTML]{80CA80}\textcolor[HTML]{000000}{25.00} & \cellcolor[HTML]{B61319}\textcolor[HTML]{FFFFFF}{4.82} \\

\refstepcounter{rowcount}\arabic{rowcount}\label{row:q3vl8b_sim_prune_intra_pseudo_att_inter_pseudo_att} & \bPrune\ \bIntraPA\ \bInterPA & -- & \cellcolor[HTML]{005B24}\textcolor[HTML]{FFFFFF}{44.71} & \cellcolor[HTML]{005A24}\textcolor[HTML]{FFFFFF}{58.51} & \cellcolor[HTML]{005F26}\textcolor[HTML]{FFFFFF}{55.30} & \cellcolor[HTML]{004A1D}\textcolor[HTML]{FFFFFF}{56.52} & \cellcolor[HTML]{005B25}\textcolor[HTML]{FFFFFF}{45.54} & \cellcolor[HTML]{3FA85B}\textcolor[HTML]{FFFFFF}{28.57} & \cellcolor[HTML]{004A1E}\textcolor[HTML]{FFFFFF}{39.58} & \cellcolor[HTML]{DDF2D8}\textcolor[HTML]{000000}{28.95} & \cellcolor[HTML]{00682A}\textcolor[HTML]{FFFFFF}{40.03} & \cellcolor[HTML]{004A1E}\textcolor[HTML]{FFFFFF}{55.55} & \cellcolor[HTML]{005522}\textcolor[HTML]{FFFFFF}{53.87} & \cellcolor[HTML]{005E26}\textcolor[HTML]{FFFFFF}{50.29} & \cellcolor[HTML]{00692A}\textcolor[HTML]{FFFFFF}{42.00} & \cellcolor[HTML]{75C477}\textcolor[HTML]{000000}{29.46} & \cellcolor[HTML]{19833E}\textcolor[HTML]{FFFFFF}{26.67} & \cellcolor[HTML]{AEDEA7}\textcolor[HTML]{000000}{22.37} & \cellcolor[HTML]{67000D}\textcolor[HTML]{FFFFFF}{5.66} \\

\refstepcounter{rowcount}\arabic{rowcount}\label{row:q3vl8b_sim_prune_intra_vnorm_inter_pseudo_att} & \bPrune\ \bIntraVN\ \bInterPA & -- & \cellcolor[HTML]{006026}\textcolor[HTML]{FFFFFF}{44.42} & \cellcolor[HTML]{006328}\textcolor[HTML]{FFFFFF}{57.62} & \cellcolor[HTML]{004E1F}\textcolor[HTML]{FFFFFF}{56.46} & \cellcolor[HTML]{004E1F}\textcolor[HTML]{FFFFFF}{56.21} & \cellcolor[HTML]{00441B}\textcolor[HTML]{FFFFFF}{46.75} & \cellcolor[HTML]{D5EFCF}\textcolor[HTML]{000000}{26.85} & \cellcolor[HTML]{004E1F}\textcolor[HTML]{FFFFFF}{39.43} & \cellcolor[HTML]{F7FCF5}\textcolor[HTML]{000000}{27.63} & \cellcolor[HTML]{006B2B}\textcolor[HTML]{FFFFFF}{39.91} & \cellcolor[HTML]{005020}\textcolor[HTML]{FFFFFF}{55.10} & \cellcolor[HTML]{005221}\textcolor[HTML]{FFFFFF}{54.04} & \cellcolor[HTML]{006227}\textcolor[HTML]{FFFFFF}{50.06} & \cellcolor[HTML]{006729}\textcolor[HTML]{FFFFFF}{42.13} & \cellcolor[HTML]{349D53}\textcolor[HTML]{FFFFFF}{30.15} & \cellcolor[HTML]{CFECC9}\textcolor[HTML]{000000}{22.90} & \cellcolor[HTML]{80CA80}\textcolor[HTML]{000000}{25.00} & \cellcolor[HTML]{C1161B}\textcolor[HTML]{FFFFFF}{4.67} \\

\refstepcounter{rowcount}\arabic{rowcount}\label{row:q3vl8b_sim_prune_intra_cross_sink_inter_pseudo_att} & \bPrune\ \bIntraCS\ \bInterPA & -- & \cellcolor[HTML]{006127}\textcolor[HTML]{FFFFFF}{44.39} & \cellcolor[HTML]{006D2C}\textcolor[HTML]{FFFFFF}{56.74} & \cellcolor[HTML]{157F3B}\textcolor[HTML]{FFFFFF}{52.50} & \cellcolor[HTML]{005622}\textcolor[HTML]{FFFFFF}{55.56} & \cellcolor[HTML]{006227}\textcolor[HTML]{FFFFFF}{45.22} & \cellcolor[HTML]{278F48}\textcolor[HTML]{FFFFFF}{28.95} & \cellcolor[HTML]{00441B}\textcolor[HTML]{FFFFFF}{39.88} & \cellcolor[HTML]{6BC072}\textcolor[HTML]{000000}{31.91} & \cellcolor[HTML]{05712F}\textcolor[HTML]{FFFFFF}{39.60} & \cellcolor[HTML]{004E1F}\textcolor[HTML]{FFFFFF}{55.24} & \cellcolor[HTML]{004D1F}\textcolor[HTML]{FFFFFF}{54.43} & \cellcolor[HTML]{0D7836}\textcolor[HTML]{FFFFFF}{48.12} & \cellcolor[HTML]{16803C}\textcolor[HTML]{FFFFFF}{40.60} & \cellcolor[HTML]{3CA559}\textcolor[HTML]{FFFFFF}{30.03} & \cellcolor[HTML]{A9DCA3}\textcolor[HTML]{000000}{23.77} & \cellcolor[HTML]{80CA80}\textcolor[HTML]{000000}{25.00} & \cellcolor[HTML]{B31218}\textcolor[HTML]{FFFFFF}{4.87} \\

\midrule
\multicolumn{20}{@{}l}{\textit{Similarity-based prune \bFIFO\ \bSink +}} \\

\refstepcounter{rowcount}\arabic{rowcount}\label{row:q3vl8b_sim_prune_intra_pseudo_att_inter_cosine_no_hat} & \bPrune\ \bIntraPA\ \bInterCos & StreamTOM~\cite{chen2025streamingtom} * & \cellcolor[HTML]{006127}\textcolor[HTML]{FFFFFF}{44.38} & \cellcolor[HTML]{005723}\textcolor[HTML]{FFFFFF}{58.75} & \cellcolor[HTML]{006D2C}\textcolor[HTML]{FFFFFF}{54.25} & \cellcolor[HTML]{004B1E}\textcolor[HTML]{FFFFFF}{56.39} & \cellcolor[HTML]{005120}\textcolor[HTML]{FFFFFF}{46.04} & \cellcolor[HTML]{F7FCF5}\textcolor[HTML]{000000}{24.98} & \cellcolor[HTML]{005924}\textcolor[HTML]{FFFFFF}{38.98} & \cellcolor[HTML]{8ACE88}\textcolor[HTML]{000000}{31.25} & \cellcolor[HTML]{006328}\textcolor[HTML]{FFFFFF}{40.23} & \cellcolor[HTML]{004E1F}\textcolor[HTML]{FFFFFF}{55.28} & \cellcolor[HTML]{005F26}\textcolor[HTML]{FFFFFF}{53.09} & \cellcolor[HTML]{067230}\textcolor[HTML]{FFFFFF}{48.76} & \cellcolor[HTML]{006428}\textcolor[HTML]{FFFFFF}{42.27} & \cellcolor[HTML]{005723}\textcolor[HTML]{FFFFFF}{31.19} & \cellcolor[HTML]{F7FCF5}\textcolor[HTML]{000000}{19.42} & \cellcolor[HTML]{157F3B}\textcolor[HTML]{FFFFFF}{31.58} & \cellcolor[HTML]{B91419}\textcolor[HTML]{FFFFFF}{4.76} \\

\midrule
\multicolumn{20}{@{}l}{\textit{Attention-based prune with adaptive head budgets \bFIFO\ \bSink\ \bHat +}} \\

\refstepcounter{rowcount}\rev{\arabic{rowcount}}\label{row:q3vl8b_adakv} & \rev{\bPrune\ \bAdaHead} & Ada-KV~\cite{feng2025adakv} & \cellcolor[HTML]{00441B}\textcolor[HTML]{FFFFFF}{45.88} & \cellcolor[HTML]{005421}\textcolor[HTML]{FFFFFF}{58.91} & \cellcolor[HTML]{006026}\textcolor[HTML]{FFFFFF}{55.18} & \cellcolor[HTML]{00441B}\textcolor[HTML]{FFFFFF}{57.04} & \cellcolor[HTML]{005B24}\textcolor[HTML]{FFFFFF}{45.54} & \cellcolor[HTML]{0E7936}\textcolor[HTML]{FFFFFF}{29.09} & \cellcolor[HTML]{00441B}\textcolor[HTML]{FFFFFF}{40.18} & \cellcolor[HTML]{005D25}\textcolor[HTML]{FFFFFF}{35.20} & \cellcolor[HTML]{005220}\textcolor[HTML]{FFFFFF}{40.89} & \cellcolor[HTML]{00451B}\textcolor[HTML]{FFFFFF}{55.91} & \cellcolor[HTML]{00441B}\textcolor[HTML]{FFFFFF}{55.82} & \cellcolor[HTML]{005823}\textcolor[HTML]{FFFFFF}{50.65} & \cellcolor[HTML]{006829}\textcolor[HTML]{FFFFFF}{42.07} & \cellcolor[HTML]{43AC5E}\textcolor[HTML]{FFFFFF}{29.92} & \cellcolor[HTML]{98D493}\textcolor[HTML]{000000}{22.90} & \cellcolor[HTML]{37A055}\textcolor[HTML]{FFFFFF}{28.95} & \cellcolor[HTML]{FB8363}\textcolor[HTML]{000000}{2.99} \\

\midrule
\multicolumn{20}{@{}l}{\textit{Offloading-based memory management \bFIFO \bSink\ \bHat}} \\

\refstepcounter{rowcount}\arabic{rowcount}\label{row:q3vl8b_rekv_offloading} & \bOff & ReKV~\cite{di2025rekv} & \cellcolor[HTML]{2E974E}\textcolor[HTML]{FFFFFF}{40.99} & \cellcolor[HTML]{2E964D}\textcolor[HTML]{FFFFFF}{51.69} & \cellcolor[HTML]{369F54}\textcolor[HTML]{FFFFFF}{49.48} & \cellcolor[HTML]{30994F}\textcolor[HTML]{FFFFFF}{48.99} & \cellcolor[HTML]{1E8741}\textcolor[HTML]{FFFFFF}{42.86} & \cellcolor[HTML]{00441B}\textcolor[HTML]{FFFFFF}{29.99} & \cellcolor[HTML]{2A924A}\textcolor[HTML]{FFFFFF}{35.98} & \cellcolor[HTML]{F1FAEE}\textcolor[HTML]{000000}{27.96} & \cellcolor[HTML]{3CA559}\textcolor[HTML]{FFFFFF}{36.83} & \cellcolor[HTML]{137D39}\textcolor[HTML]{FFFFFF}{51.36} & \cellcolor[HTML]{2F974E}\textcolor[HTML]{FFFFFF}{47.97} & \cellcolor[HTML]{3BA458}\textcolor[HTML]{FFFFFF}{44.00} & \cellcolor[HTML]{349D53}\textcolor[HTML]{FFFFFF}{38.67} & \cellcolor[HTML]{8ACE88}\textcolor[HTML]{000000}{29.23} & \cellcolor[HTML]{48AE60}\textcolor[HTML]{FFFFFF}{25.51} & \cellcolor[HTML]{C2E7BB}\textcolor[HTML]{000000}{21.05} & \cellcolor[HTML]{FDC6B0}\textcolor[HTML]{000000}{2.03} \\

\refstepcounter{rowcount}\arabic{rowcount}\label{row:q3vl8b_rekv_offloading_8bit} & \bQOff & ReKV~\cite{di2025rekv} w/ \bQOff & \cellcolor[HTML]{258D46}\textcolor[HTML]{FFFFFF}{41.63} & \cellcolor[HTML]{2C944C}\textcolor[HTML]{FFFFFF}{51.85} & \cellcolor[HTML]{2A924A}\textcolor[HTML]{FFFFFF}{50.64} & \cellcolor[HTML]{1E8742}\textcolor[HTML]{FFFFFF}{50.83} & \cellcolor[HTML]{117B38}\textcolor[HTML]{FFFFFF}{43.63} & \cellcolor[HTML]{A4DA9E}\textcolor[HTML]{000000}{27.52} & \cellcolor[HTML]{349D53}\textcolor[HTML]{FFFFFF}{35.38} & \cellcolor[HTML]{7CC87C}\textcolor[HTML]{000000}{31.58} & \cellcolor[HTML]{17813D}\textcolor[HTML]{FFFFFF}{38.77} & \cellcolor[HTML]{05712F}\textcolor[HTML]{FFFFFF}{52.61} & \cellcolor[HTML]{309950}\textcolor[HTML]{FFFFFF}{47.74} & \cellcolor[HTML]{45AD5F}\textcolor[HTML]{FFFFFF}{43.12} & \cellcolor[HTML]{2C944C}\textcolor[HTML]{FFFFFF}{39.27} & \cellcolor[HTML]{0E7936}\textcolor[HTML]{FFFFFF}{30.72} & \cellcolor[HTML]{258D47}\textcolor[HTML]{FFFFFF}{26.38} & \cellcolor[HTML]{157F3B}\textcolor[HTML]{FFFFFF}{31.58} & \cellcolor[HTML]{FEE2D5}\textcolor[HTML]{000000}{1.59} \\

\refstepcounter{rowcount}\arabic{rowcount}\label{row:streamtom_like} & \multicolumn{2}{l}{\bPrune\ \bIntraPA\ \bInterCos\ \bQOff  \hspace{2mm} StreamingTOM~\cite{chen2025streamingtom}} & \cellcolor[HTML]{107B37}\textcolor[HTML]{FFFFFF}{42.86} & \cellcolor[HTML]{1D8640}\textcolor[HTML]{FFFFFF}{53.61} & \cellcolor[HTML]{16803C}\textcolor[HTML]{FFFFFF}{52.39} & \cellcolor[HTML]{137E3A}\textcolor[HTML]{FFFFFF}{51.88} & \cellcolor[HTML]{026F2E}\textcolor[HTML]{FFFFFF}{44.51} & \cellcolor[HTML]{7AC77B}\textcolor[HTML]{000000}{27.97} & \cellcolor[HTML]{077331}\textcolor[HTML]{FFFFFF}{37.78} & \cellcolor[HTML]{6BC072}\textcolor[HTML]{000000}{31.91} & \cellcolor[HTML]{006B2B}\textcolor[HTML]{FFFFFF}{39.93} & \cellcolor[HTML]{005F26}\textcolor[HTML]{FFFFFF}{54.03} & \cellcolor[HTML]{278F48}\textcolor[HTML]{FFFFFF}{48.69} & \cellcolor[HTML]{2F984F}\textcolor[HTML]{FFFFFF}{45.06} & \cellcolor[HTML]{208843}\textcolor[HTML]{FFFFFF}{40.00} & \cellcolor[HTML]{00441B}\textcolor[HTML]{FFFFFF}{31.42} & \cellcolor[HTML]{2F984F}\textcolor[HTML]{FFFFFF}{26.09} & \cellcolor[HTML]{005A24}\textcolor[HTML]{FFFFFF}{34.21} & \cellcolor[HTML]{FEE7DC}\textcolor[HTML]{000000}{1.43} \\

\bottomrule

\end{tabular}%
}

{
\vspace{1pt}
\footnotesize
% \textbf{Badge legend.}
% \bSW\ Sliding Window;
% \bSink\ Sink tokens;
% \bHat\ Hat tokens;
% \bMerge\ Similarity merging;
% \bPrune\ Similarity pruning;
% \bIntraPA\ Intra-frame pseudo-attention;
% \bInterPA\ Inter-frame pseudo-attention;
% \bIntraCos\ Intra-frame cosine;
% \bInterCos\ Inter-frame cosine;
% \bIntraCS\ Intra-frame cross-sink;
% \bIntraVN\ Intra-frame value norm;
% \bOff\ disk offloading;
% \bQOff\ quantized offloading.
Memory budget: 9,804 tokens ($\approx 1.35GB$).
*: variant without \bQOff.
}
\end{table}

\subsection{Controlled Diagnostic Evaluation}
\label{sec:controlled_experiments}

Table~\ref{tab:controlled_eval} shows that while aggregate performance gaps among memory strategies are narrow, critical variations exist in \emph{what} is retained and for \emph{how long}. 
Established methods like StreamingVLM~\cite{xu2025streamingvlm} remain highly competitive, performing on par with recent approaches like InfiniPot-V~\cite{kim2025infinipotv}. 
%The text-only baseline remains close to chance, confirming that the benchmark cannot be solved reliably from linguistic priors alone and requires access to visual episodic evidence. 
No single method strictly dominates, showing that some are locally optimal for spatial or temporal retention (Rows~\ref{row:q3vl8b_sim_prune_intra_cosine}, \ref{row:q3vl8b_sim_prune_intra_pseudo_att}), while others (Rows~\ref{row:q3vl8b_sw_sink}, \ref{row:q3vl8b_sim_prune_intra_cosine}) balance accuracy and processing time. These trade-offs reveal key trends that can guide the design of future streaming architectures.

\textbf{The role of \bSink and \bHat tokens.} The naive Sliding Window (\bFIFO, Row~\ref{row:q3vl8b_sim_sw_pura}) yields the lowest accuracy. Adding \bSink tokens (Row~\ref{row:q3vl8b_sw_sink_aaaaa}) boosts semantic and temporal averages from 29.53\%/28.33\% to 44.74\%/39.05\%, with negligible latency change (1.05s to 1.07s). This confirms that preserving system prompts is essential for stable inference~\cite{xiao2024efficient}. Including \bHat tokens adds a small benefit under the same policy, increasing semantic/temporal accuracy from 44.38\%/40.23\% to 45.19\%/40.43\% (Rows~\ref{row:q3vl8b_sim_prune_intra_pseudo_att_inter_cosine_no_hat} and~\ref{row:0_12_q3vl8b_sim_prune_intra_pseudo_att_inter_cosine_octer}). These results show the value of protecting structural context.

\textbf{Merge vs. Prune.} \bPrune\ mechanisms consistently outperform token \bMerge. For \bIntraPA, pruning (Row~\ref{row:q3vl8b_sim_prune_intra_pseudo_att_aaaaa}) beats merging (Row~\ref{row:q3vl8b_sim_merge_intra_pseudo_att}) 45.12\% to 41.87\%. Discarding redundant tokens preserves episodic details better than averaging, which blurs fine-grained evidence (see DETL). While using a hybrid \bPrune/\bMerge\ approach (Row~\ref{row:q3vl8b_sim_merge_intra_pseudo_att_inter_pseudo_att}) partially closes this gap (43.44\%), it drastically increases processing time (5.32s vs. $\sim$2.5s for pure intra-merging).

% \textbf{Optimal in-memory strategies.} Combining intra- and inter-frame reduction yields the best streaming policies. The hybrid \bPrune\ \bIntraPA\ \bInterCos (Row~\ref{row:0_12_q3vl8b_sim_prune_intra_pseudo_att_inter_cosine_octer}) provides the strongest overall trade-off between semantic and temporal recall (45.19\%/40.43\%), excelling in Spatial and Detail memory. Blending local salience filtering with cross-frame redundancy checks optimally preserves visual anchors. However, this peak accuracy incurs the highest processing time (5.51s), more expensive than variants like \bPrune\ \bIntraCos (Row~\ref{row:q3vl8b_sim_prune_intra_cosine}, 2.28s).

\textbf{Offloading.} \bOff\ keeps evicted memory on disk, but alone is weaker than in-memory pruning. Row~\ref{row:q3vl8b_rekv_offloading} reaches 40.99\% / 36.83\% with 2.0966 GB$\times$min. \bQOff\ halves storage to 1.0515 GB$\times$min and improves accuracy (Row~\ref{row:q3vl8b_rekv_offloading_8bit}: 41.63\% / 38.77\%, 1.59s), while StreamingTOM (Row~\ref{row:streamtom_like}) obtains better results combining low-salience tokens \bPrune+\bQOff\ for 42.86\% / 39.93\%, gaining 34.21\% for Ultra-Long recall and only 0.4847 GB$\times$min.

\textbf{Adaptive head budgets.} \bAdaHead (Row~\ref{row:q3vl8b_adakv}) attains the highest semantic average in Table~\ref{tab:controlled_eval} (45.88\%) and a temporal average of 40.89\%, surpassing the intra-/inter-frame pruning hybrid of Row~\ref{row:0_12_q3vl8b_sim_prune_intra_pseudo_att_inter_cosine_octer} on both averages at about half its processing time (2.99s vs. 5.51s), with the best Detail (57.04\%), Causal (40.18\%) and Short-term recall (55.82\%) in the table. Its Ultra-Long recall (28.95\%) improves over \bPrune\ \bIntraCos\ (15.79\%) but stays below the frame-aware \bPrune\ \bIntraPA\ (35.53\%). Redistributing the budget across heads complements, rather than replaces, temporal coverage.

\textbf{Optimal in-memory strategies.} No in-memory policy dominates, in Table~\ref{tab:controlled_eval} \bPrune\ \bIntraCos\ (Row~\ref{row:q3vl8b_sim_prune_intra_cosine}) gives the best Spatial retention (60.51\%) and accuracy--latency trade-off (45.63\% at 2.28s), \bPrune\ \bIntraPA\ (Row~\ref{row:q3vl8b_sim_prune_intra_pseudo_att}) the best temporal average (41.50\%) and Ultra-Long recall (35.53\%). Inter-frame reduction (Row~\ref{row:0_12_q3vl8b_sim_prune_intra_pseudo_att_inter_cosine_octer}) adds the best Temporal retention (57.28\%) at twice the cost (5.51s); adaptive head budgets (Row~\ref{row:q3vl8b_adakv}) reach the highest semantic average (45.88\%) and Detail retention (57.04\%) but lag on long horizons; quantized offloading (Row~\ref{row:streamtom_like}) nearly matches the best Ultra-Long recall (34.21\%) at the lowest latency among reduction policies (1.43s). Across Rows~\ref{row:0_12_q3vl8b_sim_prune_intra_pseudo_att_inter_cosine_octer}--\ref{row:q3vl8b_sim_prune_intra_pseudo_att} and~\ref{row:q3vl8b_adakv} the semantic average varies by less than one point (45.12\%--45.88\%) while Ultra-Long recall spans 15.79\%--35.53\%: aggregate accuracy says little about what is retained and for how long, which the per-category and per-regime breakdown of \egostream exposes.

\subsection{State-of-the-Art Streaming Models}
\label{sec:sota_methods}

When comparing state-of-the-art streaming architectures under our unified KV-cache framework, distinct accuracy--latency trade-offs emerge rather than a strict ranking (Table~\ref{tab:controlled_eval}). \textbf{StreamingVLM}~\cite{xu2025streamingvlm} serves as the strongest low-latency baseline, proving that protecting \bSink\ tokens yields high semantic (44.74\%) and temporal (39.05\%) accuracy with minimal overhead (1.07s). For tasks requiring higher fidelity, token pruning models like \textbf{StreamMem}~\cite{yang2025streammem}, \textbf{InfiniPot-V}~\cite{kim2025infinipotv}, and \textbf{StreamingTOM}~\cite{chen2025streamingtom} (without offloading) set the upper limits among similarity-based policies---peaking at 45.23\% semantic and 40.43\% temporal accuracy---but suffer from severe processing delays (4.23s--5.51s), whereas adaptive
head budgeting (Ada-KV~\cite{feng2025adakv}, Row~\ref{row:q3vl8b_adakv}) raises the ceiling to 45.88\%/40.89\% at 2.99s. Token merging (\textbf{CaM-like}~\cite{zhang2024cam}) offers a middle ground, cutting pruning latency in half (~2.33s--2.87s) while remaining competitive. Finally, external retrieval strategies like \textbf{ReKV}~\cite{di2025rekv} and full \textbf{StreamingTOM}~\cite{chen2025streamingtom} reveal that quantized offloading is beneficial for long-horizon memory; it rescues ultra-long recall (reaching 34.21\% for StreamingTOM) and slashes processing time (down to 1.43s), albeit with a slight dip in average accuracy. Ultimately, while current SOTA in-cache methods successfully specialize in different memory profiles, all still operate well below 1 FPS, highlighting that achieving true real-time streaming ingestion with in-cache management alone remains an open challenge.
Appendix~\ref{app:compression_baselines} also reports two complementary results using pre-prefill token compression (STC~\cite{wang2026stc}) reaching 47.18\% semantic accuracy at 0.23s per frame, and the off-the-shelf streaming model LiveCC~\cite{chen2025livecc} that streams about 0.34s per frame and retains more at long horizons, but is less accurate on instant, spatial and detail recall.

\begin{figure}[t]
    \centering
    % \includegraphics[width=0.9\linewidth]{figures/category_degradation_curves.pdf}
    %\fbox{\rule{0pt}{2in} \rule{0.9\linewidth}{0pt}} % Placeholder box
    
        \includegraphics[width=0.49\linewidth]{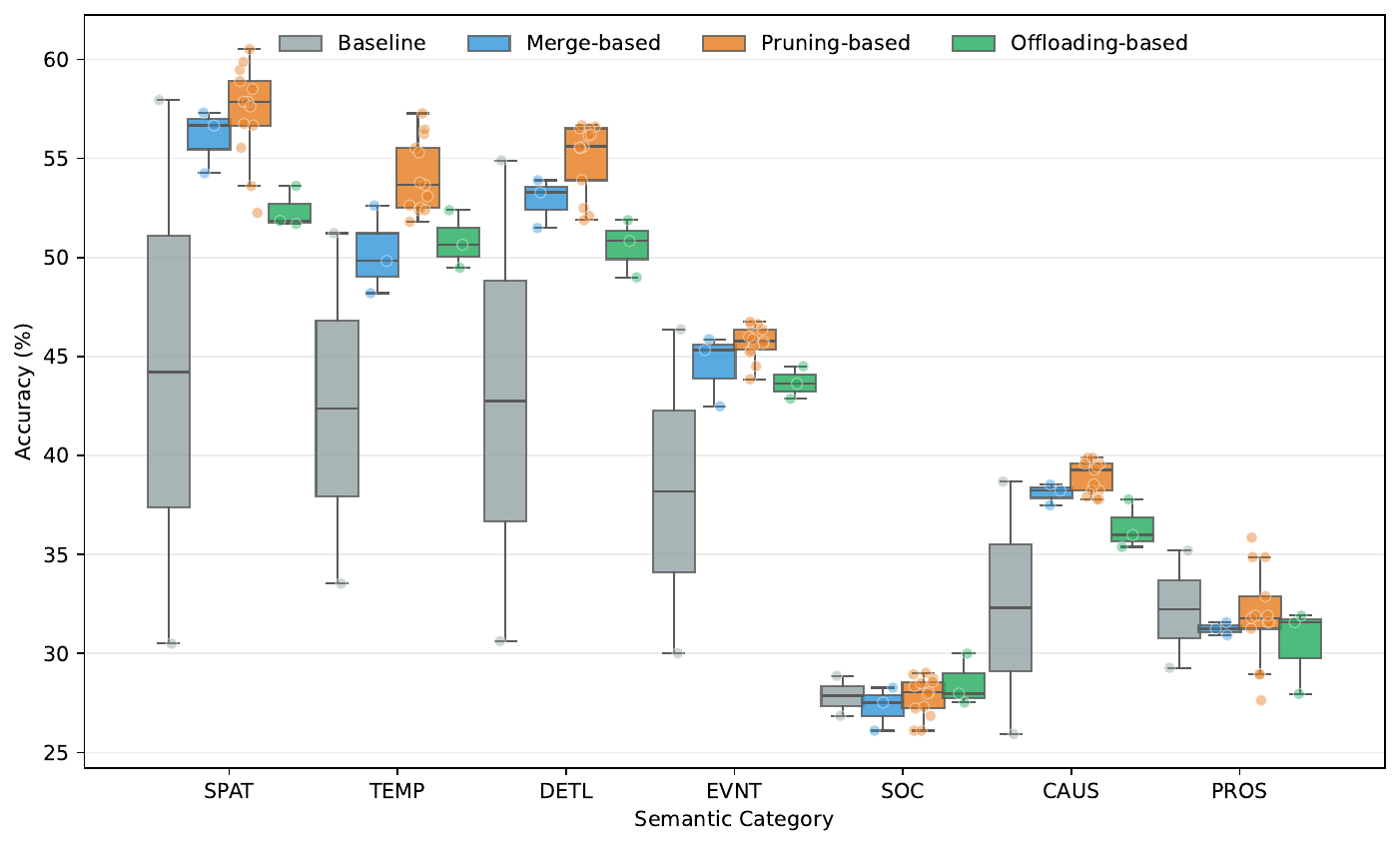}
        \includegraphics[width=0.49\linewidth]{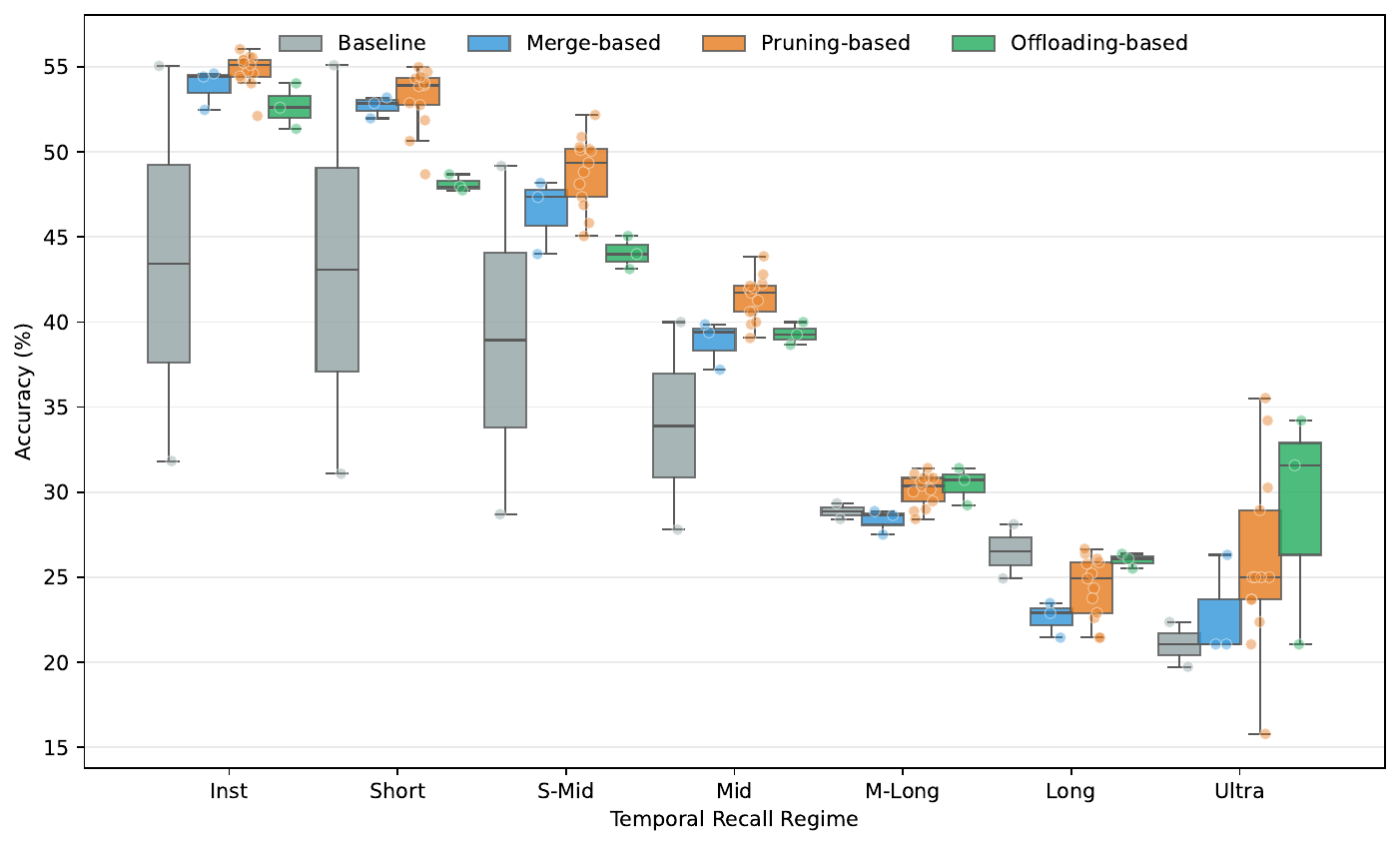}
    \caption{\textbf{Semantic and Temporal profiles.} Accuracy distributions of baselines (no \bPrune/\bMerge), merge-, prune-, and offloading-based methods wrt semantic categories (left) and recall times (right).}
    \label{fig:semantic_time_boxplots}
\end{figure}

\subsection{Semantic Profiles and Temporal Decay}
Figure~\ref{fig:semantic_time_boxplots} shows that merging, pruning and offloading techniques have different effects across semantic categories and temporal recall regimes. While all improve over baselines, pruning-based methods often have advantages, especially in the temporal and detail categories, and offloading-based approaches achieve better results on social and in mid-long to ultra-long recall. Overall, these results highlight the importance of evaluating across semantic categories and recall times, rather than just by aggregate accuracies.

\subsection{Memory Budget}
Table~\ref{tab:budgets} shows that memory budget is a driver of accuracy, but with diminishing returns. Increasing the budget from 2.6K to 9.8K tokens, from Row~\ref{bg:id_1} to Row~\ref{bg:id_4}, raises semantic accuracy from 40.88\% to 45.19\% and temporal accuracy from 35.59\% to 40.43\%. Larger budgets further improve the averages, peaking at 47.08\% semantic and 41.94\% temporal with 50.8K tokens in Row~\ref{bg:id_7}, but nearly triple processing time relative to the smallest budget in Row~\ref{bg:id_1}.

\subsection{Generalization Across Backbones}
\label{sec:backbones}
To verify that our conclusions are not specific to Qwen3-VL, we instantiate the framework on InternVL~3.5 (8B)~\cite{wang2025internvl35} and Molmo~2 (8B)~\cite{molmo2} with the same budget, protocol, prompts and policies. Table~\ref{tab:backbones} shows that the ranking is preserved across backbones: pruning beats merging on every backbone, on the semantic average (45.63 vs. 43.42, 43.09 vs. 41.92 and 41.61 vs. 39.57 for \bIntraCos) as well as on temporal recall, and keeps its advantage on Spatial and Detail memory, confirming that token averaging blurs fine-grained anchors regardless of the decoder.

\begin{table}[t]
\centering
\caption{Policy \bFIFO\ \bSink\ \bHat\ \bPrune\ \bIntraPA\ \bInterCos under different memory budgets (tokens).}
\label{tab:budgets}
\ifcsname c@rowcount\endcsname\else\newcounter{rowcount}\fi
\setcounter{rowcount}{0}
\renewcommand{\arraystretch}{1.2}
\resizebox{\textwidth}{!}{%
\begin{tabular}{@{}r r c c c c c c c c c c c c c c c c c@{}}
\toprule
 &  & \multicolumn{8}{c}{\textbf{Semantic Categories}} & \multicolumn{8}{c}{\textbf{Temporal Recall Regimes}} & \multicolumn{1}{c}{\textbf{Time (s)}} \\
\cmidrule(lr){3-10}
\cmidrule(lr){11-18}
\cmidrule(l){19-19}
 & \textbf{Tokens} & \textbf{Avg} & \textbf{SPAT} & \textbf{TEMP} & \textbf{DETL} & \textbf{EVNT} & \textbf{SOC} & \textbf{CAUS} & \textbf{PROS} & \textbf{Avg} & \textbf{Inst} & \textbf{Short} & \textbf{S-Mid} & \textbf{Mid} & \textbf{ML} & \textbf{Long} & \textbf{Ultra} & \textbf{Proc.} \\
\midrule
\refstepcounter{rowcount}\arabic{rowcount}\label{bg:id_1} & 2636 & \cellcolor[HTML]{F7FCF5}\textcolor[HTML]{000000}{40.88} & \cellcolor[HTML]{F7FCF5}\textcolor[HTML]{000000}{51.52} & \cellcolor[HTML]{F7FCF5}\textcolor[HTML]{000000}{49.83} & \cellcolor[HTML]{F7FCF5}\textcolor[HTML]{000000}{51.01} & \cellcolor[HTML]{F7FCF5}\textcolor[HTML]{000000}{42.37} & \cellcolor[HTML]{16803C}\textcolor[HTML]{FFFFFF}{27.75} & \cellcolor[HTML]{F0F9ED}\textcolor[HTML]{000000}{35.08} & \cellcolor[HTML]{CBEBC5}\textcolor[HTML]{000000}{28.62} & \cellcolor[HTML]{F7FCF5}\textcolor[HTML]{000000}{35.59} & \cellcolor[HTML]{F7FCF5}\textcolor[HTML]{000000}{51.76} & \cellcolor[HTML]{F7FCF5}\textcolor[HTML]{000000}{49.36} & \cellcolor[HTML]{F7FCF5}\textcolor[HTML]{000000}{43.35} & \cellcolor[HTML]{F7FCF5}\textcolor[HTML]{000000}{37.53} & \cellcolor[HTML]{BBE4B4}\textcolor[HTML]{000000}{31.19} & \cellcolor[HTML]{F7FCF5}\textcolor[HTML]{000000}{21.45} & \cellcolor[HTML]{F7FCF5}\textcolor[HTML]{000000}{14.47} & \cellcolor[HTML]{FFF5F0}\textcolor[HTML]{000000}{2.21} \\

\refstepcounter{rowcount}\arabic{rowcount}\label{bg:id_2} & 3660 & \cellcolor[HTML]{C8E9C1}\textcolor[HTML]{000000}{42.43} & \cellcolor[HTML]{BCE4B5}\textcolor[HTML]{000000}{54.65} & \cellcolor[HTML]{E1F3DC}\textcolor[HTML]{000000}{50.99} & \cellcolor[HTML]{CDECC7}\textcolor[HTML]{000000}{52.84} & \cellcolor[HTML]{B0DFAA}\textcolor[HTML]{000000}{44.07} & \cellcolor[HTML]{70C274}\textcolor[HTML]{000000}{26.93} & \cellcolor[HTML]{F7FCF5}\textcolor[HTML]{000000}{34.63} & \cellcolor[HTML]{097532}\textcolor[HTML]{FFFFFF}{32.89} & \cellcolor[HTML]{72C375}\textcolor[HTML]{000000}{38.80} & \cellcolor[HTML]{EBF7E7}\textcolor[HTML]{000000}{52.25} & \cellcolor[HTML]{C4E8BD}\textcolor[HTML]{000000}{50.97} & \cellcolor[HTML]{D5EFCF}\textcolor[HTML]{000000}{45.06} & \cellcolor[HTML]{BEE5B8}\textcolor[HTML]{000000}{39.47} & \cellcolor[HTML]{A5DB9F}\textcolor[HTML]{000000}{31.30} & \cellcolor[HTML]{19833E}\textcolor[HTML]{FFFFFF}{24.93} & \cellcolor[HTML]{1E8741}\textcolor[HTML]{FFFFFF}{27.63} & \cellcolor[HTML]{FED8C7}\textcolor[HTML]{000000}{2.78} \\

\refstepcounter{rowcount}\arabic{rowcount}\label{bg:id_3} & 5708 & \cellcolor[HTML]{75C477}\textcolor[HTML]{000000}{43.98} & \cellcolor[HTML]{6BC072}\textcolor[HTML]{000000}{57.22} & \cellcolor[HTML]{A2D99C}\textcolor[HTML]{000000}{52.85} & \cellcolor[HTML]{62BB6D}\textcolor[HTML]{FFFFFF}{55.42} & \cellcolor[HTML]{4BB062}\textcolor[HTML]{FFFFFF}{45.49} & \cellcolor[HTML]{3EA75A}\textcolor[HTML]{FFFFFF}{27.30} & \cellcolor[HTML]{70C274}\textcolor[HTML]{000000}{39.28} & \cellcolor[HTML]{80CA80}\textcolor[HTML]{000000}{30.26} & \cellcolor[HTML]{17813D}\textcolor[HTML]{FFFFFF}{40.61} & \cellcolor[HTML]{60BA6C}\textcolor[HTML]{FFFFFF}{54.79} & \cellcolor[HTML]{65BD6F}\textcolor[HTML]{000000}{52.70} & \cellcolor[HTML]{60BA6C}\textcolor[HTML]{FFFFFF}{48.24} & \cellcolor[HTML]{81CA81}\textcolor[HTML]{000000}{40.73} & \cellcolor[HTML]{F6FCF4}\textcolor[HTML]{000000}{30.72} & \cellcolor[HTML]{005F26}\textcolor[HTML]{FFFFFF}{25.51} & \cellcolor[HTML]{00441B}\textcolor[HTML]{FFFFFF}{31.58} & \cellcolor[HTML]{FCA486}\textcolor[HTML]{000000}{3.40} \\

\noalign{\color{red}\hrule height 1pt}
\refstepcounter{rowcount}\arabic{rowcount}\label{bg:id_4} & 9804 & \cellcolor[HTML]{309950}\textcolor[HTML]{FFFFFF}{45.19} & \cellcolor[HTML]{53B466}\textcolor[HTML]{FFFFFF}{57.87} & \cellcolor[HTML]{005F26}\textcolor[HTML]{FFFFFF}{57.28} & \cellcolor[HTML]{2F984F}\textcolor[HTML]{FFFFFF}{56.68} & \cellcolor[HTML]{137D39}\textcolor[HTML]{FFFFFF}{46.58} & \cellcolor[HTML]{00441B}\textcolor[HTML]{FFFFFF}{28.35} & \cellcolor[HTML]{AADDA4}\textcolor[HTML]{000000}{37.78} & \cellcolor[HTML]{329B51}\textcolor[HTML]{FFFFFF}{31.79} & \cellcolor[HTML]{1F8742}\textcolor[HTML]{FFFFFF}{40.43} & \cellcolor[HTML]{38A156}\textcolor[HTML]{FFFFFF}{55.42} & \cellcolor[HTML]{137D39}\textcolor[HTML]{FFFFFF}{54.37} & \cellcolor[HTML]{077331}\textcolor[HTML]{FFFFFF}{50.88} & \cellcolor[HTML]{48AE60}\textcolor[HTML]{FFFFFF}{41.73} & \cellcolor[HTML]{CEECC8}\textcolor[HTML]{000000}{31.07} & \cellcolor[HTML]{00441B}\textcolor[HTML]{FFFFFF}{25.88} & \cellcolor[HTML]{65BD6F}\textcolor[HTML]{000000}{23.68} & \cellcolor[HTML]{9C0D14}\textcolor[HTML]{FFFFFF}{5.51} \\
\noalign{\color{red}\hrule height 1pt}

\refstepcounter{rowcount}\arabic{rowcount}\label{bg:id_5} & 17996 & \cellcolor[HTML]{1A843F}\textcolor[HTML]{FFFFFF}{45.71} & \cellcolor[HTML]{006B2B}\textcolor[HTML]{FFFFFF}{61.16} & \cellcolor[HTML]{117B38}\textcolor[HTML]{FFFFFF}{56.46} & \cellcolor[HTML]{005723}\textcolor[HTML]{FFFFFF}{58.66} & \cellcolor[HTML]{006B2B}\textcolor[HTML]{FFFFFF}{46.97} & \cellcolor[HTML]{006529}\textcolor[HTML]{FFFFFF}{28.05} & \cellcolor[HTML]{1D8640}\textcolor[HTML]{FFFFFF}{41.68} & \cellcolor[HTML]{F7FCF5}\textcolor[HTML]{000000}{26.97} & \cellcolor[HTML]{006B2B}\textcolor[HTML]{FFFFFF}{41.18} & \cellcolor[HTML]{00441B}\textcolor[HTML]{FFFFFF}{57.29} & \cellcolor[HTML]{006529}\textcolor[HTML]{FFFFFF}{54.93} & \cellcolor[HTML]{077331}\textcolor[HTML]{FFFFFF}{50.88} & \cellcolor[HTML]{00481D}\textcolor[HTML]{FFFFFF}{44.33} & \cellcolor[HTML]{00441B}\textcolor[HTML]{FFFFFF}{32.34} & \cellcolor[HTML]{83CB82}\textcolor[HTML]{000000}{23.48} & \cellcolor[HTML]{45AD5F}\textcolor[HTML]{FFFFFF}{25.00} & \cellcolor[HTML]{F34C37}\textcolor[HTML]{FFFFFF}{4.35} \\

\refstepcounter{rowcount}\arabic{rowcount}\label{bg:id_6} & 34380 & \cellcolor[HTML]{026F2E}\textcolor[HTML]{FFFFFF}{46.25} & \cellcolor[HTML]{00441B}\textcolor[HTML]{FFFFFF}{62.44} & \cellcolor[HTML]{00692A}\textcolor[HTML]{FFFFFF}{57.04} & \cellcolor[HTML]{00441B}\textcolor[HTML]{FFFFFF}{59.14} & \cellcolor[HTML]{329B51}\textcolor[HTML]{FFFFFF}{45.95} & \cellcolor[HTML]{83CB82}\textcolor[HTML]{000000}{26.78} & \cellcolor[HTML]{3FA95C}\textcolor[HTML]{FFFFFF}{40.39} & \cellcolor[HTML]{2B934B}\textcolor[HTML]{FFFFFF}{32.00} & \cellcolor[HTML]{016E2D}\textcolor[HTML]{FFFFFF}{41.11} & \cellcolor[HTML]{006B2B}\textcolor[HTML]{FFFFFF}{56.62} & \cellcolor[HTML]{0E7936}\textcolor[HTML]{FFFFFF}{54.48} & \cellcolor[HTML]{00441B}\textcolor[HTML]{FFFFFF}{52.24} & \cellcolor[HTML]{00441B}\textcolor[HTML]{FFFFFF}{44.43} & \cellcolor[HTML]{006B2B}\textcolor[HTML]{FFFFFF}{32.14} & \cellcolor[HTML]{60BA6C}\textcolor[HTML]{FFFFFF}{23.88} & \cellcolor[HTML]{5EB96B}\textcolor[HTML]{FFFFFF}{23.94} & \cellcolor[HTML]{67000D}\textcolor[HTML]{FFFFFF}{5.89} \\

\refstepcounter{rowcount}\arabic{rowcount}\label{bg:id_7} & 50764 & \cellcolor[HTML]{00441B}\textcolor[HTML]{FFFFFF}{47.08} & \cellcolor[HTML]{005522}\textcolor[HTML]{FFFFFF}{61.85} & \cellcolor[HTML]{00441B}\textcolor[HTML]{FFFFFF}{57.97} & \cellcolor[HTML]{004D1F}\textcolor[HTML]{FFFFFF}{58.91} & \cellcolor[HTML]{00441B}\textcolor[HTML]{FFFFFF}{47.59} & \cellcolor[HTML]{F7FCF5}\textcolor[HTML]{000000}{25.46} & \cellcolor[HTML]{00441B}\textcolor[HTML]{FFFFFF}{43.77} & \cellcolor[HTML]{00441B}\textcolor[HTML]{FFFFFF}{34.00} & \cellcolor[HTML]{00441B}\textcolor[HTML]{FFFFFF}{41.94} & \cellcolor[HTML]{004E1F}\textcolor[HTML]{FFFFFF}{57.10} & \cellcolor[HTML]{00441B}\textcolor[HTML]{FFFFFF}{55.58} & \cellcolor[HTML]{005522}\textcolor[HTML]{FFFFFF}{51.78} & \cellcolor[HTML]{006C2C}\textcolor[HTML]{FFFFFF}{43.58} & \cellcolor[HTML]{F7FCF5}\textcolor[HTML]{000000}{30.71} & \cellcolor[HTML]{005321}\textcolor[HTML]{FFFFFF}{25.67} & \cellcolor[HTML]{05712F}\textcolor[HTML]{FFFFFF}{29.17} & \cellcolor[HTML]{67000D}\textcolor[HTML]{FFFFFF}{5.90} \\

\bottomrule
\end{tabular}%
}
\end{table}

\begin{table}[t]
\centering
\caption{\textbf{Generalization across backbones.} Intra-frame pruning and merging on three MLLM backbones under the same budget, protocol and prompts}
\label{tab:backbones}
\ifcsname c@rowcount\endcsname\else\newcounter{rowcount}\fi
\setcounter{rowcount}{0}
\renewcommand{\arraystretch}{1.05}
\resizebox{\textwidth}{!}{%
\begin{tabular}{@{}r l c c c c c c c c c c c c c c c c c@{}}
\toprule
 &  & \multicolumn{8}{c}{\textbf{Semantic Categories}} & \multicolumn{8}{c}{\textbf{Temporal Recall Regimes}} & \multicolumn{1}{c}{\textbf{Proc.}} \\
\cmidrule(lr){3-10}
\cmidrule(lr){11-18}
 & \textbf{Strategy} & \textbf{Avg} & \textbf{SPAT} & \textbf{TEMP} & \textbf{DETL} & \textbf{EVNT} & \textbf{SOC} & \textbf{CAUS} & \textbf{PROS} & \textbf{Avg} & \textbf{Inst} & \textbf{Short} & \textbf{S-Mid} & \textbf{Mid} & \textbf{ML} & \textbf{Long} & \textbf{Ultra} & \textbf{(s)} \\
\midrule
\multicolumn{19}{@{}l}{\rev{\textit{Qwen3-VL (8B)~\cite{bai2025qwen3vltechnicalreport}}}} \\

\refstepcounter{rowcount}\rev{\arabic{rowcount}}\label{row:bb_qwen_prune_cos} & \rev{\bPrune\ \bIntraCos} & \cellcolor[HTML]{00441B}\textcolor[HTML]{FFFFFF}{45.63} & \cellcolor[HTML]{00441B}\textcolor[HTML]{FFFFFF}{60.51} & \cellcolor[HTML]{04702F}\textcolor[HTML]{FFFFFF}{52.39} & \cellcolor[HTML]{00441B}\textcolor[HTML]{FFFFFF}{56.61} & \cellcolor[HTML]{00441B}\textcolor[HTML]{FFFFFF}{46.64} & \cellcolor[HTML]{107B37}\textcolor[HTML]{FFFFFF}{28.50} & \cellcolor[HTML]{00441B}\textcolor[HTML]{FFFFFF}{39.88} & \cellcolor[HTML]{00441B}\textcolor[HTML]{FFFFFF}{34.87} & \cellcolor[HTML]{8ED08B}\textcolor[HTML]{000000}{39.34} & \cellcolor[HTML]{006B2B}\textcolor[HTML]{FFFFFF}{55.60} & \cellcolor[HTML]{00441B}\textcolor[HTML]{FFFFFF}{54.99} & \cellcolor[HTML]{00441B}\textcolor[HTML]{FFFFFF}{52.18} & \cellcolor[HTML]{00441B}\textcolor[HTML]{FFFFFF}{41.27} & \cellcolor[HTML]{005924}\textcolor[HTML]{FFFFFF}{30.61} & \cellcolor[HTML]{00441B}\textcolor[HTML]{FFFFFF}{24.93} & \cellcolor[HTML]{F7FCF5}\textcolor[HTML]{000000}{15.79} & \cellcolor[HTML]{FFF5F0}\textcolor[HTML]{000000}{2.28} \\

\refstepcounter{rowcount}\rev{\arabic{rowcount}}\label{row:bb_qwen_prune_pa} & \rev{\bPrune\ \bIntraPA} & \cellcolor[HTML]{026F2D}\textcolor[HTML]{FFFFFF}{45.12} & \cellcolor[HTML]{0B7633}\textcolor[HTML]{FFFFFF}{59.47} & \cellcolor[HTML]{00441B}\textcolor[HTML]{FFFFFF}{53.08} & \cellcolor[HTML]{00491D}\textcolor[HTML]{FFFFFF}{56.52} & \cellcolor[HTML]{107B37}\textcolor[HTML]{FFFFFF}{45.87} & \cellcolor[HTML]{00441B}\textcolor[HTML]{FFFFFF}{28.72} & \cellcolor[HTML]{228A44}\textcolor[HTML]{FFFFFF}{39.28} & \cellcolor[HTML]{73C375}\textcolor[HTML]{000000}{32.89} & \cellcolor[HTML]{00441B}\textcolor[HTML]{FFFFFF}{41.50} & \cellcolor[HTML]{00441B}\textcolor[HTML]{FFFFFF}{56.04} & \cellcolor[HTML]{006127}\textcolor[HTML]{FFFFFF}{54.71} & \cellcolor[HTML]{218943}\textcolor[HTML]{FFFFFF}{50.18} & \cellcolor[HTML]{0A7533}\textcolor[HTML]{FFFFFF}{40.60} & \cellcolor[HTML]{00441B}\textcolor[HTML]{FFFFFF}{30.84} & \cellcolor[HTML]{F7FCF5}\textcolor[HTML]{000000}{22.61} & \cellcolor[HTML]{00441B}\textcolor[HTML]{FFFFFF}{35.53} & \cellcolor[HTML]{8F0912}\textcolor[HTML]{FFFFFF}{2.82} \\

\refstepcounter{rowcount}\rev{\arabic{rowcount}}\label{row:bb_qwen_merge_cos} & \rev{\bMerge\ \bIntraCos} & \cellcolor[HTML]{93D290}\textcolor[HTML]{000000}{43.42} & \cellcolor[HTML]{78C67A}\textcolor[HTML]{000000}{57.30} & \cellcolor[HTML]{ADDEA7}\textcolor[HTML]{000000}{49.83} & \cellcolor[HTML]{A8DCA2}\textcolor[HTML]{000000}{53.28} & \cellcolor[HTML]{107B37}\textcolor[HTML]{FFFFFF}{45.87} & \cellcolor[HTML]{F7FCF5}\textcolor[HTML]{000000}{27.52} & \cellcolor[HTML]{89CE87}\textcolor[HTML]{000000}{38.53} & \cellcolor[HTML]{DBF1D5}\textcolor[HTML]{000000}{31.58} & \cellcolor[HTML]{E9F6E5}\textcolor[HTML]{000000}{38.10} & \cellcolor[HTML]{60BA6C}\textcolor[HTML]{FFFFFF}{54.44} & \cellcolor[HTML]{95D391}\textcolor[HTML]{000000}{53.20} & \cellcolor[HTML]{70C274}\textcolor[HTML]{000000}{48.18} & \cellcolor[HTML]{63BB6D}\textcolor[HTML]{FFFFFF}{39.40} & \cellcolor[HTML]{F7FCF5}\textcolor[HTML]{000000}{27.50} & \cellcolor[HTML]{E4F4DF}\textcolor[HTML]{000000}{22.90} & \cellcolor[HTML]{C1E6BB}\textcolor[HTML]{000000}{21.05} & \cellcolor[HTML]{FEE5D9}\textcolor[HTML]{000000}{2.33} \\

\refstepcounter{rowcount}\rev{\arabic{rowcount}}\label{row:bb_qwen_merge_pa} & \rev{\bMerge\ \bIntraPA} & \cellcolor[HTML]{F7FCF5}\textcolor[HTML]{000000}{41.87} & \cellcolor[HTML]{F7FCF5}\textcolor[HTML]{000000}{54.25} & \cellcolor[HTML]{F7FCF5}\textcolor[HTML]{000000}{48.20} & \cellcolor[HTML]{F7FCF5}\textcolor[HTML]{000000}{51.49} & \cellcolor[HTML]{F7FCF5}\textcolor[HTML]{000000}{42.48} & \cellcolor[HTML]{40AA5C}\textcolor[HTML]{FFFFFF}{28.27} & \cellcolor[HTML]{F7FCF5}\textcolor[HTML]{000000}{37.48} & \cellcolor[HTML]{F7FCF5}\textcolor[HTML]{000000}{30.92} & \cellcolor[HTML]{F7FCF5}\textcolor[HTML]{000000}{37.73} & \cellcolor[HTML]{F7FCF5}\textcolor[HTML]{000000}{52.47} & \cellcolor[HTML]{F7FCF5}\textcolor[HTML]{000000}{51.98} & \cellcolor[HTML]{F7FCF5}\textcolor[HTML]{000000}{44.00} & \cellcolor[HTML]{F7FCF5}\textcolor[HTML]{000000}{37.20} & \cellcolor[HTML]{AADCA3}\textcolor[HTML]{000000}{28.65} & \cellcolor[HTML]{A0D89A}\textcolor[HTML]{000000}{23.48} & \cellcolor[HTML]{66BD6F}\textcolor[HTML]{000000}{26.32} & \cellcolor[HTML]{67000C}\textcolor[HTML]{FFFFFF}{2.87} \\

\midrule
\multicolumn{19}{@{}l}{\rev{\textit{InternVL~3.5 (8B)~\cite{wang2025internvl35}}}} \\

\refstepcounter{rowcount}\rev{\arabic{rowcount}}\label{row:bb_internvl_prune_cos} & \rev{\bPrune\ \bIntraCos} & \cellcolor[HTML]{00441B}\textcolor[HTML]{FFFFFF}{43.09} & \cellcolor[HTML]{00441B}\textcolor[HTML]{FFFFFF}{56.02} & \cellcolor[HTML]{56B567}\textcolor[HTML]{FFFFFF}{53.20} & \cellcolor[HTML]{00441B}\textcolor[HTML]{FFFFFF}{50.83} & \cellcolor[HTML]{00441B}\textcolor[HTML]{FFFFFF}{45.82} & \cellcolor[HTML]{A4DA9E}\textcolor[HTML]{000000}{26.55} & \cellcolor[HTML]{9CD697}\textcolor[HTML]{000000}{35.98} & \cellcolor[HTML]{00441B}\textcolor[HTML]{FFFFFF}{33.22} & \cellcolor[HTML]{64BC6E}\textcolor[HTML]{000000}{37.91} & \cellcolor[HTML]{005321}\textcolor[HTML]{FFFFFF}{52.70} & \cellcolor[HTML]{00441B}\textcolor[HTML]{FFFFFF}{50.92} & \cellcolor[HTML]{00441B}\textcolor[HTML]{FFFFFF}{48.24} & \cellcolor[HTML]{00441B}\textcolor[HTML]{FFFFFF}{39.27} & \cellcolor[HTML]{00441B}\textcolor[HTML]{FFFFFF}{27.96} & \cellcolor[HTML]{00441B}\textcolor[HTML]{FFFFFF}{25.22} & \cellcolor[HTML]{F7FCF5}\textcolor[HTML]{000000}{21.05} & \cellcolor[HTML]{FFF5F0}\textcolor[HTML]{000000}{2.34} \\

\refstepcounter{rowcount}\rev{\arabic{rowcount}}\label{row:bb_internvl_prune_pa} & \rev{\bPrune\ \bIntraPA} & \cellcolor[HTML]{005924}\textcolor[HTML]{FFFFFF}{42.92} & \cellcolor[HTML]{006227}\textcolor[HTML]{FFFFFF}{55.54} & \cellcolor[HTML]{3CA659}\textcolor[HTML]{FFFFFF}{53.32} & \cellcolor[HTML]{2E974E}\textcolor[HTML]{FFFFFF}{49.69} & \cellcolor[HTML]{238B45}\textcolor[HTML]{FFFFFF}{44.89} & \cellcolor[HTML]{00441B}\textcolor[HTML]{FFFFFF}{28.12} & \cellcolor[HTML]{00441B}\textcolor[HTML]{FFFFFF}{37.63} & \cellcolor[HTML]{ADDEA7}\textcolor[HTML]{000000}{31.25} & \cellcolor[HTML]{00441B}\textcolor[HTML]{FFFFFF}{39.00} & \cellcolor[HTML]{00441B}\textcolor[HTML]{FFFFFF}{52.79} & \cellcolor[HTML]{006829}\textcolor[HTML]{FFFFFF}{50.58} & \cellcolor[HTML]{005F26}\textcolor[HTML]{FFFFFF}{47.88} & \cellcolor[HTML]{18813D}\textcolor[HTML]{FFFFFF}{38.53} & \cellcolor[HTML]{036F2E}\textcolor[HTML]{FFFFFF}{27.27} & \cellcolor[HTML]{F7FCF5}\textcolor[HTML]{000000}{24.35} & \cellcolor[HTML]{00441B}\textcolor[HTML]{FFFFFF}{31.58} & \cellcolor[HTML]{C0151A}\textcolor[HTML]{FFFFFF}{2.49} \\

\refstepcounter{rowcount}\rev{\arabic{rowcount}}\label{row:bb_internvl_merge_cos} & \rev{\bMerge\ \bIntraCos} & \cellcolor[HTML]{66BD6F}\textcolor[HTML]{000000}{41.92} & \cellcolor[HTML]{D9F0D3}\textcolor[HTML]{000000}{51.85} & \cellcolor[HTML]{00441B}\textcolor[HTML]{FFFFFF}{53.90} & \cellcolor[HTML]{067230}\textcolor[HTML]{FFFFFF}{50.26} & \cellcolor[HTML]{81CA81}\textcolor[HTML]{000000}{43.85} & \cellcolor[HTML]{F7FCF5}\textcolor[HTML]{000000}{25.65} & \cellcolor[HTML]{006829}\textcolor[HTML]{FFFFFF}{37.33} & \cellcolor[HTML]{E7F5E2}\textcolor[HTML]{000000}{30.59} & \cellcolor[HTML]{EBF7E7}\textcolor[HTML]{000000}{36.85} & \cellcolor[HTML]{1E8742}\textcolor[HTML]{FFFFFF}{52.34} & \cellcolor[HTML]{17803C}\textcolor[HTML]{FFFFFF}{50.31} & \cellcolor[HTML]{37A155}\textcolor[HTML]{FFFFFF}{46.82} & \cellcolor[HTML]{4CB062}\textcolor[HTML]{FFFFFF}{37.87} & \cellcolor[HTML]{F7FCF5}\textcolor[HTML]{000000}{23.01} & \cellcolor[HTML]{00441B}\textcolor[HTML]{FFFFFF}{25.22} & \cellcolor[HTML]{E5F5E0}\textcolor[HTML]{000000}{22.37} & \cellcolor[HTML]{FEEAE0}\textcolor[HTML]{000000}{2.36} \\

\refstepcounter{rowcount}\rev{\arabic{rowcount}}\label{row:bb_internvl_merge_pa} & \rev{\bMerge\ \bIntraPA} & \cellcolor[HTML]{F7FCF5}\textcolor[HTML]{000000}{40.57} & \cellcolor[HTML]{F7FCF5}\textcolor[HTML]{000000}{50.96} & \cellcolor[HTML]{F7FCF5}\textcolor[HTML]{000000}{52.27} & \cellcolor[HTML]{F7FCF5}\textcolor[HTML]{000000}{47.03} & \cellcolor[HTML]{F7FCF5}\textcolor[HTML]{000000}{42.15} & \cellcolor[HTML]{B7E2B0}\textcolor[HTML]{000000}{26.40} & \cellcolor[HTML]{F7FCF5}\textcolor[HTML]{000000}{34.93} & \cellcolor[HTML]{F7FCF5}\textcolor[HTML]{000000}{30.26} & \cellcolor[HTML]{F7FCF5}\textcolor[HTML]{000000}{36.65} & \cellcolor[HTML]{F7FCF5}\textcolor[HTML]{000000}{50.91} & \cellcolor[HTML]{F7FCF5}\textcolor[HTML]{000000}{47.97} & \cellcolor[HTML]{F7FCF5}\textcolor[HTML]{000000}{44.06} & \cellcolor[HTML]{F7FCF5}\textcolor[HTML]{000000}{35.80} & \cellcolor[HTML]{B7E2B0}\textcolor[HTML]{000000}{24.51} & \cellcolor[HTML]{F7FCF5}\textcolor[HTML]{000000}{24.35} & \cellcolor[HTML]{228A44}\textcolor[HTML]{FFFFFF}{28.95} & \cellcolor[HTML]{67000C}\textcolor[HTML]{FFFFFF}{2.53} \\

\midrule
\multicolumn{19}{@{}l}{\rev{\textit{Molmo~2 (8B)~\cite{molmo2}}}} \\

\refstepcounter{rowcount}\rev{\arabic{rowcount}}\label{row:bb_molmo_prune_cos} & \rev{\bPrune\ \bIntraCos} & \cellcolor[HTML]{00441B}\textcolor[HTML]{FFFFFF}{41.61} & \cellcolor[HTML]{00441B}\textcolor[HTML]{FFFFFF}{53.63} & \cellcolor[HTML]{00441B}\textcolor[HTML]{FFFFFF}{52.01} & \cellcolor[HTML]{00441B}\textcolor[HTML]{FFFFFF}{41.97} & \cellcolor[HTML]{00441B}\textcolor[HTML]{FFFFFF}{54.61} & \cellcolor[HTML]{40AA5C}\textcolor[HTML]{FFFFFF}{27.83} & \cellcolor[HTML]{00441B}\textcolor[HTML]{FFFFFF}{28.36} & \cellcolor[HTML]{98D493}\textcolor[HTML]{000000}{32.89} & \cellcolor[HTML]{00441B}\textcolor[HTML]{FFFFFF}{36.83} & \cellcolor[HTML]{00441B}\textcolor[HTML]{FFFFFF}{51.97} & \cellcolor[HTML]{00441B}\textcolor[HTML]{FFFFFF}{45.82} & \cellcolor[HTML]{00441B}\textcolor[HTML]{FFFFFF}{39.35} & \cellcolor[HTML]{C9EAC2}\textcolor[HTML]{000000}{33.67} & \cellcolor[HTML]{00441B}\textcolor[HTML]{FFFFFF}{28.38} & \cellcolor[HTML]{00441B}\textcolor[HTML]{FFFFFF}{30.99} & \cellcolor[HTML]{F7FCF5}\textcolor[HTML]{000000}{27.63} & \cellcolor[HTML]{FFF5F0}\textcolor[HTML]{000000}{14.16} \\

\refstepcounter{rowcount}\rev{\arabic{rowcount}}\label{row:bb_molmo_prune_pa} & \rev{\bPrune\ \bIntraPA} & \cellcolor[HTML]{AEDEA8}\textcolor[HTML]{000000}{39.60} & \cellcolor[HTML]{74C476}\textcolor[HTML]{000000}{51.05} & \cellcolor[HTML]{70C274}\textcolor[HTML]{000000}{50.62} & \cellcolor[HTML]{005C25}\textcolor[HTML]{FFFFFF}{41.71} & \cellcolor[HTML]{DFF2D9}\textcolor[HTML]{000000}{51.89} & \cellcolor[HTML]{00441B}\textcolor[HTML]{FFFFFF}{29.31} & \cellcolor[HTML]{F7FCF5}\textcolor[HTML]{000000}{22.39} & \cellcolor[HTML]{F7FCF5}\textcolor[HTML]{000000}{30.26} & \cellcolor[HTML]{4BB061}\textcolor[HTML]{FFFFFF}{35.99} & \cellcolor[HTML]{279048}\textcolor[HTML]{FFFFFF}{50.85} & \cellcolor[HTML]{117C38}\textcolor[HTML]{FFFFFF}{44.69} & \cellcolor[HTML]{68BE70}\textcolor[HTML]{000000}{37.22} & \cellcolor[HTML]{F7FCF5}\textcolor[HTML]{000000}{33.02} & \cellcolor[HTML]{005C25}\textcolor[HTML]{FFFFFF}{28.23} & \cellcolor[HTML]{50B264}\textcolor[HTML]{FFFFFF}{28.95} & \cellcolor[HTML]{ADDEA7}\textcolor[HTML]{000000}{28.95} & \cellcolor[HTML]{67000C}\textcolor[HTML]{FFFFFF}{16.14} \\

\refstepcounter{rowcount}\rev{\arabic{rowcount}}\label{row:bb_molmo_merge_cos} & \rev{\bMerge\ \bIntraCos} & \cellcolor[HTML]{B2E0AB}\textcolor[HTML]{000000}{39.57} & \cellcolor[HTML]{F7FCF5}\textcolor[HTML]{000000}{48.49} & \cellcolor[HTML]{F4FAF1}\textcolor[HTML]{000000}{49.24} & \cellcolor[HTML]{C0E6B9}\textcolor[HTML]{000000}{39.46} & \cellcolor[HTML]{F5FBF3}\textcolor[HTML]{000000}{51.46} & \cellcolor[HTML]{F7FCF5}\textcolor[HTML]{000000}{25.37} & \cellcolor[HTML]{41AB5D}\textcolor[HTML]{FFFFFF}{26.12} & \cellcolor[HTML]{00441B}\textcolor[HTML]{FFFFFF}{36.84} & \cellcolor[HTML]{EDF8EA}\textcolor[HTML]{000000}{34.86} & \cellcolor[HTML]{D3EECD}\textcolor[HTML]{000000}{48.66} & \cellcolor[HTML]{AEDEA8}\textcolor[HTML]{000000}{41.85} & \cellcolor[HTML]{C8E9C1}\textcolor[HTML]{000000}{35.94} & \cellcolor[HTML]{00441B}\textcolor[HTML]{FFFFFF}{35.75} & \cellcolor[HTML]{D1EDCA}\textcolor[HTML]{000000}{26.83} & \cellcolor[HTML]{F7FCF5}\textcolor[HTML]{000000}{26.02} & \cellcolor[HTML]{ADDEA7}\textcolor[HTML]{000000}{28.95} & \cellcolor[HTML]{FEF3ED}\textcolor[HTML]{000000}{14.19} \\

\refstepcounter{rowcount}\rev{\arabic{rowcount}}\label{row:bb_molmo_merge_pa} & \rev{\bMerge\ \bIntraPA} & \cellcolor[HTML]{F7FCF5}\textcolor[HTML]{000000}{38.61} & \cellcolor[HTML]{E7F5E2}\textcolor[HTML]{000000}{49.06} & \cellcolor[HTML]{F7FCF5}\textcolor[HTML]{000000}{49.18} & \cellcolor[HTML]{F7FCF5}\textcolor[HTML]{000000}{38.52} & \cellcolor[HTML]{F7FCF5}\textcolor[HTML]{000000}{51.42} & \cellcolor[HTML]{EAF6E5}\textcolor[HTML]{000000}{25.72} & \cellcolor[HTML]{E4F4DF}\textcolor[HTML]{000000}{23.13} & \cellcolor[HTML]{85CC84}\textcolor[HTML]{000000}{33.22} & \cellcolor[HTML]{F7FCF5}\textcolor[HTML]{000000}{34.72} & \cellcolor[HTML]{F7FCF5}\textcolor[HTML]{000000}{47.85} & \cellcolor[HTML]{F7FCF5}\textcolor[HTML]{000000}{39.88} & \cellcolor[HTML]{F7FCF5}\textcolor[HTML]{000000}{34.83} & \cellcolor[HTML]{2A934B}\textcolor[HTML]{FFFFFF}{34.97} & \cellcolor[HTML]{F7FCF5}\textcolor[HTML]{000000}{26.42} & \cellcolor[HTML]{B9E3B2}\textcolor[HTML]{000000}{27.49} & \cellcolor[HTML]{00441B}\textcolor[HTML]{FFFFFF}{31.58} & \cellcolor[HTML]{70020E}\textcolor[HTML]{FFFFFF}{16.10} \\

\bottomrule
\end{tabular}%
}
{
\vspace{1pt}
\footnotesize
Memory budget: 9,804 tokens; \bFIFO\ \bSink\ \bHat\ enabled for all rows.}

\end{table}

% ---------------------------------------------------------
% SECTION 6: CONCLUSION
% ---------------------------------------------------------
\section{Conclusions}
\label{sec:conclusion}
% TO DO:
% - Summarize the core contribution (EgoStream benchmark + unified diagnostic framework).
% - Highlight the main takeaway from the experiments (e.g., current methods over-compress and lose spatial/temporal fidelity).
% - State Limitations: e.g., the benchmark relies on specific definitions of AVW, or the unified framework only explores KV-cache level memory rather than parametric weight updates.

% \egostream establishes a comprehensive diagnostic benchmark specifically designed to evaluate streaming episodic memory in egocentric vision. By organizing 2,250 curated questions across seven distinct cognitive dimensions and introducing the Answer Validity Window (AVW), the benchmark enables the controlled testing of memory retention over time, effectively separating genuine model forgetting from natural state changes in the environment. Furthermore, the introduction of a unified streaming framework provides a standardized testbed for comparing various memory-management strategies, such as token pruning and merging. Our evaluations reveal that comparable aggregate accuracies often mask significantly different semantic and temporal memory preservation profiles. Ultimately, \egostream moves beyond simple aggregate performance metrics, offering a nuanced tool to diagnose exactly what embodied systems remember and for how long, paving the way for more reliable, long-horizon assistive agents.

\egostream establishes a diagnostic benchmark and unified framework for evaluating streaming episodic memory in egocentric vision. By introducing the Answer Validity Window (AVW) to decouple genuine forgetting from natural scene changes, our evaluations reveal that comparable aggregate accuracies often mask distinct memory vulnerabilities. Moving beyond simple performance metrics, \egostream provides a precise tool to diagnose exactly what, and for how long, embodied systems remember—laying the groundwork for reliable, context-aware assistive agents.

\begin{ack}
We acknowledge ISCRA for awarding this project access to the LEONARDO supercomputer, owned by the EuroHPC Joint Undertaking, hosted by CINECA (Italy) .
We acknowledge Giulio Pedicone, Michela Maria Tasca and Emanuele Galiano for their valuable contribution to the data validation process.

\textbf{Funding and Competing Interests.} Aside from the computational resources acknowledged above, the authors received no other third-party funding for this work and declare no competing financial or professional interests over the past 36 months.
\end{ack}

\bibliographystyle{plain} 
\bibliography{parts/references}

\appendix
% \section{Technical appendices and supplementary material}
% Technical appendices with additional results, figures, graphs, and proofs may be submitted with the paper submission before the full submission deadline (see above). You can upload a ZIP file for videos or code, but do not upload a separate PDF file for the appendix. There is no page limit for the technical appendices. 

% Note: Think of the appendix as ``optional reading'' for reviewers. The paper must be able to stand alone without the appendix; for example, adding critical experiments that support the main claims to an appendix is inappropriate. 

\section{The \egostream Benchmark}
\label{app:egostream_benchmark}
This appendix reports additional details for the \egostream Benchmark construction process.

% \subsection{Source Curation and Initial Set of Questions}
% \label{app:egostream_benchmark_construction}
% % See the main paper to check what goes here. Let's put all details, prompts etc.
% In this section, we report details on the process used to obtain the initial set of $2,400$ curated question-answer sets.

\subsection{Hard Negative Generation Protocol}
\label{app:negative_generation}

To support a unified multiple-choice evaluation protocol, we generated three hard negative answers for each valid open-ended question-answer pair. The protocol is data-agnostic: given a question $q$ and a ground-truth answer $a$, it produces a set of plausible but incorrect distractors that preserve the linguistic form, level of specificity, and granularity of the correct answer while changing its factual content. This design prevents multiple-choice evaluation from becoming artificially easy due to generic, implausible, or structurally mismatched alternatives, and encourages models to rely on episodic visual evidence rather than superficial answer-format cues.

We instantiated the generation and validation pipeline using Gemini~3.1~\cite{gemini} as the LLM backend. For each valid question-answer pair, the model was queried in JSON-only mode. Before generation, we discarded items whose answer was empty or malformed. For instance, questions with placeholder answers such as ``.'' were removed from the generation pool, since they do not provide a reliable positive answer from which to construct structured distractors.

We then generated one initial negative answer conditioned on both the question and the ground-truth answer, using the Prompt~\ref{prompt:first_negative}. This first step allows the model to explicitly match the structure of the correct answer: if the ground truth is a short noun phrase, the negative is also required to be a short noun phrase; if the ground truth is a full sentence, the negative must preserve a comparable sentence structure. The prompt further instructs the model to keep similar length, tense, phrasing style, and level of detail, while changing the core content.

After this first negative was obtained, we generated the remaining two negatives without exposing the ground-truth answer to the model, using the Prompt~\ref{prompt:additional_negatives}. Instead, the model was given only the question and the first accepted negative answer. This second stage reduces the risk that all distractors are produced by directly perturbing the same ground-truth phrase, while still enforcing consistency in answer format. In other words, the first negative acts as a style and granularity anchor for the remaining distractors.

As an illustrative example, consider the following question:
\begin{quote}
\textit{What objects did I connect the white and black cable to?}
\end{quote}
with ground-truth answer:
\begin{quote}
\textit{You connected the cable to the charger and the laptop.}
\end{quote}
The generated negatives were:
\begin{quote}
\textit{You connected the cable to the external monitor and the docking station.}

\textit{You connected the cable to the wireless router and the network switch.}

\textit{You connected the cable to the external hard drive and the USB hub.}
\end{quote}
All three distractors preserve the same answer structure as the ground-truth answer, namely ``You connected the cable to [object] and [object].'' They also maintain comparable granularity by referring to plausible electronic devices and connection targets. However, the content is changed, since none of the distractors corresponds to the correct pair of objects connected in the video. This example illustrates the intended behavior of the pipeline: negatives should be difficult because they are stylistically and semantically plausible, not because they are vague or unrelated.

All candidate negatives were then passed through two levels of filtering. First, we applied lexical and string-similarity checks to reject candidates that were too close to the ground-truth answer or duplicated another already accepted negative. Answers were normalized by lowercasing, stripping punctuation and articles, and collapsing whitespace. We then computed lexical similarity using Python's \texttt{difflib.SequenceMatcher} over normalized answer strings. The resulting score lies in $[0,1]$ and is computed as $2M/T$, where $M$ is the number of matching characters in the aligned matching blocks and $T$ is the total number of characters in the two normalized strings. A score of $1$ indicates identical normalized strings. A candidate was rejected if it was identical or near-identical to the ground-truth answer, using a threshold of $0.92$, or if it duplicated an already accepted negative, using a threshold of $0.90$.

Second, every remaining candidate was evaluated by an LLM validator (Gemini 3.1) using the Prompt~\ref{prompt:negative_validation}. The validator was shown the question, the ground-truth answer, and the candidate negative, and was asked to verify that the candidate satisfies all required properties: factual content must differ from the ground truth, answer structure must be preserved, granularity must remain comparable, the answer must be plausible for the visual-question context, and the distractor must not be obviously easy. A candidate was accepted only if all validation flags were satisfied. If fewer than three candidates passed validation, we entered a regeneration loop in which additional candidates were generated one at a time using the question, the first accepted negative, and the list of banned answers, as specified in Prompt~\ref{prompt:regenerate_negative}. This loop continued until three valid negatives were obtained or a maximum number of regeneration rounds was reached. Items for which three valid negatives could not be obtained were removed from the final multiple-choice subset.

Algorithm~\ref{alg:negative_generation} summarizes the complete procedure.

\begin{algorithm}[t]
\caption{Hard negative generation and validation}
\label{alg:negative_generation}
\DontPrintSemicolon
\KwIn{Question $q$, ground-truth answer $a$, target number of negatives $K=3$}
\KwOut{Set of validated negatives $\mathcal{N}$}

\If{$a$ is empty or malformed}{
    discard item\;
}

$\mathcal{N} \leftarrow \emptyset$\;

Generate first candidate $\tilde{a}_1$ from $(q,a)$\;

\If{\textsc{IsValidNegative}$(q,a,\tilde{a}_1,\mathcal{N})$}{
    $\mathcal{N} \leftarrow \mathcal{N} \cup \{\tilde{a}_1\}$\;
}
\Else{
    retry first-negative generation up to a fixed limit\;
}

Generate additional candidates $\tilde{a}_2,\tilde{a}_3$ from $(q,\tilde{a}_1)$\;

\ForEach{candidate $\tilde{a}$ in $\{\tilde{a}_2,\tilde{a}_3\}$}{
    \If{\textsc{IsValidNegative}$(q,a,\tilde{a},\mathcal{N})$}{
        $\mathcal{N} \leftarrow \mathcal{N} \cup \{\tilde{a}\}$\;
    }
}

\While{$|\mathcal{N}| < K$ and regeneration budget not exhausted}{
    Generate a new candidate $\tilde{a}$ from $(q,\tilde{a}_1,\mathcal{N}\cup\{a\})$\;
    \If{\textsc{IsValidNegative}$(q,a,\tilde{a},\mathcal{N})$}{
        $\mathcal{N} \leftarrow \mathcal{N} \cup \{\tilde{a}\}$\;
    }
}

\If{$|\mathcal{N}| < K$}{
    discard item\;
}

\Return first $K$ elements of $\mathcal{N}$\;
\end{algorithm}

The validation function \textsc{IsValidNegative} combines deterministic similarity filtering and model-based verification. Let $\mathrm{norm}(\cdot)$ denote the normalized answer string and let $\mathrm{sim}(\cdot,\cdot)$ denote sequence similarity. A candidate $\tilde{a}$ is rejected if $\mathrm{norm}(\tilde{a})=\mathrm{norm}(a)$, if $\mathrm{sim}(\tilde{a},a)\geq0.92$, or if it is identical or near-identical to any previously accepted negative, using a threshold of $0.90$. Candidates passing these checks are then accepted only if the LLM validator confirms that the factual content changed, while structure, granularity, plausibility, and difficulty are preserved.

We applied this hard-negative generation protocol to the datasets in our benchmark pipeline that originally required multiple-choice distractor construction, namely Multi-Hop EgoQA~\cite{chen2025multihop} and EgoTempo~\cite{plizzari2025egotempo}, while keeping all other data sources intact (Ego4D GroundVQA~\cite{di2024groundvqa}, EgoLife~\cite{yang2025egolife}, and HD-EPIC~\cite{perrett2025hdepic}). After this first generation step, our dataset consists of 2,634 question-answer pairs with negatives.

\begin{PromptBlock}{First negative generation}
\label{prompt:first_negative}
\begin{PromptText}
Generate exactly one incorrect answer for a visual QA example.

Goal:
The incorrect answer must be HARD:
- it must preserve the same answer structure as the ground-truth answer
- it must preserve the same level of detail/granularity
- it must preserve the same style and syntactic form
- but it must change the core content so that it is factually incorrect

Rules:
- If the ground-truth is a full sentence, output a full sentence with the same framing.
- If the ground-truth is a short noun phrase, output a short noun phrase.
- Keep similar length.
- Keep articles, tense, and phrasing style aligned with the ground-truth when possible.
- Change the object/entity/action detail so the answer is wrong.
- Do NOT make it obviously easy or generic.
- Do NOT explain.

Question: {question}
Ground truth answer: {gt_answer}

Return format:
{
  "negative": "..."
}
\end{PromptText}
\end{PromptBlock}
\begin{PromptBlock}{Additional negative generation}

\label{prompt:additional_negatives}

\begin{PromptText}
Generate {n} additional incorrect answers for this question.

Important constraints:
- You are NOT given the ground-truth answer.
- Use only the question and the existing incorrect answer.
- The new answers must preserve the SAME structural style, granularity, and answer format
  as the existing incorrect answer.
- They must be different in content from the existing incorrect answer and from each other.
- They must remain plausible and hard.
- Do NOT make them generic or obviously easy.
- Do not explain.

Question: {question}
Existing incorrect answer: {first_negative}

Return format:
{
  "negatives": ["...", "..."]
}
\end{PromptText}
\end{PromptBlock}
\begin{PromptBlock}{Prompt for candidate regeneration.}
\label{prompt:regenerate_negative}
\begin{PromptText}
Generate exactly one additional incorrect answer for this question.

Important constraints:
- You are NOT given the ground-truth answer.
- Use only the question and the existing incorrect answer.
- The new answer must preserve the same structural style, granularity, and answer format
  as the existing incorrect answer.
- The new answer must be different from all banned answers.
- It must be plausible and hard, not generic or obviously easy.
- Do not explain.

Question: {question}
Existing incorrect answer: {first_negative}
Banned answers: {banned_answers}

Return format:
{
  "negative": "..."
}
\end{PromptText}
\end{PromptBlock}
\begin{PromptBlock}{Negative validation.}
\label{prompt:negative_validation}

\begin{PromptText}
You are validating a hard negative answer for a visual QA benchmark.

A GOOD hard negative must satisfy ALL of the following:
1. It is clearly different in factual content from the ground-truth answer.
2. It preserves the same answer format/structure as the ground-truth answer.
3. It preserves similar granularity/detail level as the ground-truth answer.
4. It is plausible for the scene/question context.
5. It is not a trivial or obviously easy distractor.
6. It should be confusable with the ground-truth by structure/style, but not by factual content.

Question: {question}
Ground-truth answer: {gt_answer}
Candidate negative: {candidate}

Return ONLY one JSON object:
{
  "valid": true,
  "content_changed": true,
  "structure_preserved": true,
  "granularity_preserved": true,
  "plausible": true,
  "too_easy": false,
  "reason": "brief explanation"
}
\end{PromptText}
\end{PromptBlock}

\subsection{Evidence Moment Annotation}
\label{app:evidence_moments}

A central requirement of our streaming evaluation protocol is to know when the visual information needed to answer a question can be observed in the video. We refer to this temporal support as the \emph{evidence moment}. Formally, for each question-answer pair, the evidence moment identifies the timestamp or temporal interval in which the answer can be grounded in the visual stream. This annotation is used to align each question with the relevant observation window and to define subsequent recall settings.

Our evidence-moment construction is dataset-dependent but follows a unified principle: we preserve the temporal grounding provided by each source dataset and convert it into a common representation. This avoids introducing unnecessary additional assumptions and keeps the benchmark faithful to the temporal structure of the original annotations.

For \textbf{Ego4D Episodic Memory VQA}~\cite{baermann2022keys}, we use the evidence annotations already provided by the dataset. These annotations identify the temporal support associated with the answer and are directly converted into our unified evidence-moment format.

For \textbf{EgoLife}~\cite{yang2025egolife}, the original temporal grounding is provided as point-level evidence rather than interval-level evidence. We preserve these point annotations as evidence timestamps. In our unified representation, such evidence points serve as temporal anchors at which the answer becomes observable in the stream.

For \textbf{Multi-Hop EgoQA}~\cite{chen2025multihop}, we use the evidence moments provided by the dataset. Since Multi-Hop EgoQA is designed around compositional questions, a single question may require multiple temporally separated pieces of evidence. We therefore retain all annotated supporting moments and represent them as a set of evidence intervals, $\mathcal{M}=\{M_1,\ldots,M_n\}$, where each $M_i$ corresponds to one required visual support segment.

For \textbf{HD-EPIC}~\cite{perrett2025hdepic}, we use the temporally grounded evidence already available in the source annotations. These moments identify the action or event segment from which the answer can be inferred.

For \textbf{EgoTempo}~\cite{plizzari2025egotempo}, the provided clips are trimmed around the action queried by the question. Since each clip is constructed to contain the queried action by design, we treat the full provided clip as the evidence moment. Thus, for an EgoTempo sample with clip duration $[t_s,t_e]$, the evidence moment is defined as $M=[t_s,t_e]$.

\begin{figure}[t]
    \centering
    \includegraphics[width=\linewidth]{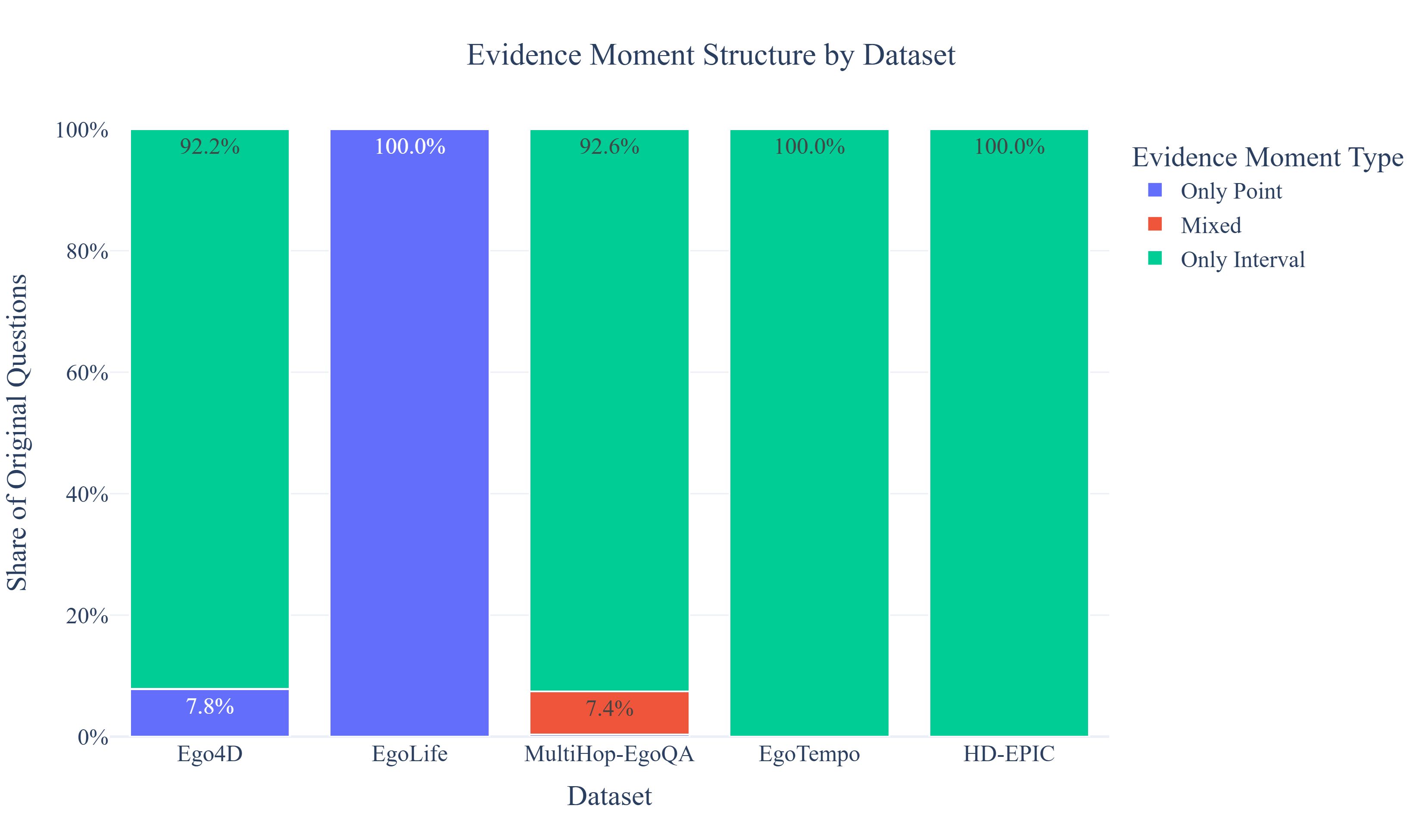}
    \caption{\textbf{Evidence moment structure by dataset.}
    We report the distribution of original questions whose evidence is represented as point-level timestamps, interval-level segments, or mixed evidence. }
    \label{fig:evidence_moment_structure}
\end{figure}

Figure~\ref{fig:evidence_moment_structure} summarizes the resulting evidence-moment structure across datasets, highlighting the diversity of the source annotations: EgoLife~\cite{yang2025egolife} is entirely point-based, EgoTempo~\cite{plizzari2025egotempo} and HD-EPIC~\cite{perrett2025hdepic} are entirely interval-based, while Ego4D Episodic Memory VQA~\cite{baermann2022keys} and Multi-Hop EgoQA~\cite{chen2025multihop} contain a mixture of evidence structures.

This process yields a consistent temporal grounding format across all five source datasets while respecting their original annotation schemes. In particular, interval annotations are preserved when available, point-level annotations are retained as temporal anchors, multi-hop annotations are represented as sets of evidence moments, and action-centered trimmed clips are treated as full-clip evidence. We paired all 2,634 examples from our hard negative generation step (See Appendix~\ref{app:negative_generation}).

\subsection{Human Validation}
\label{app:human_validation}

To improve the quality of the benchmark, we performed an additional human validation stage after the automatic and manual curation steps. Validators were asked to inspect each candidate sample through a web-based interface showing the video segment, the question, the correct answer, the distractor answers, and the annotated evidence moments. For each item, validators selected whether the question, answer, and options were valid, while also checking that the relevant action or visual evidence occurred within the annotated evidence moments. Validators could also leave a free-text note explaining the reason for rejection or suggesting a correction.

\begin{figure}[t]
    \centering
    \begin{subfigure}{0.49\linewidth}
        \centering
        \includegraphics[width=\linewidth]{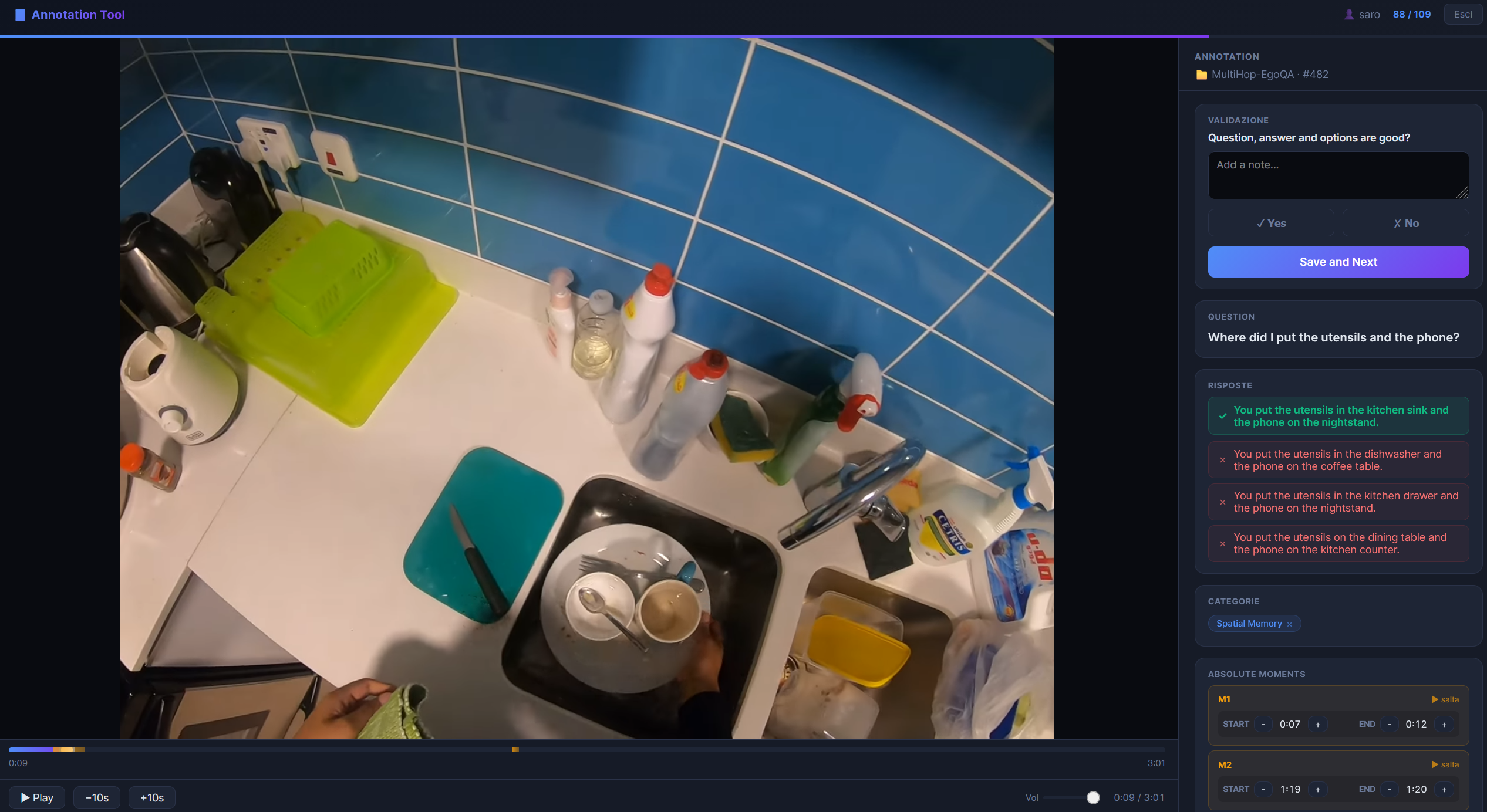}
        \caption{Validation view showing the video, the question, answer options, and evidence moments.}
    \end{subfigure}
    \hfill
    \begin{subfigure}{0.49\linewidth}
        \centering
        \includegraphics[width=\linewidth]{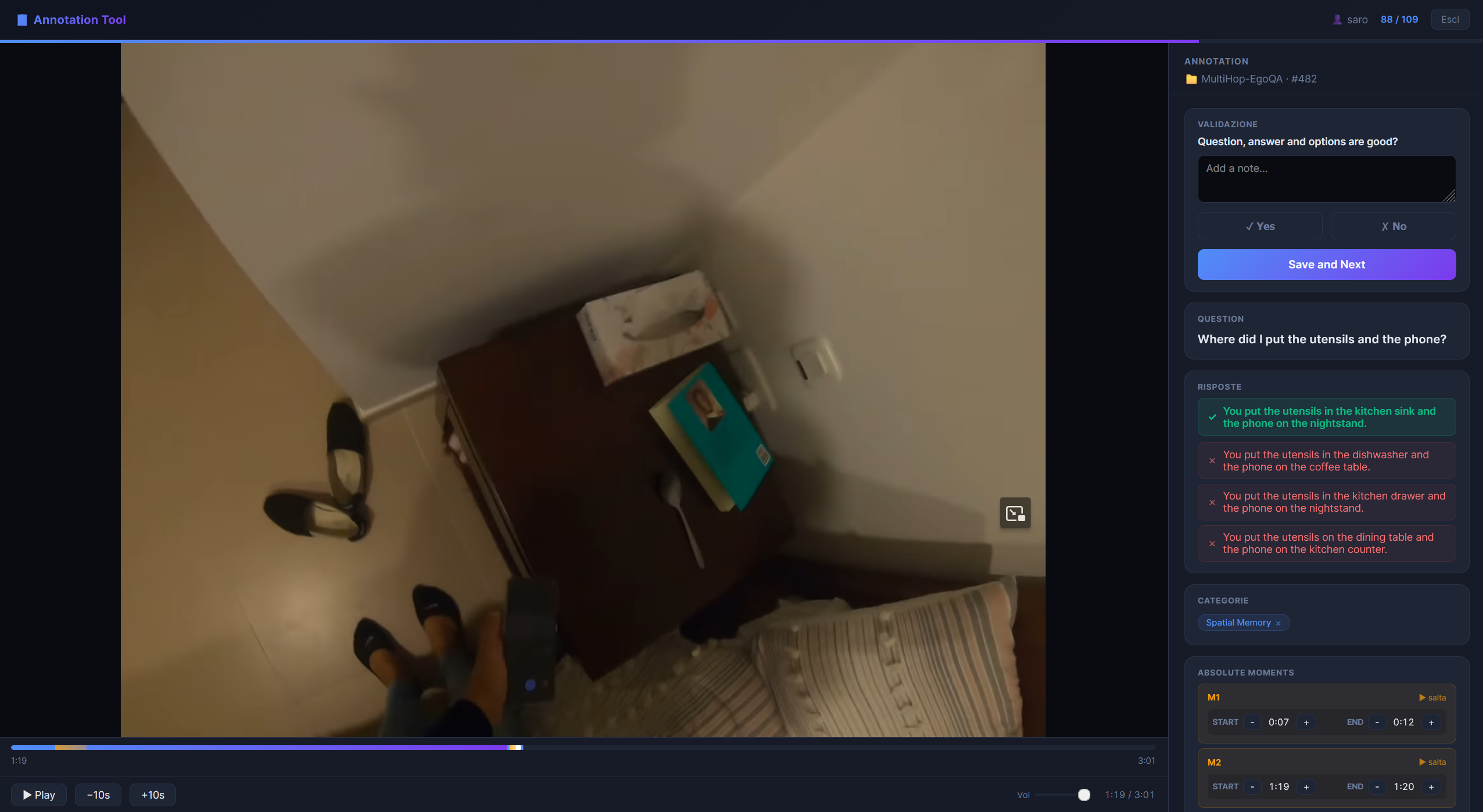}
        \caption{Example of the same validation item at a different point in the evidence window.}
    \end{subfigure}
    \caption{
    Human validation interface used to review candidate QA samples. Validators checked whether the question, correct answer, distractors, and temporal evidence were coherent with the video.
    }
    \label{fig:validation_interface}
\end{figure}

Figure~\ref{fig:validation_interface} shows the validation interface. The task was intentionally simple: validators decided whether each sample should be kept or discarded, while viewing the relevant temporal evidence and all answer options. This setup collected both binary quality labels and qualitative feedback on ambiguous questions, incorrect answers, weak distractors, missing visual evidence, or samples that did not clearly test memory. To ensure consistency, annotators received written instructions and example validations before starting. All reviewers participated voluntarily in the human validation process and received no compensation.

Overall, \textbf{24 validators} contributed to the validation process, leading to the removal of approximately \textbf{11\%} of the reviewed annotations, leading our benchmark to 2,335 samples. The free-text notes were useful for identifying recurring quality issues. Discarded samples typically fell into a few common cases: the correct answer was not clearly visible in the video, the question and answers were not well aligned, multiple answer options could be considered correct, distractors were too easy or unrealistic, the question could not be evaluated at multiple recall times, or the question did not actually test memory. Samples marked as invalid were removed from the benchmark.

Figure~\ref{fig:discarded_validation_examples} shows four examples of discarded items. These examples illustrate different failure modes identified during validation, including questions that were not suitable for multiple recall times, questions asking for future predictions rather than memory recall, ambiguous answer choices, and visually unsupported or weakly grounded annotations.

\begin{figure}[t]
    \centering
    \small

    \begin{subfigure}[t]{0.48\linewidth}
    \begin{tcolorbox}[
        height=0.275\textheight,
        valign=top,
        colback=red!3,
        colframe=red!55!black,
        title=\textbf{Invalid recall formulation},
        fonttitle=\bfseries,
        boxrule=0.7pt,
        arc=2mm,
        left=1.5mm,
        right=1.5mm,
        top=1mm,
        bottom=1mm
    ]
    \textbf{Dataset:} EgoLife

    \vspace{0.25em}
    \textbf{Question:} \emph{When was the meat bought?}

    \vspace{0.25em}
    \textbf{Correct answer:} \emph{Yesterday afternoon}

    \vspace{0.25em}
    \textbf{Distractors:}
    \emph{This afternoon; This morning; Yesterday morning}

    \vspace{0.25em}
    \textbf{Reason for removal:}
    The answer depends on the temporal reference point from which the question is asked, so it cannot be evaluated at multiple recall times.
    \end{tcolorbox}
    \end{subfigure}
    \hfill
    \begin{subfigure}[t]{0.48\linewidth}
    \begin{tcolorbox}[
        height=0.275\textheight,
        valign=top,
        colback=red!3,
        colframe=red!55!black,
        title=\textbf{Does not test memory},
        fonttitle=\bfseries,
        boxrule=0.7pt,
        arc=2mm,
        left=1.5mm,
        right=1.5mm,
        top=1mm,
        bottom=1mm
    ]
    \textbf{Dataset:} EgoTempo

    \vspace{0.25em}
    \textbf{Question:} \emph{What will the person likely do next after grabbing the envelope?}

    \vspace{0.25em}
    \textbf{Correct answer:} \emph{They will likely exit the car.}

    \vspace{0.25em}
    \textbf{Distractors:}
    \emph{They will likely start the engine; They will likely put on their seatbelt; They will likely check the rearview mirror.}

    \vspace{0.25em}
    \textbf{Reason for removal:}
    The question asks for a likely future action rather than requiring recall of information already observed in the video.
    \end{tcolorbox}
    \end{subfigure}

    \vspace{0.55em}

    \begin{subfigure}[t]{0.48\linewidth}
    \begin{tcolorbox}[
        height=0.365\textheight,
        valign=top,
        colback=red!3,
        colframe=red!55!black,
        title=\textbf{Ambiguous answer options},
        fonttitle=\bfseries,
        boxrule=0.7pt,
        arc=2mm,
        left=1.5mm,
        right=1.5mm,
        top=1mm,
        bottom=1mm
    ]
    \textbf{Dataset:} Multi-Hop EgoQA

    \vspace{0.25em}
    \textbf{Question:} \emph{What did I do with the booklet after picking it up from the stand?}

    \vspace{0.25em}
    \textbf{Correct answer:} \emph{After picking the booklet from the stand and carrying it to the seat, you read the booklet.}

    \vspace{0.25em}
    \textbf{Distractors:}
    \emph{After picking the booklet from the stand and carrying it to the seat, you flipped through the booklet; After picking the booklet from the stand and carrying it to the seat, you tucked the booklet into your bag; After picking the booklet from the stand and carrying it to the seat, you briefly examined the cover of the booklet.}

    \vspace{0.25em}
    \textbf{Reason for removal:}
    The distinction between reading and flipping through the booklet is too subtle to define a reliable single correct answer.
    \end{tcolorbox}
    \end{subfigure}
    \hfill
    \begin{subfigure}[t]{0.48\linewidth}
    \begin{tcolorbox}[
        height=0.365\textheight,
        valign=top,
        colback=red!3,
        colframe=red!55!black,
        title=\textbf{Question--answer mismatch},
        fonttitle=\bfseries,
        boxrule=0.7pt,
        arc=2mm,
        left=1.5mm,
        right=1.5mm,
        top=1mm,
        bottom=1mm
    ]
    \textbf{Dataset:} HD-EPIC

    \vspace{0.25em}
    \textbf{Question:} \emph{Why did the person open the packaging?}

    \vspace{0.25em}
    \textbf{Correct answer:} \emph{By pulling the plastic in opposite direction using both hands.}

    \vspace{0.25em}
    \textbf{Distractors:}
    \emph{By unrolling the bag opening; By unfolding the edges of the packaging holding the cube with the right hand and unfolding with the left hand; By undoing the knot on the top of the packaging using both hands.}

    \vspace{0.25em}
    \textbf{Reason for removal:}
    The question asks why the action was performed, but the answer options describe how the packaging was opened.
    \end{tcolorbox}
    \end{subfigure}

    \caption{
    Examples of discarded candidate QA samples. Human validation helped identify samples that were not suitable for the streaming memory protocol, including questions with unstable temporal references, future-prediction questions, ambiguous answer choices, and question--answer mismatches.
    }
    \label{fig:discarded_validation_examples}
\end{figure}

In addition to filtering invalid annotations, human validation confirmed that the retained samples covered a diverse set of memory demands and answer formats. Figure~\ref{fig:valid_qa_examples} reports representative valid QA pairs from the different source datasets. Each example includes the primary memory label, the correct answer, and the negative options used in the multiple-choice setting.

\begin{figure}[t]
\centering
\scriptsize

\newcommand{\validqabox}[6]{%
\begin{tcolorbox}[
    colback=green!4,
    colframe=green!45!black,
    boxrule=0.55pt,
    arc=1.5mm,
    left=1.2mm,
    right=1.2mm,
    top=0.8mm,
    bottom=0.8mm,
    width=\linewidth,
    height=#6,
    valign=top
]
\raggedright
\textbf{#1} \hfill {\scriptsize\textsc{primaryLabel: #2}}

\vspace{0.15em}
\textbf{Q:} \emph{#3}

\vspace{0.15em}
\textbf{Correct:} #4

\vspace{0.15em}
\textbf{Negatives:} #5
\end{tcolorbox}
}

\begin{tabularx}{\linewidth}{@{}X@{\hspace{0.04\linewidth}}X@{}}

% Ego4D row
\validqabox
{Ego4D}
{SPAT}
{Where did I put a pair of scissors?}
{On the table.}
{Left of the stove; on the floor; in the drawer.}
{0.105\textheight}
&
\validqabox
{Ego4D}
{DETL}
{What color was the spanner handle?}
{Black and red.}
{Black and yellow; black and white; blue and red.}
{0.105\textheight}
\\[-0.35em]

% EgoLife row
\validqabox
{EgoLife}
{PROS}
{What are we planning to eat for lunch today?}
{Dumplings.}
{Hot Pot; Toast; Pizza.}
{0.105\textheight}
&
\validqabox
{EgoLife}
{SOC}
{What does Alice not like to eat?}
{Steamed Bun.}
{Pizza; Rice; Noodles.}
{0.105\textheight}
\\[-0.35em]

% EgoTempo row
\validqabox
{EgoTempo}
{DETL}
{What does the person use to stir the ham?}
{Chopsticks.}
{A metal spatula; a wooden spoon; a large serving fork.}
{0.105\textheight}
&
\validqabox
{EgoTempo}
{EVNT}
{What object does the person move from the stovetop to the sink?}
{A frying pan.}
{A stainless steel pot; a glass lid; a small saucepan.}
{0.105\textheight}
\\[-0.35em]

% \rev{Multi-Hop EgoQA} row
\validqabox
{Multi-Hop EgoQA}
{SPAT}
{Where did I put the utensils and the phone?}
{You put the utensils in the kitchen sink and the phone on the nightstand.}
{You put the utensils in the dishwasher and the phone on the coffee table; you put the utensils in the kitchen drawer and the phone on the nightstand; you put the utensils on the dining table and the phone on the kitchen counter.}
{0.145\textheight}
&
\validqabox
{Multi-Hop EgoQA}
{SPAT}
{Where did I drop the garment and the yarn during the video?}
{You dropped the garment on your legs and the yarn on a table.}
{You dropped the garment on the chair and the yarn on the floor; you dropped the garment on the bed and the yarn on the table; you dropped the garment on the couch and the yarn on the shelf.}
{0.145\textheight}
\\[-0.35em]

% HD-EPIC row
\validqabox
{HD-EPIC}
{CAUS}
{What was the reason the person shook the bag?}
{To remove the pieces of onion skin into the food bin.}
{So that all of the coffee beans come out and go into the grinder; so that all of the raisins come out; to mix in the salt; to open the bag fully.}
{0.145\textheight}
&
\validqabox
{HD-EPIC}
{CAUS}
{What was the reason the person tapped the phone keyboard?}
{To make sure the nutritional value of the spice is correctly recorded.}
{To find the couscous measurements in the phone app; to record measurements of salad; to drop food residue attached to the spatula on the frying pan; to drop the remaining soup from the ladle into the container.}
{0.145\textheight}

\end{tabularx}

\caption{
Examples of valid QA pairs included in the benchmark. Each card reports the source dataset, the primary memory label, the question, the correct answer, and the negative options. The examples illustrate the diversity of memory demands covered by the benchmark.
}
\label{fig:valid_qa_examples}
\end{figure}

\newpage
\newpage

\subsection{Question Categorization Protocol and Labeling Guidelines}
\label{app:annotation_protocol}

To ensure the high-quality assignment of memory categories across \egostream, we utilized an iterative Human-in-the-Loop (HITL) pipeline. Crucially, the semantic categorization of each question was performed using only the text of the question and its corresponding answer. This ensured that the taxonomy remained strictly an evaluation of the cognitive demand posed by the query itself, completely independent of the visual evidence. This section details the workflow, the explicit decision rules for resolving overlapping categories, and the guidelines provided to both the LLM and human annotators.

\subsubsection{Iterative Human-in-the-Loop (HITL) Workflow}
\begin{figure}
    \centering
    \includegraphics[width=1\linewidth]{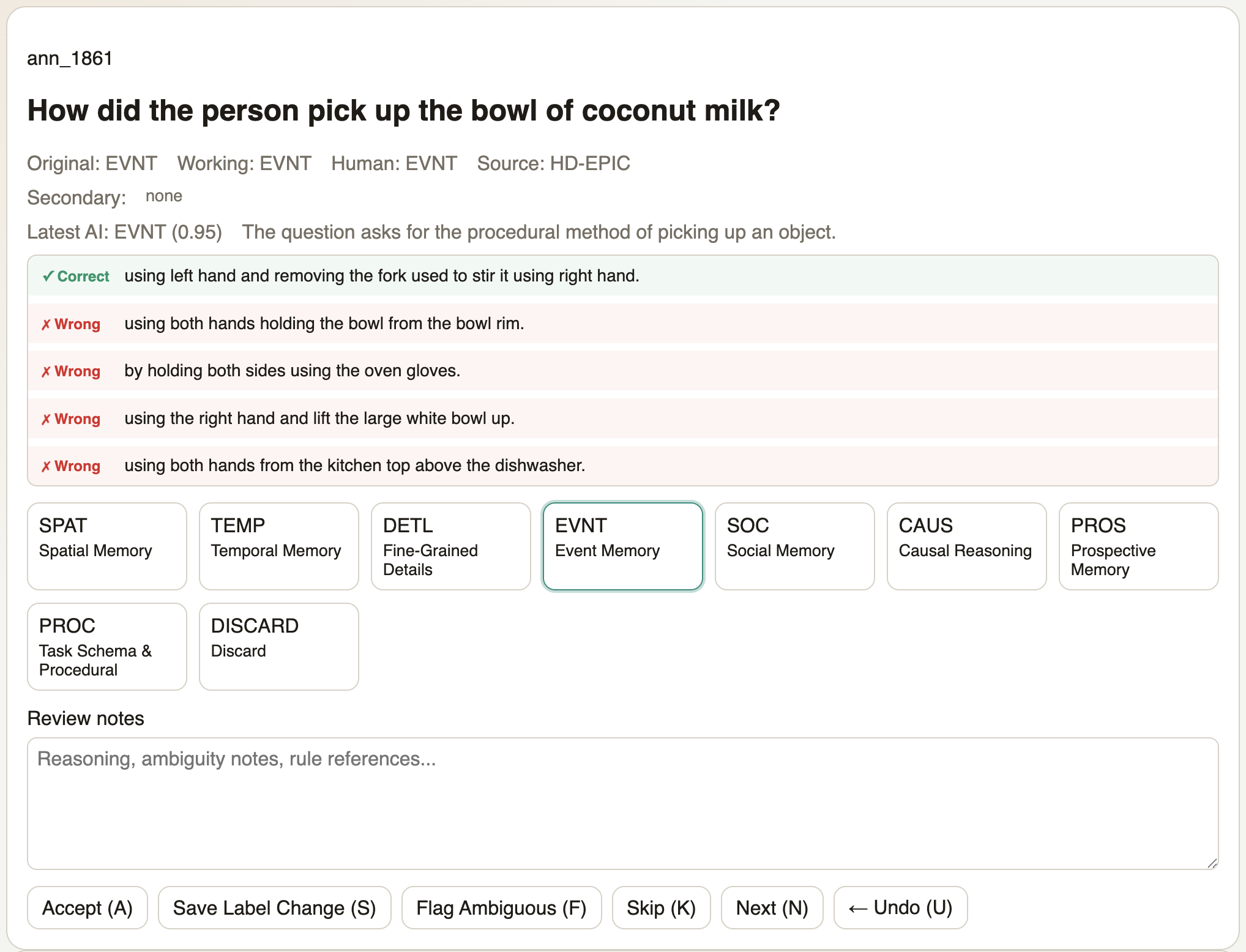}
    \caption{Interface used to refine category annotations for questions. The interface allowed to accept the automated labels, or change them, including a textual description.}
    \label{fig:refine_interface}
\end{figure}

Because episodic queries frequently exhibit overlapping cognitive demands, we assigned each question exactly one \textit{primary} label (representing the core cognitive target) and zero or more \textit{secondary} tags (capturing additional contextual dimensions). The annotation process followed an iterative cycle to scale efficiently while guaranteeing human-level accuracy:

\begin{enumerate}
    \item \textbf{Initial LLM Prediction (round 1):} A Large Language Model (Gemini 2.5 Pro) was prompted to assign initial memory categories based on early rule drafts. The used prompt is reported in Prompt~\ref{prompt:question_categories_v1}.
    \item \textbf{Manual Correction (round 1):} Human expert annotators then manually reviewed and corrected a large subset of these predictions ($847$ questions). During this phase, critical ambiguities were identified and resolved (e.g., distinguishing between a question asking for a spatial location vs. a question using a spatial phrase merely to identify an object). If some questions were considered low quality or malformed, annotators had the ability to discard them. Figure~\ref{fig:refine_interface} shows a screenshot of the interface used for this refinement.
    \item \textbf{Automated Tagging with In-Context Learning (round 2):} The manually corrected, high-quality seed set was then used to craft a refined prompt defining precise cascade rules and include few-shot, in-context examples to make automated tagging more accurate. This guided a subsequent round of LLM labeling (Gemini 2.5 Pro), significantly reducing the initial error rate. In this stage, all questions are assigned a revised category, but original human labels are retained. The prompt is reported in Prompt~\ref{prompt:question_categories_v2}.
    \item \textbf{Full Human Verification (round 2):} At this stage, human annotators used the interface introduced in Figure~\ref{fig:refine_interface} to 1) revise old questions where human and AI disagreed and 2) revise new questions with no human labels. At this stage, labeling is faster, as annotators just have to confirm automated labels in most cases.
\end{enumerate}

\begin{figure}
    \centering
    \includegraphics[width=1\linewidth]{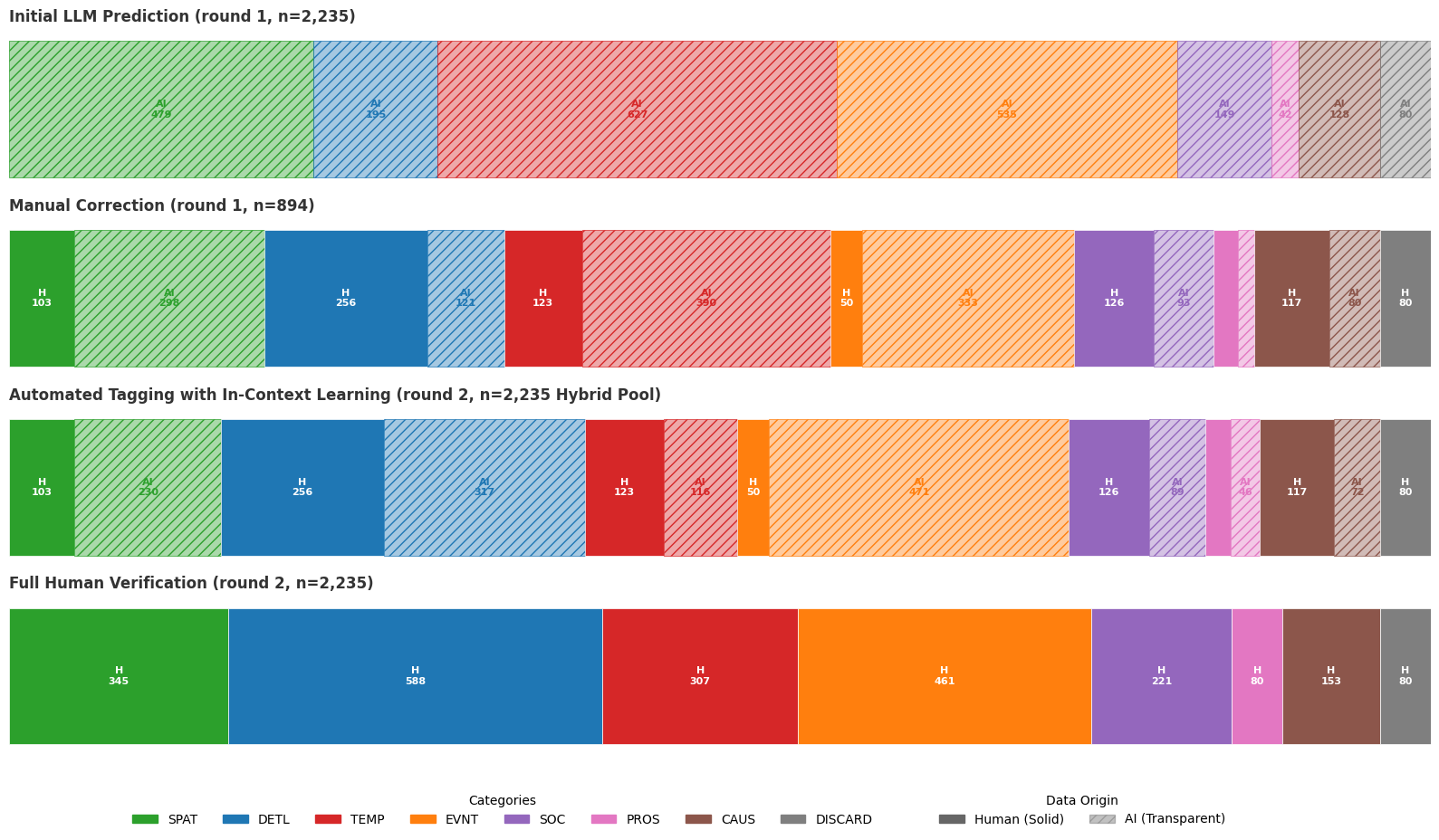}
    \caption{Evolution of AI-generated and human labels in the HITL process.}
    \label{fig:hitl_process}
\end{figure}

Figure~\ref{fig:hitl_process} shows the evolution of AI-generated and human labels obtained with the described process. The process starts with $2,335$ questions. By the end of the process, $85$ questions were discarded, leaving with a set of $2,250$ questions.

\subsubsection{Primary Labeling: The Priority Cascade}
The process described above aims to obtain primary labels isolating the single most dominant cognitive demand of a question. To resolve overlaps consistently during both LLM prompting and human review, we formalized a strict deterministic priority cascade. Annotators evaluated the question text against the following list in order, stopping at the first match:

\begin{enumerate}
    \item \textbf{\texttt{DISCARD} (Malformed/Generic):} Exits the cascade if the question is out-of-scope or asks for generic procedural knowledge (e.g., \textit{``How do I usually buy tickets?''}) rather than episodic recall.
    \item \textbf{\texttt{CAUS} (Causal):} Applies if the question explicitly asks for a reason or explanation (\textit{``Why...?''}).
    \item \textbf{\texttt{PROS} (Prospective):} Applies if the target is a future-oriented plan or intended next action, even if objects or people are involved.
    \item \textbf{\texttt{SOC} (Social):} Applies if the expected answer is a person or group (\textit{``Who...?''}). Social targets supersede temporal or spatial context.
    \item \textbf{\texttt{TEMP} (Temporal):} Applies if the core target is time, ordering, or a sequence (\textit{``When...?'', ``In what order...?''}). Time-anchored states (\textit{``before I picked it''}) default to \texttt{TEMP} unless the query explicitly asks for a person.
    \item \textbf{\texttt{SPAT} (Spatial):} Applies when the object is known, and the target answer is a physical location (\textit{``Where is X?''}).
    \item \textbf{\texttt{DETL} (Detail):} Applies when the question asks for fine-grained content (identity, color, count). \textit{Crucial Tie-Break:} If a question uses a location purely to index an object (e.g., \textit{``What did I put in the microwave?''}), the label is \texttt{DETL}, not \texttt{SPAT}, because the expected answer is an object identity.
    \item \textbf{\texttt{EVNT} (Event):} Serves as the residual category for questions probing action gist, state changes, or execution (\textit{``How did I do X?''}, \textit{``Did Y occur?''}) that do not cleanly fit the more specific categories above.
\end{enumerate}

\subsubsection{Secondary Labeling}
To capture the multidimensional nature of episodic memory, zero or more \textit{secondary} tags were assigned after the primary label was fixed. These were automatically assigned with an LLM (Gemini 2.5 Pro) using Prompt~\ref{prompt:second_tags} and then manually refined. Unlike primary labels, secondary tags do not use a priority cascade or tie-breaking rules. Any cognitive dimension that is explicitly required by the question text—other than the primary label—is tagged.

The following examples illustrate how primary and secondary labels are assigned across different source datasets:

\begin{itemize}
    \item \textbf{Ego4D} – \textit{``Where did I put a pair of scissors?''} 
    $\rightarrow$ \textbf{Primary:} \texttt{SPAT} (target is the final location of the scissors). 
    \textbf{Secondary:} \texttt{EVNT} (location is defined relative to a past placement action).

    \item \textbf{Ego4D} – \textit{``What did I pour in the dish?''} 
    $\rightarrow$ \textbf{Primary:} \texttt{DETL} (fine-grained identity of the poured content). 
    \textbf{Secondary:} \texttt{SPAT} (tied to a specific container), \texttt{EVNT} (embedded in a pouring event).

    \item \textbf{Ego4D} – \textit{``how many planks were on the floor?''} 
    $\rightarrow$ \textbf{Primary:} \texttt{DETL} (object count). 
    \textbf{Secondary:} \texttt{SPAT} (planks are defined by their location ``on the floor'').

    \item \textbf{Ego4D} – \textit{``what did I put in the wood plank?''} 
    $\rightarrow$ \textbf{Primary:} \texttt{DETL} (object inserted into the plank). 
    \textbf{Secondary:} \texttt{EVNT}, \texttt{SPAT} (object detail queried within a spatially anchored manipulation event).

    \item \textbf{EgoLife} – \textit{``Who was with me when I visited the apartment?''} 
    $\rightarrow$ \textbf{Primary:} \texttt{SOC} (person identity). 
    \textbf{Secondary:} \texttt{TEMP} (anchored to a specific visit), \texttt{EVNT} (bound to the visiting event).

    \item \textbf{EgoTempo} – \textit{``What did I do after putting the chopped dill in the food?''} 
    $\rightarrow$ \textbf{Primary:} \texttt{TEMP} (ordering of actions). 
    \textbf{Secondary:} \texttt{EVNT} (reasoning over the surrounding procedure), \texttt{DETL} (implicit reference to the dill and food state).

\end{itemize}

\begin{figure}
    \centering
    \includegraphics[width=0.8\linewidth]{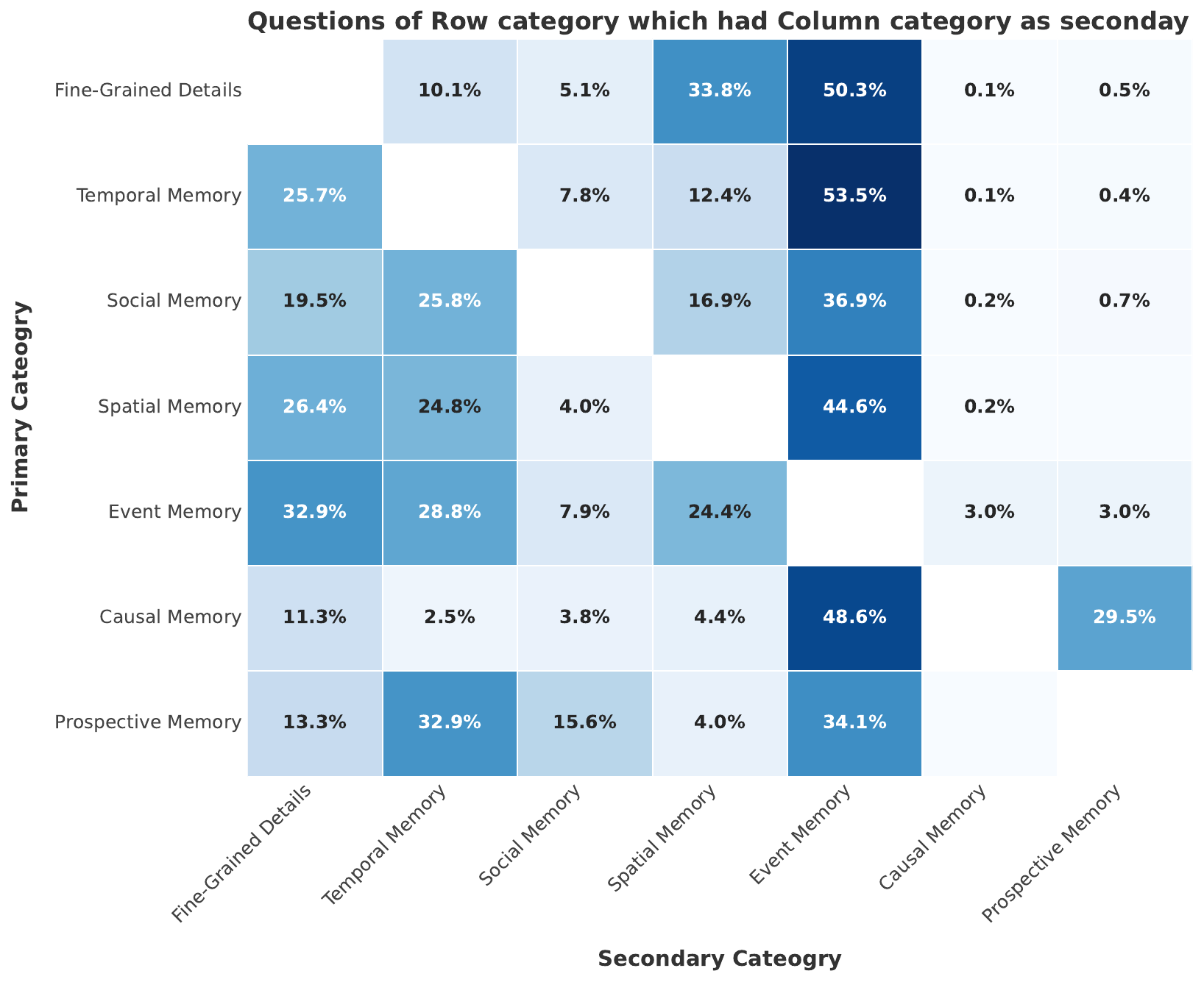}
    \caption{Co-occurrences between primary and secondary labels. Each cell shows the percentage of questions which are primarily marked with the category in the rows, but are also tagged with the category on the column.}
    \label{fig:cooccurrences}
\end{figure}

By decoupling the strict primary classification from the inclusive secondary tags, our annotation protocol supports both clean, mutually exclusive category evaluation and richer, multi-label diagnostic analysis. Figure~\ref{fig:cooccurrences} shows the co-occurrences between primary and secondary labels. As can be noted, most questions do not strictly belong to a single category, with ``Event Memory'' being the most overlapped with other categories as a secondary tag. This is expected, as most properties of episodic memory refer to specific events.

%========================================================
% Compact qualitative examples table helpers.
% Robust version: no landscape, no rotation, no oversized timeline.
%========================================================
\providecommand{\qtag}[2]{%
  \tikz[baseline=(char.base)]\node[
    anchor=base,
    fill=#1!15,
    text=#1!80!black,
    rounded corners=2.5pt,
    inner xsep=2.2pt,
    inner ysep=1.1pt,
    font=\sffamily\bfseries\tiny
  ] (char) {#2};%
}

\providecommand{\qframe}[1]{%
  \begin{tikzpicture}[baseline=-0.5ex]
    \clip[rounded corners=3pt] (0,0) rectangle (1.85,1.05);
    \node[anchor=south west, inner sep=0pt] at (0,0)
      {\includegraphics[width=1.85cm,height=1.05cm,keepaspectratio]{#1}};
    \draw[gray!50, line width=0.35pt, rounded corners=3pt] (0,0) rectangle (1.85,1.05);
  \end{tikzpicture}%
}

\providecommand{\qcorrect}[1]{%
  \textcolor{teal!85!black}{\textbf{#1}}%
}
\setlength{\tabcolsep}{3pt}

\renewcommand{\qframe}[1]{%
  \makebox[1.45cm][c]{%
    \includegraphics[width=1.45cm,height=0.85cm]{#1}%
  }%
}

\begin{table}[!htbp]
\centering
\caption{Compact qualitative examples across memory dimensions. Examples illustrate how questions are assigned primary and secondary memory labels, while also covering different Answer Validity Window (AVW) ranges.}
\label{tab:qualitative_examples_extended}

\scriptsize
\setlength{\tabcolsep}{3pt}
\renewcommand{\arraystretch}{1.15}

\begin{tabularx}{\linewidth}{@{}
>{\centering\arraybackslash}p{1.95cm}
>{\raggedright\arraybackslash}X
>{\raggedright\arraybackslash}p{2.55cm}
>{\centering\arraybackslash}p{1.35cm}
@{}}
\toprule
\textbf{Frame} &
\textbf{Question and correct answer} &
\textbf{Labels / AVW} &
\textbf{Dataset} \\
\midrule

\qframe{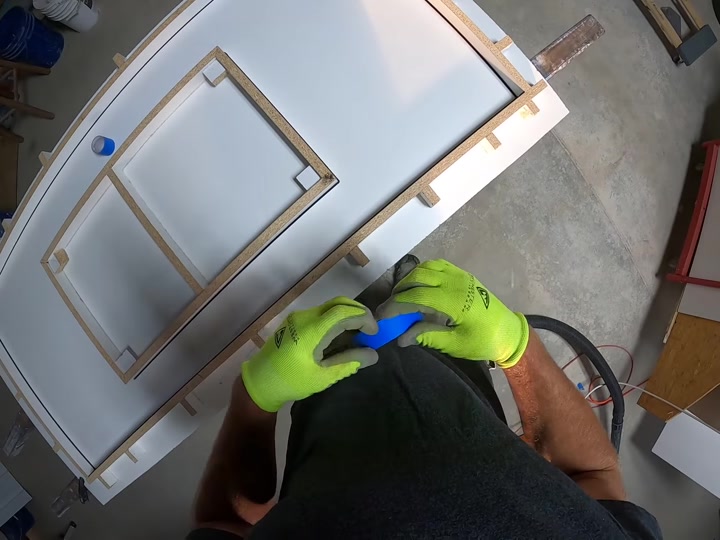} &
\textbf{Q:} Where is the roll of blue tape at the beginning of the video? \newline
\textbf{A:} \qcorrect{The roll of blue tape is on the white wooden board.}
&
\textbf{Primary:} \qtag{blue}{SPAT} \newline
\textbf{Secondary:} \qtag{green}{DETL} \qtag{orange}{TEMP} \newline
\textbf{AVW:} \rev{0s}
&
\qtag{teal}{EgoTempo}
\\[0.55em]

\qframe{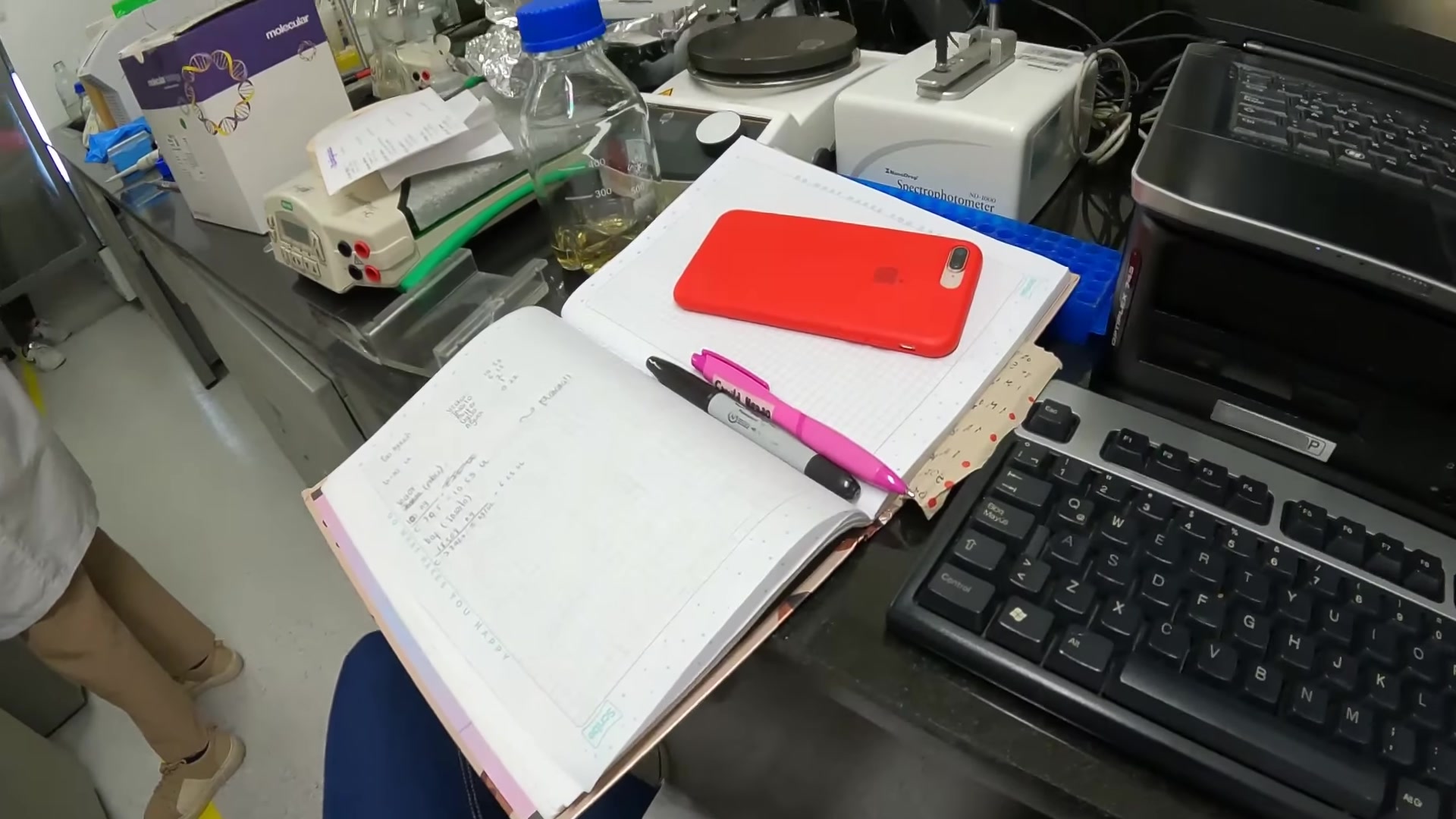} &
\textbf{Q:} Who closed the green specimen tray and the white bottle during the video? \newline
\textbf{A:} \qcorrect{You closed the specimen tray and the white bottle.}
&
\textbf{Primary:} \qtag{magenta}{SOC} \newline
\textbf{Secondary:} \qtag{purple}{EVNT} \qtag{green}{DETL} \newline
\textbf{AVW:} 1s
&
\qtag{pink}{MultiHop}
\\[0.55em]

\qframe{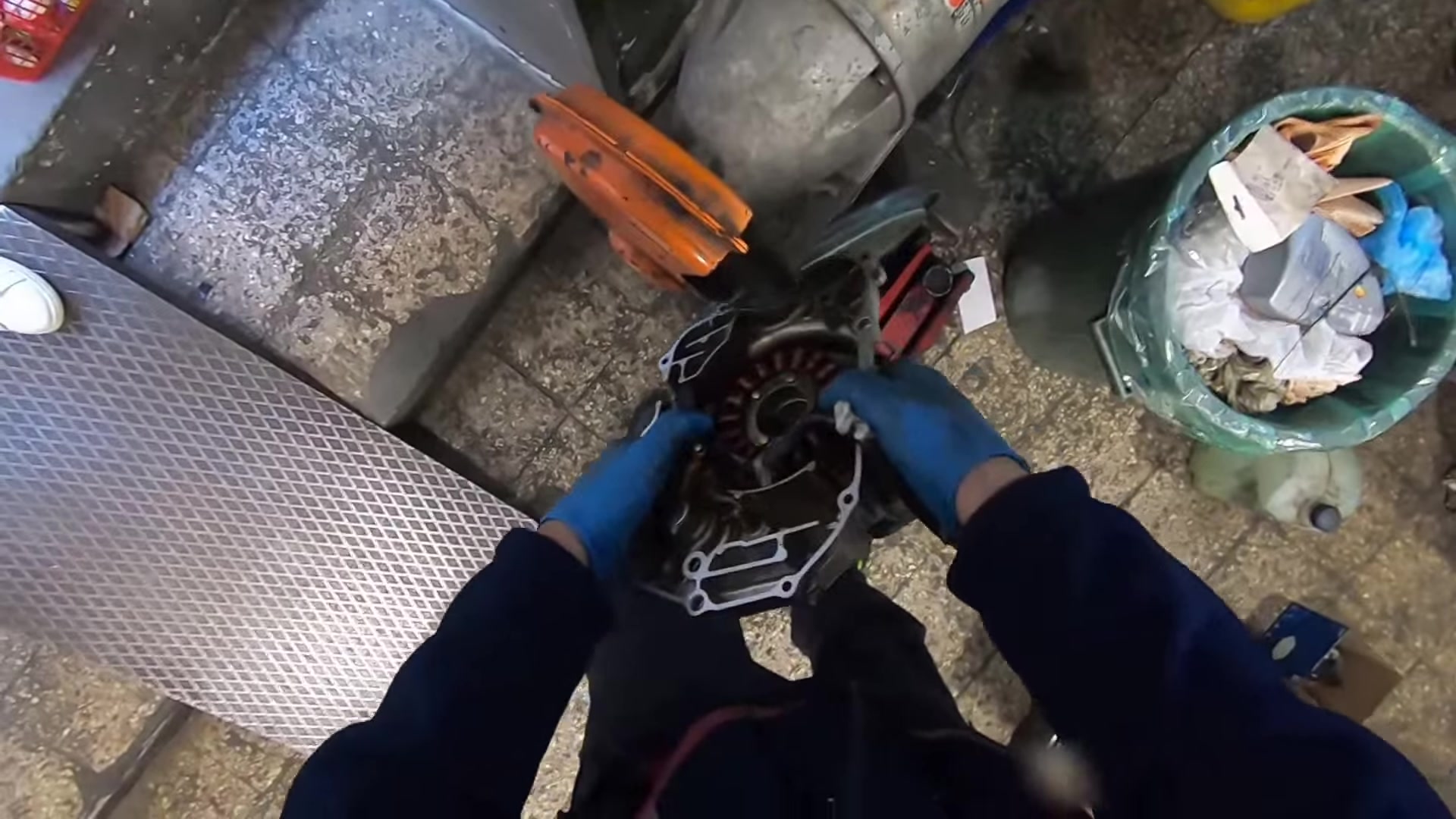} &
\textbf{Q:} Who was with me when I was machining the engine body? \newline
\textbf{A:} \qcorrect{A man with blue pants.}
&
\textbf{Primary:} \qtag{magenta}{SOC} \newline
\textbf{Secondary:} \qtag{orange}{TEMP} \qtag{purple}{EVNT} \newline
\textbf{AVW:} 10s
&
\qtag{violet}{Ego4D}
\\[0.55em]

\qframe{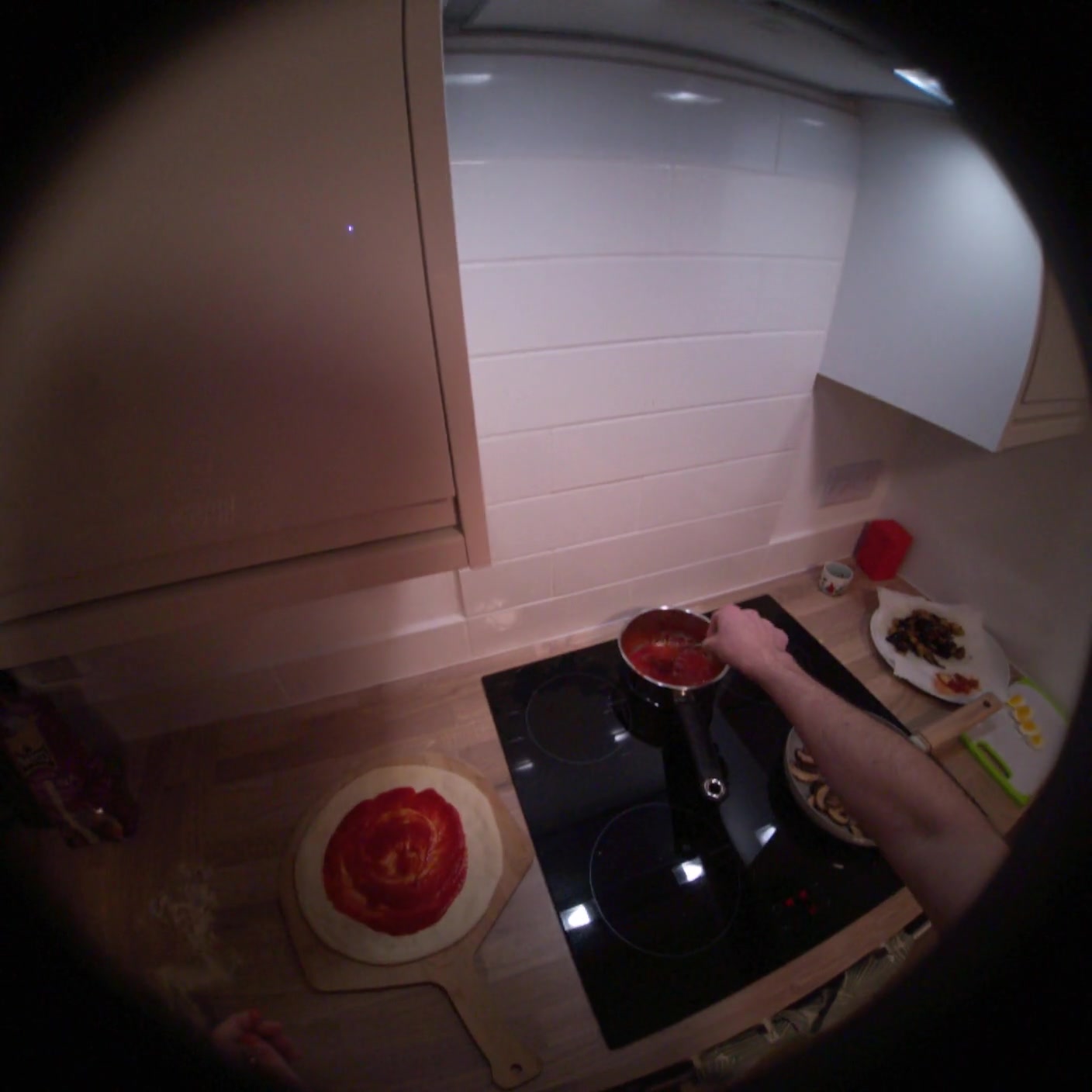} &
\textbf{Q:} How did the person put the spoon? \newline
\textbf{A:} \qcorrect{By slotting the spoon handle into the pocket of an apron.}
&
\textbf{Primary:} \qtag{green}{DETL} \newline
\textbf{Secondary:} \qtag{purple}{EVNT} \newline
\textbf{AVW:} 42s
&
\qtag{brown}{HD-EPIC}
\\[0.55em]

\qframe{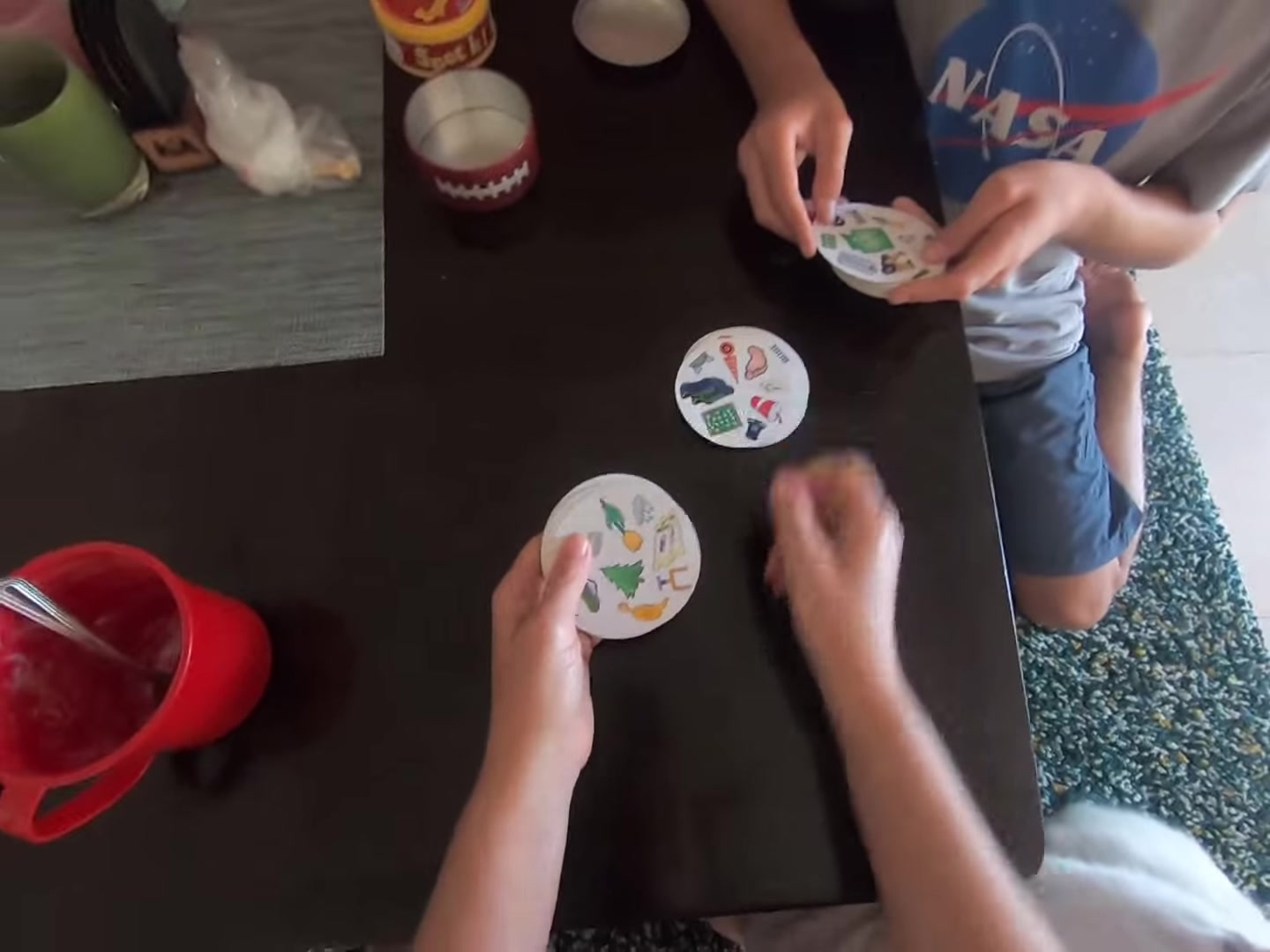} &
\textbf{Q:} Where was the steel spoon before I put the card game instructions in the paper bag? \newline
\textbf{A:} \qcorrect{Red cup.}
&
\textbf{Primary:} \qtag{blue}{SPAT} \newline
\textbf{Secondary:} \qtag{orange}{TEMP} \qtag{purple}{EVNT} \qtag{green}{DETL} \newline
\textbf{AVW:} 2.5m
&
\qtag{violet}{Ego4D}
\\[0.55em]

\qframe{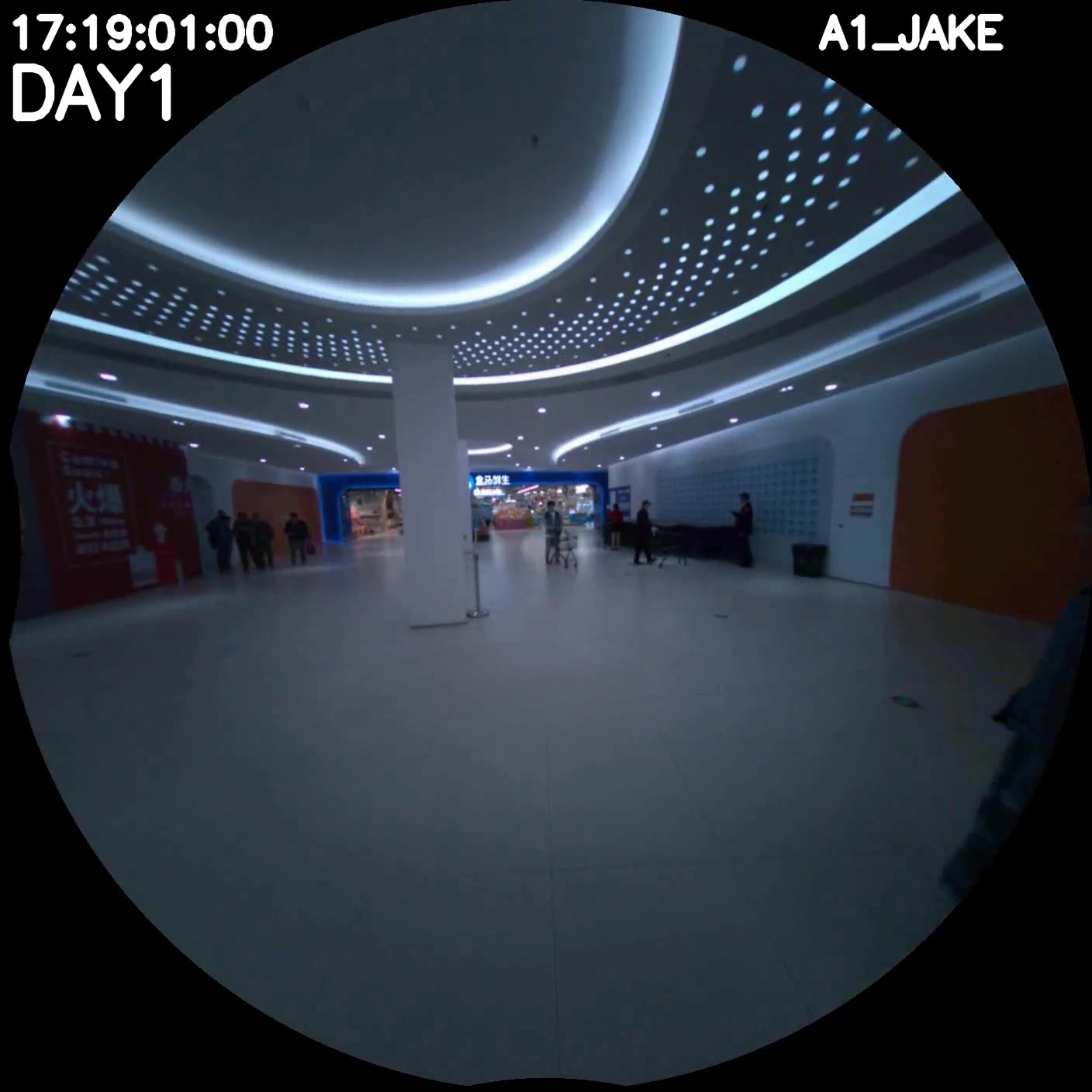} &
\textbf{Q:} Where was I the last time I pushed a cart with a friend? \newline
\textbf{A:} \qcorrect{Hema Supermarket.}
&
\textbf{Primary:} \qtag{blue}{SPAT} \newline
\textbf{Secondary:} \qtag{orange}{TEMP} \qtag{purple}{EVNT} \qtag{magenta}{SOC} \newline
\textbf{AVW:} 13.8h
&
\qtag{olive}{EgoLife}
\\

\bottomrule
\end{tabularx}

\vspace{0.35em}
\raggedright
\tiny
\textbf{Label Legend:}
\qtag{blue}{SPAT} Spatial,
\qtag{green}{DETL} Detail,
\qtag{orange}{TEMP} Temporal,
\qtag{purple}{EVNT} Event,
\qtag{magenta}{SOC} Social,
\qtag{red}{CAUS} Causal,
\qtag{cyan}{PROS} Prospective.

\end{table}

\begin{figure}
    \centering
    \includegraphics[width=0.8\linewidth]{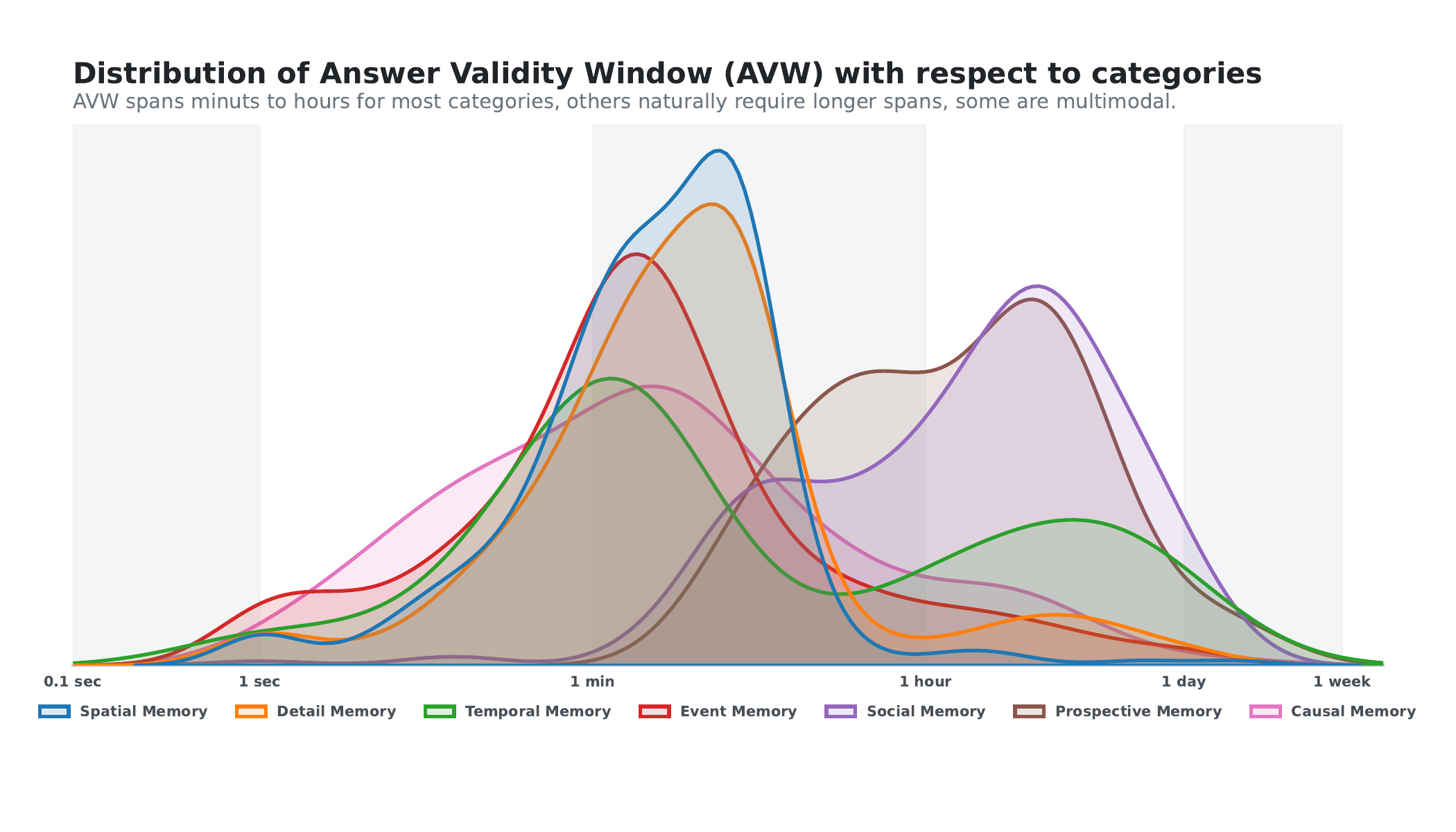}
    \caption{Distribution of Answer Validity Window across Categories.}
    \label{fig:avw_cateogires}
\end{figure}

\begin{PromptBlock}{Initial prompt for assigning primary question category.}
\label{prompt:question_categories_v1}

\begin{PromptText}
# EgoStream Memory Classification Prompt (v1)

## Task

Given a list of egocentric VideoQA questions, assign **exactly one** cognitive category to each question from the taxonomy below.  

Always label the category that reflects the question’s **primary cognitive probe**: the main memory system or reasoning demand the question is designed to stress, not just superficial wording.

***

## General principles

- **Single label**: Each question must receive one and only one category.  
- **Primary probe**: When multiple aspects are present (e.g., time and space), choose the category that captures the *hardest* or most central cognitive requirement.  
- **Ignore surface form**: Do not rely only on wh‑words; e.g., not every “Where” is SPAT, and not every “How many” is TEMP. Use the rules and examples below.  
- **Retrospective vs prospective**: Distinguish questions about **past events** (retrospective) from questions about **remembering to perform an intended future action** (prospective).

***

## Taxonomy Definitions

### 1. Spatial Memory (SPAT)

- **Focus**: Current or last known location of objects or entities in space.  
- **Primary probe**: Accessing the agent’s spatial map—“Where is X now?” or “Where did we leave X (in the last valid state)?”  
- **Typical question forms**:
  - “Where is the screwdriver now?”  
  - “On which side of the sink is the mug?”  
- **Inclusion**:
  - Questions whose answer depends mainly on knowing *where something is* in the environment at the current or last valid time.  
- **Exclusion / tie‑breaker**:
  - If the question explicitly refers to a **relative time anchor** (“before”, “after”, “earlier”), classify as **TEMP**, because the main difficulty is retrieving a prior world state given its time index.

***

### 2. Temporal Memory (TEMP)

- **Focus**: Sequence, relative order, timing, duration, and frequency of **events**.  
- **Primary probe**: Navigating the temporal stream—“When”, “Before/After”, “In what order”, “How long”, “How many times”. 
- **Typical question forms**:
  - “In what order did I open the fridge and the drawer?”  
  - “How long was the stove on before I turned it off?”  
  - “How many times did I go to the kitchen?” (count of **event occurrences**).  
  - “Where was the spanner before I carried it?” (spatial phrase but temporal pivot “before”).  
- **Inclusion**:
  - Questions about **order**, **relative time**, **event frequency**, or **past locations indexed by time**.  
- **Exclusion / tie‑breaker**:
  - Counts of **objects** (“How many eggs did I crack?”) → **DETL**, because this targets high‑fidelity numerical detail rather than temporal structure.  
  - Questions that only ask “When will I do X in the future?” *without* checking whether an intention was remembered/executed are usually better treated as CAUS/EVNT, not PROS.

***

### 3. Fine-Grained Details (DETL)

- **Focus**: High‑fidelity perceptual or specific factual details: visual features, object identity, and counts of objects.  
- **Primary probe**: Can the model recall precise attributes rather than just the gist (color, exact item, count)? 
- **Typical question forms**:
  - “What color was the mug?”  
  - “How many eggs did I crack?” (count of eggs as objects).  
  - “What did I pour into the dish?” (when the answer is a specific substance like “detergent”).  
- **Inclusion**:
  - Questions whose answer is a **specific attribute or identity** (color, shape, brand, material, exact tool type) or **object count**, even if events span time.  
- **Exclusion / tie‑breaker**:
  - If the question is about the **gist of an action** (“What did I do after entering the kitchen?”) rather than attributes of objects, use **EVNT**.  
  - If the question asks “Why” something happened, even if details are involved, use **CAUS**.

***

### 4. Event Memory (EVNT)

- **Focus**: Retrospective recall of the **gist** of actions or episodes (“what happened”), without emphasizing fine detail, spatial layout, temporal ordering, or causal explanation. - **Primary probe**: Can the model recall and summarize what the user did or what occurred at a certain time?  
- **Typical question forms**:
  - “What did I do after entering the kitchen?”  
  - “What happened when I went to the balcony?”  
- **Inclusion**:
  - Questions whose main goal is to retrieve *what* action(s) occurred, at a gist/event level.  
- **Exclusion / tie‑breaker**: If a question primarily targets:
  - **Spatial location** → SPAT  
  - **Temporal order/duration/frequency** → TEMP  
  - **Perceptual details** → DETL  
  - **Social identity or interaction** → SOC  
  - **Causal explanation/motivation** → CAUS  
  - **Remembering to execute a future intention** → PROS  

  then use the more specific category instead of EVNT.

***

### 5. Social Memory (SOC)

- **Focus**: People, identities, and social interactions.  
- **Primary probe**: Who was involved, who did what, what someone said or did.  
- **Typical question forms**:
  - “Who was talking to me in the hallway?”  
  - “What did Anna say before we left?”  
- **Inclusion**:
  - Questions where recognizing or recalling **people** or their **utterances/actions** is central (faces, names, roles, specific interlocutors).  
- **Exclusion / tie‑breaker**:
  - If the main point is **why** a person did something (“Why did Anna leave?”), use **CAUS**.  
  - If the person is incidental and the main cognitive demand is spatial/temporal/detail, use SPAT/TEMP/DETL.

***

### 6. Causal Reasoning (CAUS)

- **Focus**: Inferring causes, motivations, or consequences of actions or states.  
- **Primary probe**: Explaining *why* something happened or what it leads to, not just recalling that it happened.  
- **Typical question forms**:
  - “Why did I go to the kitchen?” (motivation for an action).  
  - “Why did the pot boil over?” (linking previous actions to an outcome).  
- **Inclusion**:
  - “Why…”, “What caused…”, “What was the reason…”, “What was the consequence of…”.  
  - Questions whose correct answer requires integrating multiple observations to infer a cause or intent.  
- **Exclusion / tie‑breaker**:
  - Questions only asking **what happened** or **who did what** without a causal explanation → **EVNT** or **SOC**.

### 7. Prospective Memory (PROS)

- **Focus**: Remembering to perform **planned actions in the future**—linking previously formed intentions to later execution at the right time or cue. 
- **Primary probe**: Did the agent remember and carry out a prior intention?  
- **Typical question forms**:
  - “Did I remember to lock the door when I left, as I planned earlier?”  
  - “Did I take my medicine at 8 pm like I intended?”  
  - “What was I planning to do after finishing this call?” (if the intention was set earlier).  
- **Inclusion**:
  - Questions that explicitly reference a **prior intention** and a **later check** on whether it was remembered/executed.  
- **Exclusion / tie‑breaker**:
  - Questions that merely describe future plans without checking remembering (“What am I going to do later?” asked at planning time only) are better assigned to EVNT/CAUS depending on content.  
  - Questions about past events with no future-intention component belong to TEMP/EVNT/others, not PROS.

***

## Corner Cases & Tie-Breaker Table

| Question | Possible categories | Decision & Rationale |
| :--- | :--- | :--- |
| “What did I pour in the dish?” | EVNT vs DETL | **DETL**: The answer is the **specific identity** of the substance (e.g., detergent), a high‑fidelity object detail rather than just the action “I poured something.” |
| “Where was the spanner before I carried it?” | SPAT vs TEMP | **TEMP**: The key challenge is retrieving a **past state** indexed by “before”, i.e., temporal navigation, even though the answer is a location. |
| “How many eggs did I crack?” | TEMP vs DETL | **DETL**: This is a **count of objects**, not a count of event occurrences or visits, so we treat it as fine-grained detail. |
| “How many times did I go to the kitchen?” | TEMP vs EVNT | **TEMP**: This is a **frequency of events**, which directly probes temporal/event counting. |
| “Why did I go to the kitchen?” | EVNT vs CAUS | **CAUS**: The question asks for the **motivation or reason** for the action, not just what happened. |
| “Did I remember to turn off the stove after deciding to cook pasta later?” | TEMP vs PROS | **PROS**: This is explicitly about whether a **previous intention** was remembered and executed at the appropriate later time. |

***

## Input Format

You will receive as input a json file `input_questions.json`.

***

## Output Format

Return a JSON array `input_questions_automated_round1.json` following the format of `input_questions.json`. Keep only one category per action decided according to the rules above.

Add rationales for your choices and keep them short but specific (1–2 sentences), linking the question to the rules above.
\end{PromptText}
\end{PromptBlock}
\begin{PromptBlock}{Refined prompt for assigning primary question category.}
\label{prompt:question_categories_v2}

\begin{PromptText}
# EgoStream Memory Classification Prompt (v2)

You are given **questions about an egocentric episode** (first‑person video).  
Your task is to assign **exactly one primary category** to each question from this set:

- `DETL` – Fine‑grained details  
- `SPAT` – Spatial memory  
- `TEMP` – Temporal memory  
- `EVNT` – Event memory  
- `SOC` – Social memory  
- `PROS` – Prospective memory  
- `CAUS` – Causal memory  
- `DISCARD` – Out‑of‑scope / malformed

Follow the **decision cascade** below in order. At each step, if a rule matches, assign that label and stop.

---

## 1. DISCARD filter (early exit)

Assign `DISCARD` if:

- The question is malformed, incoherent, or clearly unanswerable from the episode.  
- The question is **generic procedural knowledge** rather than episode‑specific, e.g.:
  - "How do I usually buy plane tickets?"  
  - "What is the best way to scan my house for things I forgot?"

If in doubt, prefer a real category rather than DISCARD; use DISCARD only when the question truly does not fit any memory category or is nonsensical.

**If DISCARD applies, stop.**

---

## 2. CAUS – Causal memory

Assign `CAUS` if the **primary purpose** of the question is to ask for a **reason or explanation**, typically introduced by "why" or "for what reason", and the expected answer is a **cause**, not a procedure.

Examples:

- "Why didn't we buy chestnut kernels?"  
- "Why did we put the durian back?"  

If the question is **"how did I do X?"** and the answer is an **action sequence** (not a reason), this is **EVNT**, not CAUS.

**If CAUS applies, stop.**

---

## 3. PROS – Prospective memory

Assign `PROS` when the question asks about **plans, intentions, or future actions** (things that are supposed to happen or be done).

Typical patterns:

- "What are we planning to eat for lunch?"  
- "What are we going to do tomorrow?"  
- "What does Lucia need to buy to decorate her room?"  
- "Who plans to grow flowers?" (future‑oriented social plan → still PROS)

Prospective trumps other dimensions (including SOC) when the core target is an intended future action or plan.

**If PROS applies, stop.**

---

## 4. SOC – Social memory

Assign `SOC` when the **answer is one or more people** (identity of a person or group), or a stable property of people in the episode.

Use SOC whenever **"who" is the main query**, even if there are temporal or contextual cues.

Examples:

- "Who did I talk to in the workshop?"  
- "Who did I talk to in the room?"  
- "Who did I lose track of at the supermarket on day 1?"  
- "Who was next to me the last time I was taste‑testing something?"  
- "Who usually picks up the takeout?"  
- "Who usually frequents the vegetable market?"

Even if the question mentions time ("last time…", "on day 1…") or place, if the **target** is the person, use `SOC`.

**If SOC applies, stop.**

---

## 5. TEMP – Temporal memory

Assign `TEMP` when the **core target** is **time or order of events**: when something happened, in what sequence, or at which occurrence ("last time", "first time", etc.). The expected answer is a **time**, **temporal position**, or **ordering**.

Typical patterns:

- Explicit when/last time/first time:
  - "When was the meat bought?"  
  - "When was the last time eggs were bought?"  
  - "When did we finish washing the dishes?"

- Order / sequence:
  - "In what order did I open the dishwasher and the container?"  

- Location in a **temporal chain**, where **time is primary**:
  - "Where was the knife before I picked it?" → `TEMP`  
    (You are being tested on recalling an earlier state *indexed by time* "before", not just "where is it now".)

Use this tie‑breaker:

> If the question's main point is **"at what time / in what order / which occurrence"**, label `TEMP`, even if the answer mentions a location or object.

Exceptions (override TEMP):

- If the question is clearly about **who** (person) and "last time" is just context, use `SOC`.  
- If "before/after" is only a weak cue and the **answer space is purely locations**, and the annotator's intent is clearly spatial, `SPAT` can trump TEMP (see SPAT below).

**If TEMP applies, stop.**

---

## 6. SPAT – Spatial memory

Assign `SPAT` when the **target of the question is a location or spatial configuration**. The answer should be **"where"** something is or was, or a surface/region relation ("on", "in", "next to", etc.), not which object it is.

Key pattern:

> The **object is given**, and the question asks **where** it is/was.

Examples:

- "Where is my dreamers yora?"  
- "Where is the stool?"  
- "Where did I put the shoe?"  
- "Where did I put the cabbage?"  
- "Where did I put the chopsticks?"  
- "Where did I put the books?"  
- "Where was my phone?"  
- "In what location did I see the orange couch?"  
- "In what location did I see the toolbox?"  
- "Where did I carry the carton to?"  
- "Where was the AD calcium milk taken from?"  

For "before/after" spatial questions:

- "Where was the candle jar before I operated my phone?" → `SPAT`  
  (Here the practical focus is the location, time just anchors which state.)

Tie‑break rule vs DETL:

- If the **answer is a location** (shelf, table, room, "next to the sink", etc.), choose `SPAT`.  
- If the **answer is an object identity, color, or count**, even if the question contains "in/on/at [location]", choose `DETL` (see next section).

**If SPAT applies, stop.**

---

## 7. DETL – Fine‑grained detail memory

Assign `DETL` when the question asks for **fine‑grained content**:

- Identity of an object  
- Color  
- Number / count  
- Type / category / brand  
- Text written on an object  
- Specific item used/consumed/etc.

Crucially, **DETL includes questions where a location/time is used only as a cue to select the object**, but the real target is the object or its properties.

### 7.1 Typical DETL patterns

**Object identity:**

- "What did I pour in the dish?"  
- "What did I pour in the cup?"  
- "What did I put in the saucepan?"  
- "What did I put in the microwave?"  
- "What did I put in the bucket?" (identity of what was put in)  
- "What vegetable did I slice?"  
- "What ingredient did I pick?"  
- "What tool did I remove from the drawer?"  
- "What cable did I connect to the circular saw?"

**Color / appearance:**

- "What color was the spanner handle?"  
- "What color is the towel?"  
- "What color is the air fryer?"  
- "What color was the machine on the floor?"  
- "What color was the jerrycan on the shelf?"  
- "What color was the wallpaper I picked from the box?"  
- "What color is the washing machine that I touched?"

**Number / count:**

- "How many oven trays did I take from the trolley?"  
- "How many pastries did I put on the tray?"  
- "How many planks were on the floor?"  
- "How many blue bottles did I see?"  
- "How many cables did I pick?"  
- "How many spoons were in the cooking pot?"

**Text / label / semantic content:**

- "What words were written on the box close to the wall?"

**Preferences / stable content facts (when answer is not a person):**

- "What kind of sauce do Shure and I usually have with hot pot?"  
- "What brand of mineral water do we typically buy?"

### 7.2 Location as index → still DETL

If a question uses location or time to specify **which object** we are talking about, but the **answer is still the object identity/color/number**, use `DETL`.

Examples:

- "What color was the cup on the wall?" → DETL  
- "What color was the drum on the floor?" → DETL  
- "What color is the shopping bag on the floor in the room?" → DETL  
- "What color was the cloth on the carton?" → DETL  
- "What did I put in the bucket?" → DETL (identity of object)  
- "What did I pour into the steel bowl?" → DETL

**Rule of thumb for DETL vs SPAT:**

- **If the answer is an object identity, color, or count → DETL.**  
- **If the answer is a location → SPAT.**

### 7.3 DETL vs EVNT

Use `DETL` when the question is *"which specific object/value?"*, even if phrased as "what did I X?":

- "What vegetable did I slice?"  
- "What did I use to stir the soup?" (identity of utensil)  
- "What ingredient did I pick?"  

Use `EVNT` (see next section) when the question is about **what happened in an event or how an action was performed**, not about which object.

**If DETL applies, stop.**

---

## 8. EVNT – Event memory

Assign `EVNT` when the question targets the **event/gist** itself: what happened, how it unfolded, or whether a particular action/state occurred. It is not about fine object details, nor about pure when/where/who.

Typical patterns:

- **How an action was done (procedural within the episode):**
  - "How did the person open the coffee beans bag?"  
  - "How did the person remove the excess water?"  
  - "How did I open the recipe kit yellow tub?"  
  - "How did I take the nails from storage?"

- **Gist of a past episode or outcome:**
  - "What did we end up eating the last time we ordered takeout?"  
    (overall event outcome, not a specific ingredient color/count)  
  - "What items are still not bought?" (status of a plan over time)  

- **Binary / state questions about whether something happened or how:**
  - "Did I leave the door open?"  
  - "Did I close the fridge door after taking the milk?"

Use EVNT as a **residual category** when:

- The question is clearly about **the action or state change itself**, and  
- It does not fit CAUS, PROS, SOC, TEMP, SPAT, or DETL more specifically.

---

## 9. Summary of priority order

Always apply categories in this **strict order**; once matched, stop:

1. `DISCARD` – malformed / out‑of‑scope generic  
2. `CAUS` – why / reason questions  
3. `PROS` – future plans / intended actions  
4. `SOC` – answer is person/people  
5. `TEMP` – when / order / last time (time answer)  
6. `SPAT` – where / spatial relation (location answer)  
7. `DETL` – object identity, color, number, type, text, detailed content  
8. `EVNT` – event gist: what happened, how it was done, or whether it occurred

When in doubt:

- Ask: **What type of thing is the answer?**  
  - Person → SOC  
  - Time / order → TEMP  
  - Location → SPAT  
  - Object identity / color / count / text → DETL  
  - Reason → CAUS  
  - Future plan → PROS  
  - Otherwise → EVNT
\end{PromptText}
\end{PromptBlock}
\begin{PromptBlock}{Prompt to assign secondary tags to questions.}
\label{prompt:second_tags}

\begin{PromptText}
# Secondary Memory Tags for Episodic QA

You are given questions about a specific egocentric episode and a fixed primary label in `primaryLabel`.

Primary labels come from this set:
- `DETL` Fine-grained details
- `SPAT` Spatial memory
- `TEMP` Temporal memory
- `EVNT` Event memory
- `SOC` Social memory
- `PROS` Prospective memory
- `CAUS` Causal memory
- `DISCARD` Out-of-scope or malformed

## Goal
Assign `secondaryLabels`: zero or more additional labels from the same set that also apply.

## Hard constraints
1. Never change the primary label.
2. Secondary labels can be empty.
3. Never include the primary label inside `secondaryLabels`.
4. Use only the question text and primary label.
5. If `primaryLabel` is `DISCARD`, `secondaryLabels` must be empty.
6. Return labels as a set (no duplicates).

## Semantics for secondary labeling
- Add `DETL` when fine-grained item identity/count/text/color is truly required.
- Add `SPAT` when location or spatial relation is essential.
- Add `TEMP` when order/time anchors (before/after/last/first/when) are essential.
- Add `EVNT` when whether/how an action happened is a meaningful dimension.
- Add `SOC` when person identity/co-presence/role is meaningful.
- Add `PROS` when intent/plan/likely next action is meaningful.
- Add `CAUS` when reason/explanation is meaningful.
- Do not add `DISCARD` secondarily.

## Input format
You will receive a JSON array where each item has:
- `id`
- `question`
- `primaryLabel`

## Output format
Return only a JSON array with one object per input item:

[
  {
    "id": "ann_1",
    "secondaryLabels": ["TEMP", "SPAT"]
  }
]

No markdown. No extra keys. No prose.
\end{PromptText}
\end{PromptBlock}
\subsection{Answer Validity Window and Recall Regimes}
\label{app:avw_recall_Regimes}

A key objective of \egostream is to evaluate not only whether a model can answer a question, but also how long the required information remains valid and retrievable in a streaming setting. We therefore associate each question-answer pair with an \emph{Answer Validity Window} (AVW), defined as the duration for which the annotated answer remains valid after the supporting evidence has been observed.

Let $\mathcal{M}=\{M_1,\ldots,M_n\}$ be the set of evidence moments for a question-answer pair, where each moment is either an interval $M_i=[s_i,e_i]$ or a point timestamp represented as a degenerate interval. We define the evidence completion time as
\[
t_{\mathrm{ev}} = \max_i e_i,
\]
i.e., the time at which all evidence required to answer the question has appeared. The Answer Validity Window is then
\[
\mathrm{AVW} = t_{\mathrm{valid}} - t_{\mathrm{ev}},
\]
where $t_{\mathrm{valid}}$ is the last timestamp at which the annotated answer is still valid. For clip-bound sources, $t_{\mathrm{valid}}$ is the last timestamp at which validity can be verified from the observed stream, so the AVW is a conservative bound (Section~\ref{sec:avw_and_regimes}). A query at time $t_q$ is admissible only if
\[
t_{\mathrm{ev}} \leq t_q \leq t_{\mathrm{valid}}.
\]
The recall delay is
\[
\Delta t = t_q - t_{\mathrm{ev}},
\]
Since admissible queries must satisfy $t_{\mathrm{ev}} \leq t_q \leq t_{\mathrm{valid}}$, the set of possible recall delays for a given question is
\[
0 \leq \Delta t \leq \mathrm{AVW}.
\]
Thus, the AVW defines the maximum admissible recall delay: instant recall queries are issued at evidence completion, while the largest possible delayed query occurs at $t_{\mathrm{valid}}$.

The definition of $t_{\mathrm{valid}}$ depends on the temporal structure of each source dataset. For \textbf{EgoLife}~\cite{yang2025egolife}, which provides point-level evidence and a query timestamp, we set $t_{\mathrm{valid}}=t_q$ and compute $\mathrm{AVW}=t_q-t_{\mathrm{ev}}$. For \textbf{EgoTempo}~\cite{plizzari2025egotempo}, clips are centered around the queried action and evaluated immediately after it; thus $t_q=t_{\mathrm{ev}}$ and $\mathrm{AVW}=0$, placing all samples in the instant-recall regime.

For \textbf{Ego4D Episodic Memory VQA}~\cite{baermann2022keys} and \textbf{Multi-Hop EgoQA}~\cite{chen2025multihop}, we use the temporal evidence provided by the source annotations. In the single-evidence case, $t_{\mathrm{ev}}$ is the end of the annotated supporting moment; for Multi-Hop EgoQA, it is the end of the last required evidence segment. Since both datasets are originally designed for offline VideoQA, the answer is assumed to remain valid until the end of the clip. We therefore set $t_{\mathrm{valid}}$ to the final timestamp of the clip.

For \textbf{HD-EPIC}~\cite{perrett2025hdepic}, we derive validity boundaries from the narration stream. We convert narrations into SRT-like timestamped entries, split them into overlapping chunks, and query Gemini~3.1~\cite{gemini} with the memory question and narration text using Prompt~\ref{prompt:hdepic_validity_spans}. The model returns candidate answerable spans
\[
\mathcal{C}=\{M_i=[s_i,e_i]\},
\]
together with supporting narration lines, confidence, and a short evidence description. We keep medium- and high-confidence spans, merge adjacent predictions, and remove duplicates. A candidate span $M_x$ is considered reliable if it overlaps the ground-truth evidence interval $M_{\mathrm{gt}}$ with
\[
\mathrm{IoU}(M_x,M_{\mathrm{gt}}) > 0.5.
\]
To ensure that the ground-truth answer remains the only valid answer during inference, we trim the video between the neighboring answerable spans. If $M_{x-1}=[s_{x-1},e_{x-1}]$ and $M_{x+1}=[s_{x+1},e_{x+1}]$ are the previous and next candidate spans for the same question, the evaluation subclip is defined as
\[
[e_{x-1},\, s_{x+1}].
\]
We then set $t_q$ to the last second of this subclip and compute $\mathrm{AVW}=t_q-t_{\mathrm{ev}}$ as above. This yields an unambiguous validity-preserving clip while keeping recall delay measured from evidence completion.

\paragraph{Audit of the AVW annotations.}\label{app:avw_audit} Human review covered the full benchmark and inspected both the evidence completion time $t_{\mathrm{ev}}$ and the expiration time $t_{\mathrm{valid}}$. Table~\ref{tab:avw_audit} reports, per source, the fraction of questions whose $t_{\mathrm{valid}}$ was adjusted and the mean absolute displacement of the adjusted boundaries. Overall, 98.04\% of the automatically derived windows were kept unchanged. Adjustments concentrated on questions inherited from Ego4D, EgoTempo and Multi-Hop EgoQA.

\begin{table}[h]

\centering
\caption{Audit of the automatically derived Answer Validity Windows. }
\label{tab:avw_audit}
\small
\begin{tabular}{@{}lcc@{}}
\toprule
\textbf{Source} & \textbf{Adjusted AVW (\%)} & \textbf{Mean displacement (s)} \\
\midrule
Ego4D & 7.23 & 120.42 \\
EgoLife & 0.00 & 0.00 \\
EgoTempo & 0.94 & 13.50 \\
HD-EPIC & 0.00 & 0.00 \\
Multi-Hop EgoQA & 0.69 & 36.50 \\
\midrule
Total & 1.96 & 103.07 \\
\bottomrule
\end{tabular}
\end{table}

\begin{PromptBlock}{HD-EPIC narration-based validity span grounding}
\label{prompt:hdepic_validity_spans}

\begin{PromptText}
You are a temporal grounding engine that works ONLY from narrated subtitles (SRT-like text).
Your job is to locate candidate time spans where a given memory question is answerable based on the narration.

You will be given:
- a memory_nlq_question (natural language memory question)
- a narration_srt_chunk (SRT entries with timestamps)

TASK
Identify ALL candidate time spans within the provided narration_srt_chunk where the question is answerable.
A candidate span is valid if the narration in that interval provides enough textual evidence to answer the question.
Return multiple spans if the question is answerable multiple times in the chunk.
If the chunk contains no answerable evidence, return an empty list.

IMPORTANT RULES
- Use ONLY the narration text provided in narration_srt_chunk. Do NOT assume visual information.
- Do NOT invent events or add details not present in the narration.
- Do NOT use or infer any answer choices.
- A span should be as tight as possible, while still covering the evidence needed to answer the question.
- If evidence is spread across adjacent lines, extend the span to include all required adjacent lines.
- Timestamps MUST come from the SRT entries.
- Use the earliest start time and latest end time of the supporting lines.
- supporting_line_idxs must list the SRT indices used as evidence.
- Confidence:
  - "high" = the narration directly provides enough evidence to answer the question
  - "medium" = the narration likely provides enough evidence, but requires mild inference
  - "low" = the narration is only weakly relevant
- If you are unsure, prefer returning no span.

OUTPUT FORMAT
Return exactly ONE valid JSON object and nothing else.
Use this schema:

{
  "spans": [
    {
      "start_sec": <float>,
      "end_sec": <float>,
      "supporting_line_idxs": [<int>, ...],
      "confidence": "low" | "medium" | "high",
      "evidence": "<short quote or paraphrase from the narration>"
    }
  ]
}

If there are no answerable spans in this chunk, return:

{
  "spans": []
}

INPUT:
{
  "question": "{memory_question}",
  "narration_srt_chunk": "{srt_chunk}"
}
\end{PromptText}
\end{PromptBlock}

\subsection{Recall Regimes}
\label{app:recall_regimes}

Once the Answer Validity Windows are assigned, we derive recall-conditioned evaluations by sampling query times within the admissible validity duration. This ensures that delayed queries measure memory retention rather than answer invalidation. If a model fails at a delayed query time, the annotated answer is still valid by construction, so the failure can be interpreted as degradation under increasing temporal distance.

We define recall regimes using a psychology-inspired, approximately logarithmic temporal scale. Our design follows the classical paradigm of probing memory at multiple delays to measure retention and forgetting, dating back to Ebbinghaus' forgetting-curve studies~\cite{ebbinghaus1885,murre2015replication}. It also reflects the distinction between short-lived working-memory processes, typically operating over seconds~\cite{baddeley2003working}, and longer consolidation processes unfolding over minutes, hours, and beyond~\cite{dudai2004memory,moscovitch2006consolidation}.

Rather than querying each regime at a fixed deterministic offset, we sample recall delays from regime-specific Gaussian distributions centered around representative temporal scales. In particular, we use
\[
\begin{aligned}
&\text{Instant recall:} && \Delta t = 0,\\
&\text{Short-term recall:} && \Delta t \sim \mathcal{N}(2.5,(5/6)^2), \quad \Delta t \in [0,5],\\
&\text{Short-mid-term recall:} && \Delta t \sim \mathcal{N}(17.5,(25/6)^2), \quad \Delta t \in [5,30],\\
&\text{Mid-term recall:} && \Delta t \sim \mathcal{N}(105,25^2), \quad \Delta t \in [30,180],\\
&\text{Mid-long-term recall:} && \Delta t \sim \mathcal{N}(990,270^2), \quad \Delta t \in [180,1800],\\
&\text{Long-term recall:} && \Delta t \sim \mathcal{N}(15300,4500^2), \quad \Delta t \in [1800,28800],\\
&\text{Ultra-long-term recall:} && \Delta t = \mathrm{AVW}, \quad \mathrm{AVW} > 28800.
\end{aligned}
\]
Samples are accepted only if they fall within the admissible AVW of the corresponding question; otherwise, the regime is not instantiated for that question. This allows the same base question to contribute to multiple recall regimes when its answer remains valid long enough, while short-lived answers contribute only to early regimes.

These regimes should not be interpreted as rigid cognitive boundaries. They are practical evaluation strata: they follow a memory-motivated temporal progression, provide fine resolution near the evidence moment, and become progressively broader at longer horizons.

\begin{figure}[t]
    \centering
    \includegraphics[width=0.98\linewidth]{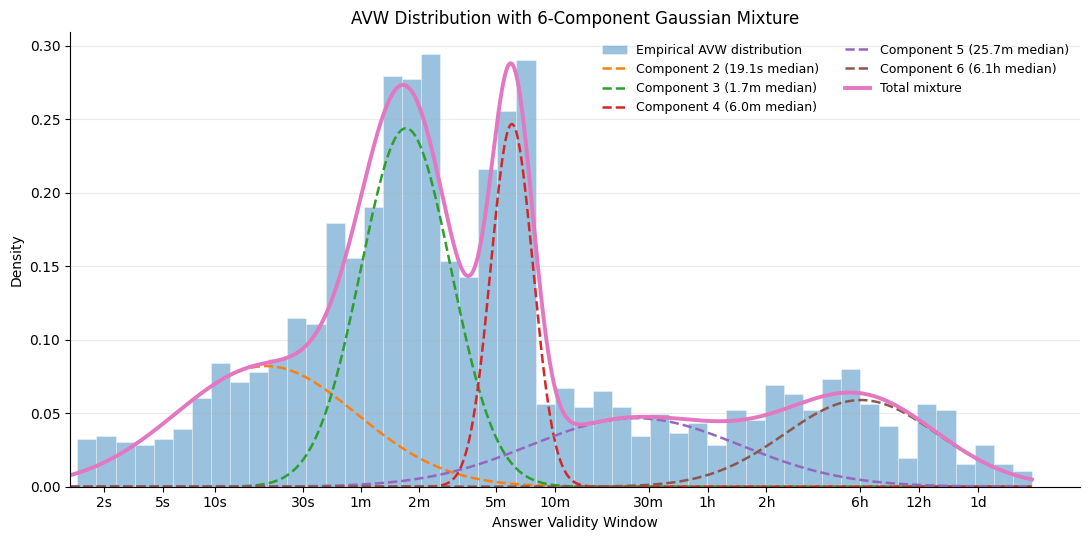}
    \caption{\textbf{Answer Validity Window distribution and Gaussian mixture fit.}
    We plot the empirical distribution of non-zero AVW durations in log-time space and overlay the fitted $6$-component Gaussian Mixture Model. The distribution is concentrated around short validity windows but exhibits a long tail of persistent answers, providing empirical support for recall regimes with fine resolution near the evidence moment and broader coverage at longer delays.}
    \label{fig:recall_regimes}
\end{figure}

To verify that this discretization is also compatible with the empirical structure of the data, we fit a Gaussian Mixture Model (GMM) to the non-zero AVW durations in log-time space shown in 
Figure~\ref{fig:recall_regimes}. The distribution is strongly concentrated near short validity windows, but exhibits a long tail of answers that remain valid over minutes, hours, or longer. The fitted components approximately align with the proposed recall regimes, supporting our use of a logarithmic-like scale while leaving room for practical choices due to dataset composition and annotation biases.

\subsection{Inference Prompts}
\label{app:model_inference}
We use two prompts during inference, corresponding to the two phases of our streaming protocol: stream ingestion and question answering. The ingestion prompt is used before any question is known, and only serves to initialize the model in a streaming-video setting while incoming frames are encoded into the KV cache. The QA prompt is then appended at the target question time and queries the model using the visual context currently stored in the cache. This separation prevents question-conditioned video encoding and ensures that all memory policies are evaluated under the same online setting.
The ingestion prompt defines the streaming context and marks the beginning of the video stream. No answer is generated during this phase (Prompt~\ref{prompt:ingestion_prompt}).
The QA prompt is used after ingestion has reached the desired timestamp. For multiple-choice questions, the model is instructed to output only the selected option index (Prompt~\ref{prompt:qa_prompt}).
\newpage
\begin{PromptBlock}{Ingestion Phase.}
\label{prompt:ingestion_prompt}

\begin{PromptText}

You are experiencing the video from a first-person perspective. Your goal during ingestion is not to write a polished description, but to build a detailed memory that will later be used to answer questions about the video.

Store as much QA-relevant information as possible.
Use first-person narration.
Follow the chronological order of the video.
Do not invent details that are not visible or strongly implied.

Prioritize information useful for future questions:
- Where I am and how the scene is organized.
- What objects are visible and where they are located.
- What objects I interact with.
- Where each object is before I interact with it.
- Where each object is after I interact with it.
- Object attributes such as color, shape, size, material, label, texture, condition, and distinguishing features.
- Object states such as open or closed, full or empty, clean or dirty, broken or intact, switched on or off, covered or uncovered, inside or outside.
- Counts and repetitions, including how many items I see, pick up, place, use, cut, eat, drink, or move.
- The temporal order of events, especially what happens before, during, and after another action.
- People present in the scene, including where they are, what they wear, what they do, and whether I talk to them or work with them.
- Tools, appliances, ingredients, food, containers, drawers, shelves, cabinets, doors, buttons, handles, furniture, vehicles, signs, and landmarks.

For every important object interaction, encode it as a QA memory:
I take [object] from [old location].
I put [object] in/on/under/beside/behind/next to [new location].
I use [tool/object] to [action] [target].
I open or close [object], leaving it [state].
I see [number] [objects] at/in/on [location].
I interact with [person/role] at/in [location].

Avoid vague summaries such as "I prepare food" or "I work on something".
Instead, preserve the concrete objects, locations, quantities, states, and action order that would allow later question answering. Remember as much as possible to answer a question.


\end{PromptText}
\end{PromptBlock}
\begin{PromptBlock}{Question Answering Phase.}
\label{prompt:qa_prompt}

\begin{PromptText}

Question: [QUESTION]
Options:
1) [OPTION_1]
2) [OPTION_2]
3) [OPTION_3]
4) [OPTION_4]
Reply with number only.


\end{PromptText}
\end{PromptBlock}
\subsection{Text Baseline}
\label{app:text_baseline}

To assess whether the base QA set can be solved by exploiting textual priors alone, we additionally evaluated a text-only configuration in which the model receives only the question and answer options, without access to the corresponding video (see Table~\ref{fig:text_only}). As shown below, performance is close to random chance across all semantic categories, except Social (34.63\%). These results are reported here only as a sanity check and are not directly comparable with the video-based methods discussed in the main paper, since they do not use the visual modality. Nevertheless, they further support the conclusion that the benchmark cannot be solved reliably by leveraging textual bias alone and instead requires visual grounding.

\begin{table}[h]
\caption{Text-only baseline.}
\label{fig:text_only}
\centering
\small
\begin{tabular}{@{}l l c c c c c c c c@{}}
\toprule
\textbf{Strategy} & \textbf{Configuration} & \textbf{Avg} & \textbf{SPAT} & \textbf{TEMP} & \textbf{DETL} & \textbf{EVNT} & \textbf{SOC} & \textbf{CAUS} & \textbf{PROS} \\
\midrule
Qwen 8B text only & Text-only MCQA & 25.42 & 23.06 & 24.06 & 24.27 & 27.71 & 34.63 & 23.27 & 20.93 \\
\bottomrule
\end{tabular}
\end{table}

\subsection{Multiple-Choice Bias Analysis}
\label{app:mcq_bias}
Our distractor generation pipeline enforces length, syntactic and structural alignment between the correct answer and the distractors (Appendix~\ref{app:negative_generation}); here we quantify the residual biases.

\paragraph{Sensitivity to option order.} In the main evaluation, options are shuffled once with a fixed seed and kept identical across methods. To measure sensitivity to option order, we ran \bPrune\ \bIntraCos\ three times with different shuffling seeds. Table~\ref{tab:shuffle_ci} reports mean accuracy and 95\% bootstrap confidence intervals across runs: the variance is low and the intervals are narrow relative to the gaps discussed in Section~\ref{sec:experiments}, so the comparative rankings are robust to option order.

\begin{table}[h]
\centering
\caption{Stability of \bPrune\ \bIntraCos\ across three option-shuffling seeds: mean accuracy and 95\% bootstrap confidence intervals; Acc is the overall accuracy over all evaluations.}
\label{tab:shuffle_ci}
\resizebox{\textwidth}{!}{%
\begin{tabular}{@{}c cccccccc cccccccc@{}}
\toprule
 & \multicolumn{8}{c}{\textbf{Semantic Categories}} & \multicolumn{8}{c}{\textbf{Temporal Recall Regimes}} \\
\cmidrule(lr){2-9} \cmidrule(lr){10-17}
\textbf{Acc} & \textbf{Avg} & \textbf{SPAT} & \textbf{TEMP} & \textbf{DETL} & \textbf{EVNT} & \textbf{SOC} & \textbf{CAUS} & \textbf{PROS} & \textbf{Avg} & \textbf{Inst} & \textbf{Short} & \textbf{S-Mid} & \textbf{Mid} & \textbf{ML} & \textbf{Long} & \textbf{Ultra} \\
\midrule
48.7$\pm$1.6 & 46.1$\pm$3.7 & 59.4$\pm$3.9 & 54.8$\pm$4.0 & 57.3$\pm$2.9 & 46.5$\pm$3.5 & 30.7$\pm$2.5 & 40.2$\pm$4.8 & 34.0$\pm$4.5 & 40.7$\pm$2.8 & 56.6$\pm$1.9 & 55.3$\pm$2.1 & 52.0$\pm$2.3 & 41.0$\pm$2.2 & 32.3$\pm$2.5 & 28.1$\pm$2.9 & 19.7$\pm$5.5 \\
\bottomrule
\end{tabular}%
}
\end{table}

\paragraph{Option-length and option-position biases.} Across the three shuffled runs, the uniquely longest option (by token count) accounts for 12.07\% of the model selections, while the correct answer is the longest option in 12.41\% of the questions; the two rates match, giving no evidence of a length preference. The distribution of selected positions is 31.98\%, 21.97\%, 23.35\% and 22.70\% for options 1--4: a moderate preference for the first position exists, but choices are broadly spread, and the shuffling analysis above shows that it does not alter the rankings.

\section{Implementation Details}
\label{app:implementation_details}

\subsection{KV-Cache Layout and Streaming Update}
\label{app:cache_layout}

Our benchmark uses a unified KV-cache implementation that exposes the same streaming interface for all compression, offloading, and retrieval variants. For each decoder layer, the active cache is stored as a contiguous sequence of three logical regions:
\[
K,V
=
\big[
K_{\mathrm{sink}},V_{\mathrm{sink}};
K_{\mathrm{mut}},V_{\mathrm{mut}};
K_{\mathrm{hat}},V_{\mathrm{hat}}
\big].
\]
The sink region contains the priming tokens from the system prompt and initial stream instruction. When sink tokens are enabled, this region is fixed after priming and is never modified by pruning, merging, offloading, or retrieval. The mutable region contains historical visual tokens and is the only region modified by memory-management operations. The hat region contains the most recent visual context and is protected from reduction during streaming. Depending on the configuration, the hat size is either a fixed number of tokens or the token count of the most recent frames.

At each streaming step, a batch of incoming frames is encoded by the visual encoder, projected into the language-model embedding space, and inserted into the language model through the standard image-token placeholders. In frame-aware modes, the cache records the visual-token positions before the forward pass and finalizes the boundary after the reduction step. This removes the auxiliary text/template tokens used to route the images through the processor while preserving the true visual-token frame boundaries.

After each append, a forward pass is done, then the cache checks whether the active length exceeds the configured budget. If it does, the layer first splits the cache into sink, mutable, and hat regions; applies the selected reduction pipeline only to the mutable region; and finally reassembles the cache in the same order. The first decoder layer computes the token indices or trace operations used in a reduction step, and the subsequent layers replay the same operations. This keeps the retained token layout consistent across layers while avoiding independent pruning decisions per layer.

\begin{algorithm}[t]
\caption{Streaming cache update}
\label{alg:cache_update}
\DontPrintSemicolon
\KwIn{Incoming frame stream $\{x_t\}$, sink budget $B_{\mathrm{sink}}$, mutable budget $B_{\mathrm{mut}}$, hat budget $B_{\mathrm{hat}}$}
\KwOut{Updated active KV cache}

Initialize sink cache $K_{\mathrm{sink}}, V_{\mathrm{sink}}$ from the priming prompt\;
Initialize mutable cache $K_{\mathrm{mut}}, V_{\mathrm{mut}} \leftarrow \emptyset$\;
Initialize hat cache $K_{\mathrm{hat}}, V_{\mathrm{hat}} \leftarrow \emptyset$\;

\ForEach{incoming frame batch $X_t$}{
    Encode $X_t$ with the visual encoder and projector to obtain visual embeddings\;

    \If{frame-aware pruning or storage offload is enabled}{
        Record frame token counts and visual-token positions before the forward pass\;
    }

    Run the language-model forward pass with the current active cache and the new visual embeddings\;
    Append the newly produced KV states to the right of the cache during the forward pass\;

    \If{frame-aware pruning or storage offload is enabled}{
        Keep only the visual KV states from the newly appended segment\;
        Update frame-boundary metadata\;
    }
    \Else{
        Update frame-token bookkeeping directly\;
    }

    Split the resulting cache into sink, mutable, and hat regions\;

    \If{$|K_{\mathrm{mut}}| > B_{\mathrm{mut}}$}{
        Apply the configured mutable-region reduction pipeline\;

        \If{$|K_{\mathrm{mut}}| > B_{\mathrm{mut}}$}{
            Apply the selected overflow policy to the remaining excess mutable tokens\;
        }
    }

    Reassemble the active cache as
    $\big[K_{\mathrm{sink}},V_{\mathrm{sink}};\,
    K_{\mathrm{mut}},V_{\mathrm{mut}};\,
    K_{\mathrm{hat}},V_{\mathrm{hat}}\big]$\;
}

\Return updated active KV cache\;
\end{algorithm}

\subsection{Memory Budgets and Region Sizes}
\label{app:memory_budgets}

We separate the active cache into three budgets:
\[
B_{\mathrm{active}}
=
B_{\mathrm{sink}}
+
B_{\mathrm{mut}}
+
B_{\mathrm{hat}} .
\]
Here, $B_{\mathrm{sink}}$ is determined by the number of priming tokens retained as sink tokens, $B_{\mathrm{mut}}$ is the live budget for historical visual tokens, and $B_{\mathrm{hat}}$ is the protected recent-context budget. The mutable budget is the main control variable for memory compression. All methods compared under the same setting use the same total active-cache budget and the same sink and hat configuration.

When hat tokens are configured by frames, the hat size is computed dynamically from the tracked token counts of the most recent frames. Otherwise, a fixed token-level hat size is used. In both cases, the hat region is excluded from pruning and offloading. If a cache variant does not use frame-aware pruning, the implementation falls back to token-level bookkeeping while preserving the same active-cache accounting.

\subsection{Intra-Frame Reducers}
\label{app:intra_frame_reduction}

Intra-frame reducers operate independently within each mutable frame. This preserves temporal coverage because candidates are generated with frame boundaries and, for token-wise salience methods, the implementation can enforce a minimum number of retained tokens per frame.

\paragraph{Adjacent cosine redundancy (\bIntraCos).}
For each mutable frame, we compute cosine similarity between adjacent visual-token key states. Candidate adjacent pairs are ranked by similarity. Under pruning, the selected token from each pair is removed. Under merging, the selected token's value state is averaged into the retained entry. Non-overlap constraints prevent one token from participating in multiple pair reductions in the same pass.

\paragraph{Pseudo-attention salience (\bIntraPA).}
The query-agnostic pseudo-attention scorer constructs a normalized context vector from the mean key representation of the tokens in a frame. It then scores tokens by their similarity to this context with a softmax-style normalization. Low-importance candidates are selected for reduction.

\paragraph{Value-norm ranking (\bIntraVN).}
The value-norm reducer scores each visual token by the norm of its value state and removes low-norm tokens. In the current implementation, this reducer is token-wise and supports pruning only. It does not merge low-norm value states into retained tokens.

\paragraph{Sink-guided cross-attention (\bIntraCS).}
The sink-guided reducer uses the cached sink key states as a fixed guidance template. For each mutable frame, it computes a StreamMem-style attention score between the sink guidance keys and the frame's visual key states, aggregates the score across heads and guidance tokens using either mean or max reduction, and removes the lowest-scoring visual tokens.

\subsection{Inter-Frame Reducers}
\label{app:inter_frame_reduction}

Inter-frame reducers compare neighboring temporal blocks while preserving the streaming order. The implementation constructs source--reference pairs from consecutive frames in the mutable region and, when a hat region is present, also compares the last mutable frame against the first hat frame. The source side is always a mutable frame, so only mutable tokens are eligible for removal.

\paragraph{Cross-frame cosine redundancy (\bInterCos).}
For each neighboring frame pair, we compare source-frame visual-token key states against the reference-frame key states. The implementation allows unrestricted matching between tokens across the two frames rather than assuming grid-aligned positions. Each source token receives the maximum cosine similarity to the reference frame, and highly redundant source tokens are selected for removal. A per-source-frame constraint prevents the reducer from removing all tokens from a source frame.

\paragraph{Cross-frame pseudo-attention salience (\bInterPA).}
For pseudo-attention inter-frame scoring, the reference frame is summarized into a normalized context representation, and source-frame tokens are scored against that context. Tokens with the lowest salience are selected from the mutable source frame. In the current implementation, inter-frame reduction supports pruning only; merging is not used for inter-frame steps.

% \subsection{Pipeline Execution and Cross-Layer Replay}
% \label{app:pipeline_replay}
% The pruning pipeline is executed on the first decoder layer and produces a replayable trace. A trace operation records the operation type, the selected pair indices or token indices, and lightweight metadata such as the scorer name and chunk size. Follower layers do not recompute scores. Instead, they replay the same trace on their own mutable KV tensors. This design ensures that all decoder layers keep the same sequence length and token layout after reduction.
% The pipeline repeatedly applies enabled steps until the mutable length is at or below the target budget, or until no step can make progress. If no configured reducer can remove enough tokens and fallback trimming is allowed, the implementation applies a final left trim to the mutable region. When a storage/offload step is present, fallback trimming is disabled at the pipeline level; instead, the remaining overflow is passed to the offload mechanism.

\subsection{Offloading and Quantized Storage}
\label{app:offloading_details}

Disk offloading is implemented as an optional storage step after mutable-region reduction. If the mutable region still exceeds $B_{\mathrm{mut}}$, the offload policy removes the leftmost excess mutable tokens and writes them to disk as a contiguous chunk. Each chunk stores keys, values, frame-length metadata, token counts, and bookkeeping fields such as physical bytes and equivalent uncompressed bytes. The offload step may split a frame if only part of that frame lies in the overflow range, and the remaining live frame lengths are updated accordingly.

The offload store supports two formats. In raw mode, keys and values are stored in their original dtype on CPU. In quantized mode, offloaded keys and values are quantized with symmetric per-tensor int8 quantization and saved together with their scales. This quantization affects only the external offload storage. The active cache used during the forward pass remains in the model dtype, and offloaded chunks are dequantized when loaded for retrieval.

\subsection{Adaptive Head Budgets (\texorpdfstring{\bAdaHead}{ADA-HEAD})}
\label{app:adakv}
Ada-KV~\cite{feng2025adakv} evicts independently in every decoder layer. At overflow, mutable tokens are scored by the attention they receive from the most recent visual queries (query heads pooled onto their KV head); the scores of all heads are flattened and the global top-$B$ entries are kept. Each head keeps its $b_h$ highest-scoring entries in chronological order, stored as a padded tensor with a validity mask. The reducer treats the mutable region as a single pool without per-frame minima, is query-agnostic with respect to the question, and is not combined with similarity reducers, offloading or retrieval.

\subsection{Pre-Prefill Token Compression (STC)}
\label{app:stc}
STC~\cite{wang2026stc} prunes the visual tokens of each incoming frame before prefill, between the projector and the language model (STC-Pruner only, without feature caching). Each token is scored by the weighted sum ($\alpha=0.5$) of its cosine distances from a spatial anchor (the mean of the frame's tokens) and a temporal anchor (the mean of the previous frames' spatial anchors); the top $\lfloor N(1-\rho)\rfloor$ tokens are kept in order, with $\rho=0.75$, i.e., 112 of the 448 tokens of a frame. Neither future frames nor the question are visible to the method; retained tokens then follow the standard \bFIFO\ \bSink\ \bHat\ lifecycle.

\subsection{Inference Protocol and Reproducibility}
\label{app:reproducibility_notes}

All methods use the same video frame order, frame sampling configuration, query times, system prompt, visual encoder, projector, decoder, and generation interface. During streaming ingestion, position IDs are advanced with a global token counter, while attention masks are constructed from the physical cache length plus the current input length.

For each question, the benchmark clones the current cache state, defined as the master cache, into a secondary cache, identically to the master cache. If retrieval is enabled, relevant blocks are retrieved from the disk and loaded in the secondary cache. The master cache remains unchanged and continues to represent the streaming state of the video, while the secondary cache is used to prompt the model with the question. This design ensures that question-time retrieval does not leak information across questions or alter the cache used for later timestamps.

Reducers, offloading, or retrieval never modify sink and hat regions. Unless otherwise specified, all token-selection scores are computed only over mutable visual tokens. Offloaded tokens are counted separately from live tokens in the reported metrics, and the active cache budget refers to the physical live cache consumed by the decoder at inference time. Retrieval does not increase the active-cache budget; it replaces the historical mutable portion of the active cache with the highest-scoring retrieved blocks under the configured retrieval token budget.

\section{Experimental Settings}
\label{app::experimental}
\subsection{Hardware and Configuration}
\label{app:hardware_configuration}

All experiments use the same inference setup. Inferences are run on a single NVIDIA A100 GPU with 64 GB of VRAM, 32 CPU cores, and 64 GB of system RAM. Videos are processed at 1 FPS and resized to a target resolution of 896$\times$512. We use a sliding-window streaming setup with stride 16 and window size 16, corresponding to batches of 16 processed frames, without window overlap. Generation is performed with temperature 0.01 and nucleus sampling with top-$p=0.9$. This fixed configuration ensures that all memory-management strategies are compared under identical input resolution, temporal sampling, batching, and decoding conditions. The backbone study of Table~\ref{tab:backbones} uses the same settings for InternVL~3.5 and Molmo~2: only the vision encoder, projector and decoder change, with frame boundaries taken from each model's visual-token layout.

\subsection{Pre-Prefill Compression and Off-the-Shelf Streaming Models}
\label{app:compression_baselines}
Table~\ref{tab:compression_baselines} reports two results outside the in-cache policies of Table~\ref{tab:controlled_eval}: the \bFIFO\ \bSink\ \bHat\ baseline augmented with STC (Appendix~\ref{app:stc}), which filters tokens before they enter the cache and reaches 47.18\% at 0.23s per frame, above our best similarity-based pruning policy (\bPrune\ \bIntraCos, Row~\ref{row:q3vl8b_sim_prune_intra_cosine}); and LiveCC~\cite{chen2025livecc}, a model trained for streaming video understanding, evaluated at 1~FPS under the same protocol. LiveCC streams faster (0.34s per frame, including the commentary it generates) and retains more at long horizons, but is less accurate on Instant, Spatial and Detail recall; as it uses a different backbone, training recipe and question subset, the comparison is indicative rather than controlled.

\begin{table}[t]
\centering
\caption{Pre-prefill compression (STC) and an off-the-shelf streaming model (LiveCC) under the \egostream protocol.}
\label{tab:compression_baselines}
\resizebox{\textwidth}{!}{%
\begin{tabular}{@{}l *{8}{c} *{8}{c} c@{}}
\toprule
 & \multicolumn{8}{c}{\textbf{Semantic Categories}} & \multicolumn{8}{c}{\textbf{Temporal Recall Regimes}} & \textbf{Proc.} \\
\cmidrule(lr){2-9} \cmidrule(lr){10-17}
\textbf{Method} & \textbf{Avg} & \textbf{SPAT} & \textbf{TEMP} & \textbf{DETL} & \textbf{EVNT} & \textbf{SOC} & \textbf{CAUS} & \textbf{PROS} & \textbf{Avg} & \textbf{Inst} & \textbf{Short} & \textbf{S-Mid} & \textbf{Mid} & \textbf{ML} & \textbf{Long} & \textbf{Ultra} & \textbf{(s)} \\
\midrule
\multicolumn{18}{@{}l}{\textit{Qwen3-VL (8B)}, \bFIFO\ \bSink\ \bHat} \\
\quad \bPrune\ \bIntraCos & 45.63 & \textbf{60.51} & 52.39 & 56.61 & 46.64 & 28.50 & 39.88 & \textbf{34.87} & 39.34 & 55.60 & 54.99 & 52.18 & 41.27 & 30.61 & 24.93 & 15.79 & 2.28 \\
\quad + STC~\cite{wang2026stc} & \textbf{47.18} & 59.07 & \textbf{56.46} & \textbf{58.66} & \textbf{51.23} & \textbf{30.52} & \textbf{41.08} & 33.22 & \textbf{41.90} & \textbf{56.75} & \textbf{55.43} & \textbf{54.71} & \textbf{45.53} & \textbf{34.75} & 23.77 & 22.37 & \textbf{0.23} \\
\midrule
\multicolumn{18}{@{}l}{\textit{Off-the-shelf streaming model}} \\
\quad LiveCC~\cite{chen2025livecc} (1\,FPS) & 36.59 & 40.58 & 39.71 & 40.75 & 39.39 & 30.47 & 31.03 & 34.23 & 34.50 & 42.90 & 39.04 & 36.08 & 33.25 & 32.37 & \textbf{30.25} & \textbf{27.63} & 0.34 \\
\bottomrule
\end{tabular}%
}
\end{table}

\section{Limitations and Societal Impact}
\egostream establishes a foundation benchmark for streaming episodic memory, but several limitations remain. Although we provide robust baselines on three 8B backbones, memory decay may vary across larger backbones and implementations. While \egostream can support open-ended QA, we focus on multiple choice, leaving free-form generation unexplored. Finally, \egostream inherits environmental and geographic biases from its source datasets and narrations. Streaming episodic memory may benefit assistive technologies, but also raises privacy and surveillance risks by capturing bystanders or sensitive interactions; these risks require mitigation before deployment.

%%%%%%%%%%%%%%%%%%%%%%%%%%%%%%%%%%%%%%%%%%%%%%%%%%%%%%%%%%%%

%\newpage
\section*{NeurIPS Paper Checklist}

\begin{enumerate}

\item {\bf Claims}
    \item[] Question: Do the main claims made in the abstract and introduction accurately reflect the paper's contributions and scope?
    \item[] Answer: \answerYes{} % Replace by \answerYes{}, \answerNo{}, or \answerNA{}.
    \item[] Justification: The paper’s claims are supported by extensive experiments on a novel egocentric streaming benchmark, introducing AVW and evaluating memory categories and recall regimes under the same streaming constraint and backbone.

\item {\bf Limitations}
    \item[] Question: Does the paper discuss the limitations of the work performed by the authors?
    \item[] Answer: \answerYes{} %\answerTODO{} % Replace by \answerYes{}, \answerNo{}, or \answerNA{}.
    \item[] Justification: We provide a dedicated Limitations paragraph at the end of the Appendix. %\justificationTODO{}

\item {\bf Theory assumptions and proofs}
    \item[] Question: For each theoretical result, does the paper provide the full set of assumptions and a complete (and correct) proof?
    \item[] Answer: 
    %\answerNo{} % Replace by \answerYes{}, \answerNo{}, or 
    \answerNA{}.
    \item[] Justification: The paper does not present formal theorems or theoretical guarantees.

    \item {\bf Experimental result reproducibility}
    \item[] Question: Does the paper fully disclose all the information needed to reproduce the main experimental results of the paper to the extent that it affects the main claims and/or conclusions of the paper (regardless of whether the code and data are provided or not)?
    \item[] Answer: \answerYes{} % Replace by \answerYes{}, \answerNo{}, or \answerNA{}.
    \item[] Justification: The paper describes dataset construction, AVW generation, recall-regime construction, evaluation protocol, cache layout, memory budgets, reducers, and implementation details. We also provide code.

\item {\bf Open access to data and code}
    \item[] Question: Does the paper provide open access to the data and code, with sufficient instructions to faithfully reproduce the main experimental results, as described in supplemental material?
    \item[] Answer: \answerYes{} % Replace by \answerYes{}, \answerNo{}, or \answerNA{}.
    \item[] Justification: We provide download scripts, annotations and the unified benchmark code in order to reproduce our experiments.

\item {\bf Experimental setting/details}
    \item[] Question: Does the paper specify all the training and test details (e.g., data splits, hyperparameters, how they were chosen, type of optimizer) necessary to understand the results?
    \item[] Answer: \answerYes{} % Replace by \answerYes{}, \answerNo{}, or \answerNA{}.
    \item[] Justification: Our dataset is designed as a testbed for evaluating the memory capabilities of streaming multimodal models.

\item {\bf Experiment statistical significance}
    \item[] Question: Does the paper report error bars suitably and correctly defined or other appropriate information about the statistical significance of the experiments?
    \item[] Answer: \answerNo{} % Replace by \answerYes{}, \answerNo{}, or \answerNA{}.
    \item[] Justification: The paper reports accuracy and diagnostic results but does not include confidence intervals, as the computational cost required to perform statistical inference at this scale is prohibitively high.

\item {\bf Experiments compute resources}
    \item[] Question: For each experiment, does the paper provide sufficient information on the computer resources (type of compute workers, memory, time of execution) needed to reproduce the experiments?
    \item[] Answer: \answerYes{} % Replace by \answerYes{}, \answerNo{}, or \answerNA{}.
    \item[] Justification: We report the hardware used and the per-frame execution time for our experiments.
    
\item {\bf Code of ethics}
    \item[] Question: Does the research conducted in the paper conform, in every respect, with the NeurIPS Code of Ethics \url{https://neurips.cc/public/EthicsGuidelines}?
    \item[] Answer: \answerYes{} % Replace by \answerYes{}, \answerNo{}, or \answerNA{}.
    \item[] Justification: We don't violate any term.

\item {\bf Broader impacts}
    \item[] Question: Does the paper discuss both potential positive societal impacts and negative societal impacts of the work performed?
    \item[] Answer: \answerYes{} % Replace by \answerYes{}, \answerNo{}, or \answerNA{}.
    \item[] Justification: The paper highlights how advancing episodic memory enables context-aware assistive wearable technologies, while acknowledging negative risks like privacy infringement from continuous recording and potential surveillance misuse, noting on-device processing as a mitigation.%\justificationTODO{DA SCRIVERE}
    
\item {\bf Safeguards}
    \item[] Question: Does the paper describe safeguards that have been put in place for responsible release of data or models that have a high risk for misuse (e.g., pre-trained language models, image generators, or scraped datasets)?
    \item[] Answer: \answerNA{} % Replace by \answerYes{}, \answerNo{}, or \answerNA{}.
    \item[] Justification: The paper does not release a pretrained generative model or a newly scraped raw Internet dataset.

\item {\bf Licenses for existing assets}
    \item[] Question: Are the creators or original owners of assets (e.g., code, data, models), used in the paper, properly credited and are the license and terms of use explicitly mentioned and properly respected?
    \item[] Answer: \answerYes{} % Replace by \answerYes{}, \answerNo{}, or \answerNA{}.
    \item[] Justification: We do not re-distribute existing dataset used. We cite original sources and provide download scripts, inviting users to comply to original licenses.

\item {\bf New assets}
    \item[] Question: Are new assets introduced in the paper well documented and is the documentation provided alongside the assets?
    \item[] Answer: \answerYes{} % Replace by \answerYes{}, \answerNo{}, or \answerNA{}.
    \item[] Justification: The paper introduces EgoStream annotations and a unified streaming evaluation framework. We provide code and documentation, along with details in the main paper and supplementary material.

\item {\bf Crowdsourcing and research with human subjects}
    \item[] Question: For crowdsourcing experiments and research with human subjects, does the paper include the full text of instructions given to participants and screenshots, if applicable, as well as details about compensation (if any)? 
    \item[] Answer: \answerYes{} % Replace by \answerYes{}, \answerNo{}, or \answerNA{}.
    \item[] Justification: We describe expert human validation and human-in-the-loop categorization in the supplementary material.

\item {\bf Institutional review board (IRB) approvals or equivalent for research with human subjects}
    \item[] Question: Does the paper describe potential risks incurred by study participants, whether such risks were disclosed to the subjects, and whether Institutional Review Board (IRB) approvals (or an equivalent approval/review based on the requirements of your country or institution) were obtained?
    \item[] Answer: \answerNA{} % Replace by \answerYes{}, \answerNo{}, or \answerNA{}.
    \item[] Justification: Human validation was conducted by expert volunteer researchers involved in the project. No personal or sensitive information was collected from validators, and the task consisted of reviewing existing benchmark annotations through the provided interface. Since this did not constitute a human-subject experiment and did not involve external annotation labor, no compensation scheme or IRB approval was required under the applicable institutional procedures. experiments.

\item {\bf Declaration of LLM usage}
    \item[] Question: Does the paper describe the usage of LLMs if it is an important, original, or non-standard component of the core methods in this research? Note that if the LLM is used only for writing, editing, or formatting purposes and does \emph{not} impact the core methodology, scientific rigor, or originality of the research, declaration is not required.
    %this research? 
    \item[] Answer: \answerYes{} % Replace by \answerYes{}, \answerNo{}, or \answerNA{}.
    \item[] Justification: LLMs are part of the methodology. These uses are disclosed in the dataset construction and appendix sections.

\end{enumerate}

\end{document}